%% file: main.tex
\documentclass[10pt,twocolumn,letterpaper]{article}

\usepackage{iccv}              
\usepackage[accsupp]{axessibility}
\usepackage{times}
\usepackage{epsfig}
\usepackage{graphicx}
\usepackage{amsmath}
\usepackage{amssymb}
\usepackage{graphicx}
\usepackage{amsmath}
\usepackage{amssymb}
\usepackage{booktabs}
\usepackage{amsthm}
\usepackage{algorithm}
\usepackage{algorithmic}
\usepackage{nicefrac}       
\usepackage{microtype} 
\usepackage{multirow}
\usepackage{xcolor}
\usepackage{color, colortbl}
\definecolor{Gray}{gray}{0.9}

\usepackage[hang,flushmargin]{footmisc}

\usepackage{babel}
\providecommand{\lemmaname}{Lemma}
\providecommand{\theoremname}{Theorem}

\theoremstyle{plain}
\newtheorem{thm}{\protect\theoremname}
\theoremstyle{plain}

\newcolumntype{a}{>{\columncolor{Gray}}c}
\newcolumntype{H}{>{\setbox0=\hbox\bgroup}c<{\egroup}@{}}
\newcolumntype{Z}{>{\setbox0=\hbox\bgroup}c<{\egroup}@{\hspace*{-\tabcolsep}}}

\usepackage[pagebackref,breaklinks,colorlinks,allcolors=iccvblue]{hyperref}
\usepackage[capitalize]{cleveref}
\crefname{section}{Sec.}{Secs.}
\Crefname{section}{Section}{Sections}
\Crefname{table}{Table}{Tables}
\crefname{table}{Tab.}{Tabs.}

\definecolor{iccvblue}{rgb}{0.21,0.49,0.74}

\def\paperID{15406} 
\def\confName{ICCV}
\def\confYear{2025}

\title{Beyond Losses Reweighting: Empowering Multi-Task Learning\\ via the Generalization Perspective}

\author{
Hoang Phan$^{1,*}$, Lam Tran$^{2,*}$, Quyen Tran$^{2}$, Ngoc Tran$^{3}$, Tuan Truong$^{2,\dagger}$, \\ Qi Lei$^{1}$,  Nhat Ho$^{4}$, Dinh Phung$^{2,5}$, Trung Le$^{5}$ \\
\small{
$^1$New York University, $^2$Qualcomm AI Research$^{+}$, $^3$Vanderbilt University, $^4$University of Texas at Austin, $^5$Monash University}
}

\begin{document}
\maketitle
\def\thefootnote{\textsuperscript{$*$}}\footnotetext{Equal contributions.}
\def\thefootnote{\textsuperscript{$\dagger$}}\footnotetext{Work done while at Qualcomm AI Research.}
\def\thefootnote{\textsuperscript{$+$}}\footnotetext{Qualcomm AI Research is an initiative of Qualcomm Technologies, Inc.}

\input{sec/0_abstract}    
\input{sec/1_intro}

\input{sec/2_related_work}
\input{sec/4_method}

\input{sec/5_experiment}

\input{sec/6_conclusion}

\clearpage
\section*{Acknowledgements}
We thank Chau Pham for his contributions to the scene understanding experiments. TL was supported by ARC DP23 grant DP230101176 and by the Air Force Office of Scientific Research under award number FA2386-23-1-4044.  QL acknowledges support of NSF DMS-2523382 and DOE Office of Science under Award \#DE-SC0024721.
{
    \small
\bibliographystyle{ieeenat_fullname}
    \bibliography{ref}
}
\input{sec/X_suppl}

\end{document}

%% file: sec/0_abstract.tex
\begin{abstract}

Multi-task learning (MTL) trains deep neural networks to optimize several objectives simultaneously using a shared backbone, which leads to reduced computational costs, improved data efficiency, and enhanced performance through cross-task knowledge sharing. Although recent gradient manipulation techniques aim to find a common descent direction that benefits all tasks, conventional empirical loss minimization still leaves models vulnerable to overfitting and gradient conflicts. To address this, we introduce a novel MTL framework that leverages weight perturbation to regulate gradient norms, thus improving generalization. By adaptively modulating weight perturbations, our approach harmonizes task-specific gradients, reducing conflicts and encouraging more robust learning across tasks. Theoretical insights reveal that controlling the gradient norm through weight perturbation directly contributes to better generalization. Extensive experiments across diverse applications demonstrate that our method significantly outperforms existing gradient-based MTL techniques in terms of task performance and overall model robustness.
\end{abstract}

%% file: sec/1_intro.tex
\section{Introduction}
\label{sec:intro}

Over the past few years, deep learning has emerged as a powerful tool for functional approximation, demonstrating superior performance and even surpassing human abilities in various applications. Despite these impressive achievements, training massive independent neural networks for individual tasks demands significant computational and storage resources, as well as extended runtime. Consequently, multi-task learning has become a preferable approach in many 
situations~\citep{zhang2014facial, liu2019end, wang2020gradient} as it aims to learn a shared  network among tasks, reducing redundant feature calculations while promoting positive task transfer.

However, learning such a shared backbone faces performance degradation due to gradient conflict \citep{yu2020gradient}, where task-specific gradients may differ in direction and magnitude, resulting in tasks canceling each other and a subset of tasks being under-optimized. To tackle this, a common approach is to manipulate task gradients to find a better update direction so that all task losses decrease in a more balanced manner. This has been found to consistently exhibit improved performance \citep{ banfair, liu2023famo}.
However, existing state-of-the-art methods in this vein ~\citep{sener2018multi, yu2020gradient, liu2021conflict, liu2021towards, javaloy2021rotograd, navon2022multi} often overlook the geometrical properties of the loss landscape, focusing solely on minimizing the empirical error in the optimization process, which can easily be prone to overfitting problems \citep{kaddour2022fair, zhao2022penalizing}.

Meanwhile, the overfitting problem in modern neural networks is often attributed to high-dimensional and nonconvex loss functions, which result in complex loss landscapes containing multiple local optima. Consequently, understanding the loss surface is crucial for training robust models, and developing flat minimizers remains one of the most effective strategies~\citep{keskar2017large, kaddour2022fair, li2022analyzing, lyu2022understanding}. Specifically, recent studies~\citep{he2019asymmetric, zheng2021regularizing} demonstrate that directly minimizing empirical risk often leads to a loss landscape with many sharp minima, resulting in poor generalization to unseen data. This issue is apparently further exacerbated when optimizing multiple objectives simultaneously, as in the context of multi-task learning. In fact, sharp minima of each constituent objective might appear at different locations, which can result in large generalization errors on the associated task. Therefore, finding a common flat and low-loss valued region for all tasks is desirable for improving the current methods of multi-task learning.

\begin{figure*}[t]
    \centering
     \includegraphics[width=.8\textwidth, trim=.cm 0cm 2.cm 0cm,clip]{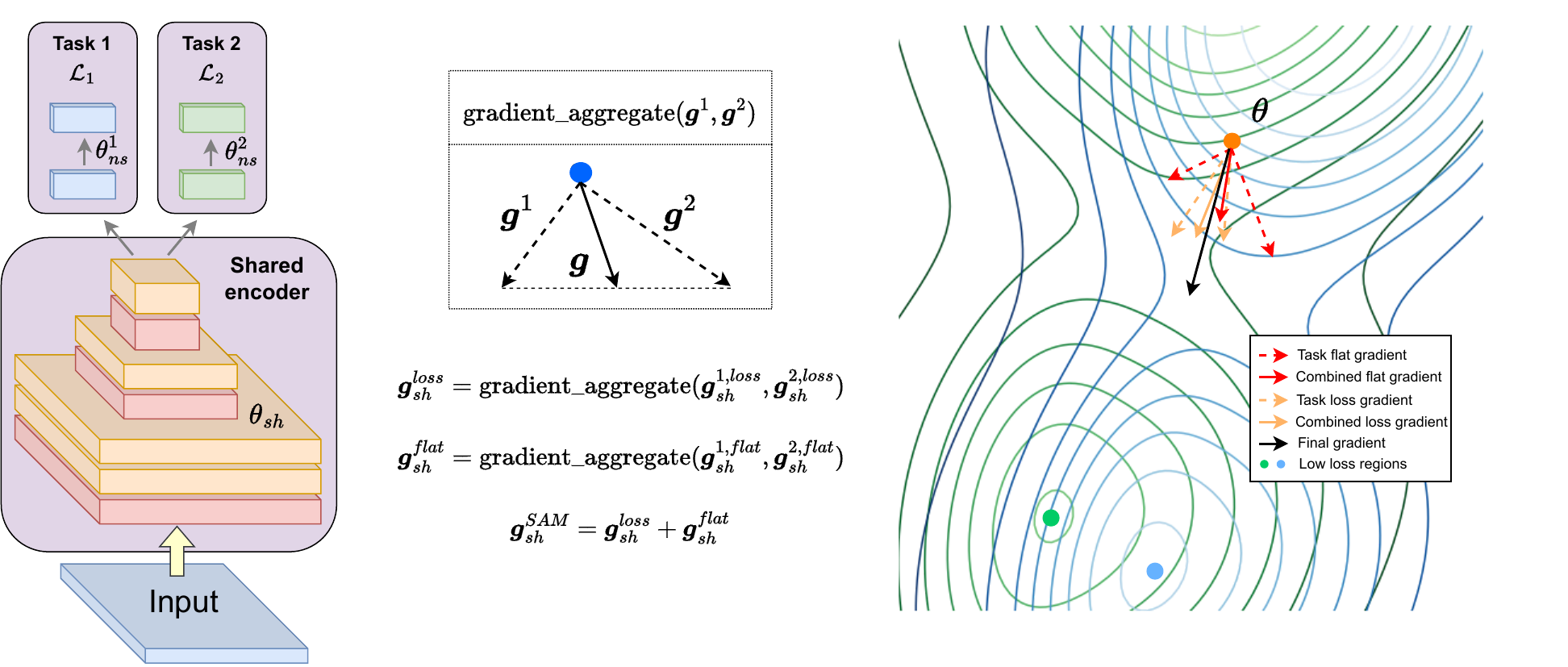}
    \caption{ We demonstrate our framework in a 2-task problem.  For the shared part, \textit{task-based flat gradients}  (\textcolor{red}{red} dashed arrows) steer the model away from sharp areas, while \textit{task-based loss gradients} (\textcolor{orange}{orange} dashed arrows) lead the model into their corresponding low-loss regions. In our method, we aggregate them to find the combined flat gradient $\boldsymbol{g}^{flat}_{sh}$ and combined loss gradient $\boldsymbol{g}^{loss}_{sh}$, respectively. Finally, we add these two output gradients to target the joint low-loss and flat regions across the tasks. Conversely, updating task-specific non-shared parts is straightforward and much easier as only one objective is involved.
    }
    \label{fig:overview}
        \vspace*{-\baselineskip}
\end{figure*}

\textbf{Contribution.} To address the above desideratum, we propose a novel MTL training method that enhances existing gradient manipulation strategies by promoting the learning of flat loss landscapes across all tasks. Specifically, we penalize each task's sharpness, the gap between the largest and the empirical losses within a weight perturbation ~\citep{zheng2021regularizing, foret2021sharpnessaware}, to improve the generalization of all tasks. This is theoretically supported by the generalization error in Theorem~\ref{thm:moo-sam}, which shows that our approach not only orients the model toward the joint low empirical loss value across tasks but also encourages the model to reach the task-based flat regions. Importantly, our approach is model-agnostic and compatible with current gradient-based MTL approaches (see Figure~\ref{fig:overview} for an overview of our approach). By using our proposed framework, gradient conflict across tasks is significantly mitigated, which is the goal of recent gradient-based MTL studies in alleviating negative transfer between tasks. Finally, we conduct comprehensive experiments on a variety of applications to demonstrate the merit of our approach for improving not only task performance but also model robustness and calibration. Last but not least, to the best of our knowledge, ours is the first work to improve multi-task learning by investigating the geometrical properties of the model loss landscape.



%% file: sec/2_related_work.tex
\section{Related Work}
\label{sec:related_work}

\subsection{Multi-task learning}
In multi-task learning (MTL), we often aim to jointly train a single model to tackle multiple different but correlated tasks. It has been proven in prior work~\citep{caruana1997multitask, liu2019end, liu2019mt-dnn, ruder2017overview} that
 it can not only enhance the overall performance but also reduce the memory usage and speed up the inference process. Previous studies on MTL often employ a hard parameter-sharing mechanism along with lightweight, task-specific modules to handle multiple tasks. 
 

 \textbf{Pareto multi-task learning.} Originating from Multiple-Gradient Descent Algorithm (MGDA), a popular line of gradient-based MTL methods aims to find Pareto stationary solutions, where it is impossible to improve model performance on any particular task without diminishing performance on another~\citep{sener2018multi}. Moreover, recent studies suggest exploring the entire Pareto front by learning diverse solutions~\citep{ lin2019pareto, liu2021profiling, mahapatra2020multi, mahapatra2021exact, phan2022stochastic}, or profiling the entire Pareto front with a hyper-network~\citep{lin2020controllable, navon2021learning}. While these methods are theoretically grounded and guaranteed to converge to Pareto-stationary points, the experimental results are often limited and lack comparisons in practical settings.

\textbf{Loss and gradient balancing.} Another branch of preliminary work in MTL capitalizes on the idea of dynamically reweighting loss functions based on gradient magnitudes~\citep{chen2018gradnorm}, task homoscedastic uncertainty~\citep{kendall2018multi}, or difficulty prioritization~\citep{guo2018dynamic} to balance the gradients across tasks. More recently, PCGrad \citep{yu2020gradient} introduces a gradient manipulation procedure to avoid conflicts among tasks by projecting
random task gradients onto the normal plane of the other. Similarly,~\citep{liu2021conflict} proposes a provably convergent method to minimize the average loss, and~\citep{liu2021towards}  calculates loss-scaling coefficients such that the combined gradient has equal-length projections onto individual task gradients.

\subsection{Flat minima}
\label{sec:flat}
    Flat minimizer has been found to improve the generalization ability of neural networks because it enables models to find wider local minima, by which they will be more robust against shifts between train and test losses~\citep{DBLP:conf/iclr/JiangNMKB20, DBLP:conf/nips/PetzkaKASB21, DBLP:conf/uai/DziugaiteR17}. This relationship between generalization ability and the width of minima is theoretically and empirically studied in many studies ~\citep{DBLP:conf/nips/HochreiterS94, neyshabur2017exploring, dinh2017sharp, fort2019emergent}, and subsequently, a variety of methods seeking flat minima have been proposed~\citep{DBLP:conf/iclr/PereyraTCKH17, Chaudhari2017EntropySGDBG, DBLP:conf/iclr/KeskarMNST17, DBLP:conf/uai/IzmailovPGVW18}. 

    Recently, SAM~\citep{foret2021sharpnessaware}, which seeks flat regions by explicitly minimizing the worst-case loss around the current model, has received significant attention due to its effectiveness and scalability compared to previous methods. Particularly, it has been exploited in a variety of tasks and domains, such as domain generalization ~\citep{cha2021swad, wang2023sharpness, zhang2023flatness}, federated learning \citep{caldarola2022improving,qu2022generalized}, Bayesian networks \citep{nguyen2023flat, llenhoff2023sam}, meta-learning \citep{abbas2022sharp}.
    In addition, SAM shows its generalization ability in vision models~\citep{chen2021vision} and language models~\citep{bahri-etal-2022-sharpness}.
    However, these studies have only focused on single-task problems. Closer to our setting are the works in \citep{deng2021flattening, yang2023data} that apply SAM to Continual Learning, but their focus is the relationship between flatness and catastrophic forgetting.
    In this work, we leverage SAM's principle to develop theory and devise practical methods, allowing for seeking flat minima in gradient-based multi-task learning models.
    


%% file: sec/4_method.tex
\section{Methodology}
\label{sec:framework}

This section outlines our proposed framework for enhancing gradient-based MTL methods. We begin by recalling the goal of multi-task learning and then establish upper bounds for the general loss of each task. Based on these bounds, we develop our framework to improve model generalization by guiding it toward flatter regions for each task.
\subsection{Multi-task Learning and Gradient-based methods}
In multi-task learning, we are given a data-label distribution $\mathcal{D}$ from which we can sample a training set $\mathcal{S}= \{(\boldsymbol{x}_i,y_i^1,...,y_i^m)_{i=1}^n\}$, where $\boldsymbol{x}_i$ is a data example and $y_i^1,...,y_i^m$ are the labels of the tasks $1,2,...,m$ respectively. 

The model for each task $\boldsymbol{\theta}^i = [\boldsymbol{\theta}_{sh}, \boldsymbol{\theta}^i_{ns}]$ consists of the shared part $\boldsymbol{\theta}_{sh}$ and the individual non-shared part $\boldsymbol{\theta}^i_{ns}$. We denote the general loss for the task $i$ by $\mathcal{L}^i_{\mathcal{D}}(\boldsymbol{\theta}^i)$, while its empirical loss over the training set $\mathcal{S}$ by $\mathcal{L}^i_{\mathcal{S}}(\boldsymbol{\theta}^i)$.
Existing work in MTL, typically MGDA \citep{sener2018multi}, PCGrad \citep{yu2020gradient}, CAGrad \citep{liu2021conflict}, and IMTL \citep{liu2021towards}, aim to find a model that simultaneously minimizes the empirical losses for all tasks:
\begin{equation} \label{eq:mtl_obj}
\min_{\boldsymbol{\theta}_{sh},\boldsymbol{\theta}_{ns}^{1:m}}\left[\mathcal{L}_{\mathcal{S}}^{1}\left(\boldsymbol{\theta}^{1}\right),...,\mathcal{L}_{\mathcal{S}}^{m}\left(\boldsymbol{\theta}^{m}\right)\right],
\end{equation}
by calculating gradient $\boldsymbol{g}^i$ for i-th task ($i \in [m]$). The current model parameter is then updated using the unified gradient $\boldsymbol{g}= \text{gradient\_aggregate}(\boldsymbol{g}^{1}, \boldsymbol{g}^{2}, \dots, \boldsymbol{g}^{m} )$, where the generic operation {gradient\_aggregate} combines multiple task gradients, as proposed in gradient-based MTL studies. Details on this operation can be found in the Appendix.

However, prior works only focused on minimizing the empirical losses and tend to be overfitting. To alleviate this, inspired by ~\citep{foret2021sharpnessaware, pmlr-v139-kwon21b, zheng2021regularizing, wu2020adversarial},
 it is desirable to develop sharpness-aware MTL approaches wherein the task models simultaneously seek low loss and flat regions, which is discussed below.

\subsection{Sharpness minimization for MTL} 
Intuitively, flat minima are those where neighboring points also exhibit low loss values. One effective way to find such minima is to minimize the worst-case perturbation loss, as demonstrated in ~\citep{foret2021sharpnessaware, pmlr-v139-kwon21b}. Here, we propose applying this concept to each task objective in MTL. Formally, the worst-case loss for each task is defined as follows:
\begin{align}
    \underset{||\boldsymbol{\epsilon}_{sh}||_2 \leq \rho_{sh}}{\max} \bigg[\underset{||\boldsymbol{\epsilon}_{ns}^i||_2 \leq \rho_{ns}}{\max} \mathcal{L}^i_S\left({\boldsymbol{\theta}_{sh}+\boldsymbol{\epsilon}}_{sh}, \boldsymbol{\theta}_{ns}^i+\boldsymbol{\epsilon}_{ns}^i\right)\bigg]_{i=1}^m \label{worst-case-loss-mtl}
\end{align}
where $||\cdot||_2$ denotes the $l_2$ norm, $\rho_{sh}$ and $\rho_{ns}$ represent the radii of the neighborhoods for the shared and non-shared parts, respectively.

The formulation of the worst-case loss in Eq.~(\ref{worst-case-loss-mtl}) differs from that in the single-task setting, as it involves multiple objective functions, each consisting of shared and individual non-shared parameters. This complexity makes extending the generalization error bound in ~\citep{foret2021sharpnessaware} non-trivial. In the next sub-section, we provide such bounds for the true risks in the context of MTL, highlighting the concept of sharpness for the shared and non-shared parts.


\subsection{Theoretical development}
We informally state our main theorem that bounds the generalization performance of individual tasks by the empirical error on the training set: 




\begin{thm} \label{thm:main_theo}
 For any perturbation radius $\rho_{sh}, \rho_{ns} > 0$, under the bounded-loss and mild assumptions, with probability $1-\delta$ (over the choice of training set $\mathcal{S} \sim \mathcal{D}$) we obtain
\begin{multline}
\left[\mathcal{L}_{\mathcal{D}}^{i}\left(\boldsymbol{\theta}^{i}\right)\right]_{i=1}^{m}\leq
\max_{\Vert\boldsymbol{\epsilon}_{sh}\Vert_{2}\leq\rho_{sh}} \biggl[ \\
\max_{\Vert\boldsymbol{\epsilon}_{ns}^{i}\Vert_2\leq\rho_{ns}}\mathcal{L}_{\mathcal{S}}^{i}\left(\boldsymbol{\theta}_{sh}+\boldsymbol{\epsilon}_{sh},\boldsymbol{\theta}_{ns}^{i}+\boldsymbol{\epsilon}_{ns}^{i}\right) +f^{i}\left(\Vert\boldsymbol{\theta}^{i}\Vert_{2}^{2}\right)\biggr]_{i=1}^{m},
\end{multline}
\label{main_bound}
where $f^i: \mathbb{R}_+ \rightarrow \mathbb{R}_+, i \in [m]$ are strictly increasing functions. 
\label{thm:moo-sam}
\end{thm}

Theorem \ref{thm:moo-sam} establishes the connection between the generalization error of each task and its empirical training error via worst-case perturbation on the parameter space. The formally stated theorem and proof are provided in the Appendix. We note that the worst-case shared perturbation $\boldsymbol{\epsilon}_{sh}$ is common for all tasks, while the worst-case non-shared perturbation $\boldsymbol{\epsilon}^i_{ns}$ is tailored for each task $i$. This requires addressing multiple objectives with both non-shared and shared components in our theory development.

Additionally and importantly, the proof in \cite{foret2021sharpnessaware} invokes the PAC-Bayesian generalization bound \cite{mcallester1999pac}; hence, it only applies to the 0-1 loss in the binary classification setting. 
In contrast, as a theoretical contribution, we employ a more general PAC-Bayesian generalization bound \citep{JMLR:v17:15-290}, which only requires the loss to be bounded, thus tackling a notably wider range of losses in MTL. Hence, our theory development is not a trivial extension of prior works due to the nature of multi-objective optimization. 

\subsection{Practical method}
Guided by Theorem \ref{thm:main_theo}, we first aim to solve the bi-level maximization problem for each task loss as follows:
\begin{align}
&\max_{\Vert\boldsymbol{\epsilon}_{sh}\Vert_{2}\leq\rho_{sh}} \biggl[ 
\max_{ \Vert\boldsymbol{\epsilon}_{ns}^{i}\Vert_2\leq\rho_{ns}}\mathcal{L}_{\mathcal{S}}^{i}\left(\boldsymbol{\theta}_{sh}+\boldsymbol{\epsilon}_{sh},\boldsymbol{\theta}_{ns}^{i}+\boldsymbol{\epsilon}_{ns}^{i}\right) \biggr]  \label{bilevel_max} \\
&\approx\max_{\Vert\boldsymbol{\epsilon}_{sh}\Vert_{2}\leq\rho_{sh}} \biggl[ 
\max_{ \Vert\boldsymbol{\epsilon}_{ns}^{i}\Vert_2\leq\rho_{ns}}\mathcal{L}_{\mathcal{S}}^i(\boldsymbol{\theta}_{sh}, \boldsymbol{\theta}_{ns}^i)+ \nonumber \\
&(\boldsymbol{\epsilon}_{ns}^i)^{\mathbb{T}}\nabla_{\boldsymbol{\theta}_{ns}^i}\mathcal{L}_{\mathcal{S}}^i(\boldsymbol{\theta}_{sh}, \boldsymbol{\theta}_{ns}^i)+(\boldsymbol{\epsilon}_{sh})^{\mathbb{T}}\nabla_{\boldsymbol{\theta}_{sh}}\mathcal{L}_{\mathcal{S}}^i(\boldsymbol{\theta}_{sh}, \boldsymbol{\theta}_{ns}^i)\biggl], \\
&=\max_{\Vert\boldsymbol{\epsilon}_{sh}\Vert_{2}\leq\rho_{sh}} \biggl[\mathcal{L}_{\mathcal{S}}^i(\boldsymbol{\theta}_{sh}, \boldsymbol{\theta}_{ns}^i)+(\boldsymbol{\epsilon}_{sh})^{\mathbb{T}}\nabla_{\boldsymbol{\theta}_{sh}}\mathcal{L}_{\mathcal{S}}^i(\boldsymbol{\theta}_{sh}, \boldsymbol{\theta}_{ns}^i) \nonumber \\
&+\max_{ \Vert\boldsymbol{\epsilon}_{ns}^{i}\Vert_2\leq\rho_{ns}}(\boldsymbol{\epsilon}_{ns}^i)^{\mathbb{T}}\nabla_{\boldsymbol{\theta}_{ns}^i}\mathcal{L}_{\mathcal{S}}^i(\boldsymbol{\theta}_{sh}, \boldsymbol{\theta}_{ns}^i)\biggl] \label{taylor_bi_level}
\end{align}
where approximation is the first order Taylor expansion with a note that $\boldsymbol{\epsilon}_{sh}$ and $\boldsymbol{\epsilon}_{ns}^i$ are independent.
Now following the dual norm problem as in \citep{foret2021sharpnessaware}, the solution for the inner maximization is
\begin{align}
    \boldsymbol{\epsilon}_{ns}^{i,*} = \frac{\rho_{ns}\nabla_{\boldsymbol{\theta}_{ns}^i}\mathcal{L}_{\mathcal{S}}^i(\boldsymbol{\theta}_{sh}, \boldsymbol{\theta}_{ns}^i)}{\Vert\nabla_{\boldsymbol{\theta}_{ns}^i}\mathcal{L}_{\mathcal{S}}^i(\boldsymbol{\theta}_{sh}, \boldsymbol{\theta}_{ns}^i)\Vert_2} \label{ep_ns_optimal}
\end{align}
Next, our goal is to find $\boldsymbol{\epsilon}_{sh}$ that simultaneously maximizes the following objectives 
\begin{align}
&\max_{\Vert\boldsymbol{\epsilon}_{sh}\Vert_{2}\leq\rho_{sh}}\biggl[\mathcal{L}_{\mathcal{S}}^i(\boldsymbol{\theta}_{sh}, \boldsymbol{\theta}_{ns}^i)+(\boldsymbol{\epsilon}_{sh})^{\mathbb{T}}\nabla_{\boldsymbol{\theta}_{sh}}\mathcal{L}_{\mathcal{S}}^i(\boldsymbol{\theta}_{sh}, \boldsymbol{\theta}_{ns}^i) \nonumber \\
&\quad\quad\quad\quad+\rho_{ns}\Vert\nabla_{\boldsymbol{\theta}_{ns}^i}\mathcal{L}_{\mathcal{S}}^i(\boldsymbol{\theta}_{sh}, \boldsymbol{\theta}_{ns}^i)\Vert_2\biggr]_{i=1}^m \\
&= \max_{\Vert\boldsymbol{\epsilon}_{sh}\Vert_{2}\leq\rho_{sh}}\biggl[(\boldsymbol{\epsilon}_{sh})^{\mathbb{T}}\nabla_{\boldsymbol{\theta}_{sh}}\mathcal{L}_{\mathcal{S}}^i(\boldsymbol{\theta}_{sh}, \boldsymbol{\theta}_{ns}^i)\biggr]_{i=1}^m + const \label{one-share} \\
&\leq \biggl[\max_{\Vert\boldsymbol{\epsilon}_{sh}^i\Vert_{2}\leq\rho_{sh}}(\boldsymbol{\epsilon}_{sh}^i)^{\mathbb{T}}\nabla_{\boldsymbol{\theta}_{sh}}\mathcal{L}_{\mathcal{S}}^i(\boldsymbol{\theta}_{sh}, \boldsymbol{\theta}_{ns}^i)\biggr]_{i=1}^m + const, \label{each-share}
\end{align}
where $const$ is the constant independent of $\boldsymbol{\epsilon}_{sh}$. It is non-trivial to find the closed-form solution for this problem because the worst-cased perturbation $\boldsymbol{\epsilon}_{sh}$ is shared among the tasks (cf. Eq.~(\ref{one-share})). We hence relax it by separately finding $\boldsymbol{\epsilon}_{sh}^i$ for each task (cf. Eq.~(\ref{each-share})).  
Similarly to the inner maximization, we now have 
\begin{align}
    \boldsymbol{\epsilon}_{sh}^{i,*} = \frac{\rho_{sh}\nabla_{\boldsymbol{\theta}_{sh}}\mathcal{L}_{\mathcal{S}}^i(\boldsymbol{\theta}_{sh}, \boldsymbol{\theta}_{ns}^i)}{\Vert\nabla_{\boldsymbol{\theta}_{sh}}\mathcal{L}_{\mathcal{S}}^i(\boldsymbol{\theta}_{sh}, \boldsymbol{\theta}_{ns}^i)\Vert_2}.\label{ep_sh_optimal}
\end{align}
Note that this relaxation gives us an upper bound for the optimal value of the maximization problem in Eq.~(\ref{one-share}) because $(\boldsymbol{\epsilon}_{sh}^*)^{\mathbb{T}}\nabla_{\boldsymbol{\theta}_{sh}}\mathcal{L}_{\mathcal{S}}^i(\boldsymbol{\theta}_{sh}, \boldsymbol{\theta}_{ns}^i) \leq (\boldsymbol{\epsilon}_{sh}^{i,*})^{\mathbb{T}}\nabla_{\boldsymbol{\theta}_{sh}}\mathcal{L}_{\mathcal{S}}^i(\boldsymbol{\theta}_{sh}, \boldsymbol{\theta}_{ns}^i), \forall i=1,\cdots m$. Therefore, we aim to minimize this upper-bound multi-objectives in practice. 
Finally, substituting Eq.~(\ref{ep_ns_optimal}) and Eq.~(\ref{ep_sh_optimal}) back to Eq.~(\ref{taylor_bi_level}), the bi-level maximization in Eq.~(\ref{bilevel_max}) has the following approximate solution:
\begin{align}
    &\biggl[\mathcal{L}_{\mathcal{S}}^{i}\left(\boldsymbol{\theta}_{sh}+\boldsymbol{\epsilon}_{sh}^{i,*},\boldsymbol{\theta}_{ns}^{i}+\boldsymbol{\epsilon}_{ns}^{i,*}\right)\biggr]_{i=1}^m \label{worst-case-mtl} \\
    & \approx \biggl[\mathcal{L}_{\mathcal{S}}^i(\boldsymbol{\theta}_{sh}, \boldsymbol{\theta}_{ns}^i)\biggr]_{i=1}^m + \rho_{sh}\biggl[\Vert\nabla_{\boldsymbol{\theta}_{sh}}\mathcal{L}_{\mathcal{S}}^i(\boldsymbol{\theta}_{sh}, \boldsymbol{\theta}_{ns}^i)\Vert_2\biggr]_{i=1}^m \nonumber \\
    &\quad\quad\quad\quad +\rho_{ns}\biggl[\Vert\nabla_{\boldsymbol{\theta}_{ns}^i}\mathcal{L}_{\mathcal{S}}^i(\boldsymbol{\theta}_{sh}, \boldsymbol{\theta}_{ns}^i)\Vert_2\biggr]_{i=1}^m
    \label{3-obj}
\end{align}
The above equation shows that sharpness-aware minimization for MTL requires us to minimize the following sub-objectives for each task: (i) the conventional task-loss, (ii) the norm of gradient w.r.t the task-specific head, and (iii) the norm of gradient w.r.t the shared backbone. Minimizing the first objective leads the model toward local, but possibly sharp minima, which causes overfitting and severe task conflict. Hence, to reduce this, the last two objectives help steer the model away from such sharp minima, favoring flatter ones, i.e. gradient norm minimization seeks flat minima.
However, in comparison with traditional MTL, we are tasked with more objectives, possibly leading to more conflicts between all of them. In the following, we will present how we solve this problem by separately treating each sub-objectives for all tasks.

\subsubsection{Update the non-shared parts} 
Since each task $i$ has its own task-specific head $\boldsymbol{\theta}_{ns}^i$, we can find flat minima for $\boldsymbol{\theta}_{ns}^i$ as follows:
\begin{align}
    \boldsymbol{g}_{ns}^{i,\text{SAM}} & =\nabla_{\boldsymbol{\theta}_{ns}^{i}}\mathcal{L}_{\mathcal{S}}^{i}\left(\boldsymbol{\theta}_{sh}+\boldsymbol{\epsilon}_{sh}^{i,*},\boldsymbol{\theta}_{ns}^{i}+\boldsymbol{\epsilon}_{ns}^{i,*}\right)\nonumber\\&
    \approx\nabla_{\boldsymbol{\theta}_{ns}^{i}}\mathcal{L}_{\mathcal{S}}^{i}\left(\boldsymbol{\theta}_{sh},\boldsymbol{\theta}_{ns}^{i}\right)\bigg|_{\boldsymbol{\theta}_{sh}=\boldsymbol{\theta}_{sh}+\boldsymbol{\epsilon}_{sh}^{i,*}, \boldsymbol{\theta}_{ns}^{i}=\boldsymbol{\theta}_{ns}^{i}+\boldsymbol{\epsilon}_{ns}^{i,*}}, \nonumber\\
\boldsymbol{\theta}_{ns}^{i} & =\boldsymbol{\theta}_{ns}^{i}-\eta\boldsymbol{g}_{ns}^{i,\text{SAM}}
\label{ns_update}
\end{align}
Note that computing gradient directly on the gradient norm sub-objective requires the heavy computation of Hessian matrix. Instead, we resort to the perturbed loss (\ref{worst-case-mtl}) and approximate its gradient w.r.t $\boldsymbol{\theta}_{ns}^i$ as in Eq.~(\ref{ns_update}). This procedure is similar to single-task SAM.

\subsubsection{Update the shared part}
This is a challenging task since as shown in Eq.~(\ref{3-obj}), we have to find a common $\boldsymbol{\theta}_{sh}$ to not only reduce losses of all tasks, but also to reduce their gradient norms. Specifically, define $\mathcal{L}_{loss}^i:=\mathcal{L}_{\mathcal{S}}^i(\boldsymbol{\theta}_{sh}, \boldsymbol{\theta}_{ns}^i)$ and $\mathcal{L}_{flat}^i:=\rho_{sh}\Vert\nabla_{\boldsymbol{\theta}_{sh}}\mathcal{L}_{\mathcal{S}}^i(\boldsymbol{\theta}_{sh}, \boldsymbol{\theta}_{ns}^i)\Vert_2 +\rho_{ns}\Vert\nabla_{\boldsymbol{\theta}_{ns}^i}\mathcal{L}_{\mathcal{S}}^i(\boldsymbol{\theta}_{sh}, \boldsymbol{\theta}_{ns}^i)\Vert_2$, we have $2\times m$ objectives in total. 

It has been shown in \citep{zhuang2022surrogate,wang2023sharpness} that there may exist conflict between $\mathcal{L}_{loss}^i$ and $\mathcal{L}_{flat}^i$, leading to a risk of increasing the loss when minimizing sharpness. This problem can worsen in the scope of MTL where conflicts can arise not only between task objectives (inter-conflict) but also between the two purposes of each task (intra-conflict). Inspired by this evidence and the inherently different goals between these two types of losses, we propose to consider them individually. Conceptually, we decompose the original MTL problem, $\text{MTL}[\mathcal{L}_{loss}^i + \mathcal{L}_{flat}^i]_{i=1}^m$, into two sub-MTLs, $\text{MTL}[\mathcal{L}_{loss}^i]_{i=1}^m + \text{MTL}[\mathcal{L}_{flat}^i]_{i=1}^m$. 

To solve each sub-MTL problem, we have to compute gradients of $\mathcal{L}_{loss}^i$ and $\mathcal{L}_{flat}^i$ w.r.t $\boldsymbol{\theta}_{sh}$. The former is straightforward, but the latter requires heavy Hessian computation. We bypass this by noticing from Eq.~(\ref{3-obj}) that the perturbed loss is the sum of $\mathcal{L}_{loss}^i$ and $\mathcal{L}_{flat}^i$. Hence, the gradient of $\mathcal{L}_{flat}^i$ can be approximated by the difference between the gradient of the perturbed loss and that of $\mathcal{L}_{loss}^i$. Formally,
\begin{align}
    &\boldsymbol{g}_{sh}^{i,\text{SAM}}  =\nabla_{\boldsymbol{\theta}_{sh}}\mathcal{L}_{\mathcal{S}}^{i}\left(\boldsymbol{\theta}_{sh}+\boldsymbol{\epsilon}_{sh}^{i,*},\boldsymbol{\theta}_{ns}^{i}+\boldsymbol{\epsilon}_{ns}^{i,*}\right)\nonumber\\
    &\approx\nabla_{\boldsymbol{\theta}_{sh}}\mathcal{L}_{\mathcal{S}}^{i}\left(\boldsymbol{\theta}_{sh},\boldsymbol{\theta}_{ns}^{i}\right)\bigg|_{\boldsymbol{\theta}_{sh}=\boldsymbol{\theta}_{sh}+\boldsymbol{\epsilon}_{sh}^{i,*}, \boldsymbol{\theta}_{ns}^{i}=\boldsymbol{\theta}_{ns}^{i}+\boldsymbol{\epsilon}_{ns}^{i,*}} \nonumber \\
    &\boldsymbol{g}_{sh}^{i,loss} = \nabla_{\boldsymbol{\theta}_{sh}}\mathcal{L}_{\mathcal{S}}^{i}\left(\boldsymbol{\theta}_{sh},\boldsymbol{\theta}_{ns}^{i}\right),
    \\
    &\boldsymbol{g}_{sh}^{i,flat} = \boldsymbol{g}_{sh}^{i,SAM} - \boldsymbol{g}_{sh}^{i,loss},
\end{align}
The purpose of the negative gradient $-\boldsymbol{g}_{sh}^{i,\text{loss}}$ is to orient the model to minimize the loss of the task $i$, while $-\boldsymbol{g}_{sh}^{i,\text{flat}}$ navigates the model to the task $i$'s flatter region. Therefore, the gradients $\boldsymbol{g}_{sh}^{i, \text{loss}}$ share a similar nature, making them likely congruent. A similar relationship holds for $\boldsymbol{g}_{sh}^{i, \text{flat}}$. To solve each sub-MTL, following gradient-based MTL methods that aggregate gradients such that their conflict is reduced, we aim to find a common direction that leads the joint low-valued losses for all tasks and the joint flatter region for them as:
\begin{align*}
\boldsymbol{g}_{sh}^{\text{loss}} & =\text{gradient\_aggregate}(\boldsymbol{g}_{sh}^{1,\text{loss}},\ldots,\boldsymbol{g}_{sh}^{m,\text{loss}}),\\
\boldsymbol{g}_{sh}^{\text{flat}} & =\text{gradient\_aggregate}(\boldsymbol{g}_{sh}^{1,\text{flat}},\ldots,\boldsymbol{g}_{sh}^{m,\text{flat}}).
\end{align*}
Finally, to combine two sub-MTL problems, we can similarly aggregate $\boldsymbol{g}_{sh}^{\text{loss}}$ and $\boldsymbol{g}_{sh}^{\text{flat}}$ based on gradient-based MTL methods. However, in practice, we find that simply adding them can work well in most cases (e.g., Ours vs Second-aggre in Table \ref{tab:mnist-abl}), so we adopt this strategy to save computation: $\boldsymbol{g}_{sh}^{\text{SAM}} =\boldsymbol{g}_{sh}^{\text{loss}}+\boldsymbol{g}_{sh}^{\text{flat}};\; \boldsymbol{\theta}_{sh}  =\boldsymbol{\theta}_{sh}-\eta\boldsymbol{g}_{sh}^{\text{SAM}}$.

The key steps of our proposed framework are summarized in Algorithm \ref{alg:moo-sam}, and the overall schema of our proposed method is demonstrated in Figure \ref{fig:overview}.

Note that one can apply gradient-based methods to remove intra-conflict between $\boldsymbol{g}_{sh}^{i,loss}$ and $\boldsymbol{g}_{sh}^{i,flat}$ for each task to obtain $\boldsymbol{g}_{sh}^i$, then aggregate them once more time. This strategy, however, is extremely computationally expensive as we have to use gradient-aggregator $m+1$ times, and still results in similar performance compared to our method, i.e., Each-aggre vs Ours in Table \ref{tab:mnist-abl}.

Another approach is to directly aggregate $\boldsymbol{g}_{sh}^{i,SAM}$ of each task, i.e, ignoring the intra-conflict. This might still result in a higher level of gradient conflict than in our method which considers intra-conflict. We empirically demonstrate 
this in Figure \ref{fig:decomposition_gradnorm} where the our strategy gains lower loss values and gradient norms than the direct strategy, and the effectiveness of our method in Table \ref{tab:aggregation}.

{ \begin{algorithm}
\caption{\small Sharpness minimization for multi-task learning}
\small \textbf{Input:}{  Model parameter $\boldsymbol{\theta}=[\boldsymbol{\theta}_{sh},\boldsymbol{\theta}_{ns}^{1:m}]$, perturbation radius $\rho=[\rho_{sh},\rho_{ns}]$, step size $\eta$ and a list of $m$ differentiable loss functions $\left\{\mathcal{L}^i\right\}_{i=1}^m$}. \\
\small\textbf{Output:}{ Updated parameter $\boldsymbol{\theta}^*$}
\begin{algorithmic}[1]

\FOR{task $i\in [m]$}
\STATE Compute gradient $\boldsymbol{g}^{i,\text{loss}}_{sh}, \boldsymbol{g}^i_{ns} \leftarrow \nabla_{\boldsymbol{\theta}} \mathcal{L}^i(\boldsymbol{\theta})$
\STATE Worst-case perturbation direction \\  $\boldsymbol{\epsilon}^i_{sh}=\rho_{sh} \cdot \boldsymbol{g}^{i,\text{loss}}_{sh} /\left\|\boldsymbol{g}^{i,\text{loss}}_{sh}\right\|$,  \quad$\boldsymbol{\epsilon}^i_{ns}=\rho_{ns} \cdot \boldsymbol{g}^{i}_{ns} /\left\|\boldsymbol{g}^{i}_{ns}\right\|$
\STATE Approximate SAM's gradient\\ 
\hspace*{10mm}$\boldsymbol{g}^{i, \text{SAM}}_{sh}, \boldsymbol{g}^{i, \text{SAM}}_{ns}=\left.\nabla \mathcal{L}^i(\boldsymbol{\theta}_{sh}+\boldsymbol{\epsilon}^i_{sh},\boldsymbol{\theta}_{ns}^i+\boldsymbol{\epsilon}^i_{ns})\right.$ 
\STATE Compute flat gradient \\  \hspace*{10mm}
$\boldsymbol{g}^{i,\text{flat}}_{sh} = \boldsymbol{g}^{i,\text{SAM}}_{sh} - \boldsymbol{g}^{i,\text{loss}}_{sh}$
\ENDFOR
\STATE 
Calculate combined update gradients:
\vspace{-3mm}
\begin{align*}
  \boldsymbol{g}_{sh}^{\text{loss}} & =\text{gradient\_aggregate}(\boldsymbol{g}^{1,\text{loss}}_{sh}, \boldsymbol{g}^{2,\text{loss}}_{sh}, \dots, \boldsymbol{g}^{m,\text{loss}}_{sh} )\\
  \boldsymbol{g}_{sh}^{\text{flat}} & =\text{gradient\_aggregate}(\boldsymbol{g}^{1,\text{flat}}_{sh}, \boldsymbol{g}^{2,\text{flat}}_{sh}, \dots, \boldsymbol{g}^{m,\text{flat}}_{sh} )
\end{align*}
\vspace*{-5mm}
\STATE Calculate shared gradient update $\boldsymbol{g}^{\text{SAM}}_{sh} = \boldsymbol{g}_{sh}^{\text{loss}} + \boldsymbol{g}_{sh}^{\text{flat}}$
\STATE Update model parameter\\ $\boldsymbol{\theta^*} = [\boldsymbol{\theta}_{sh}, \boldsymbol{\theta}_{ns}^{1:m}] - \eta [\boldsymbol{g}^{\text{SAM}}_{sh}, \boldsymbol{g}^{1:m,\text{SAM}}_{ns}]$
\end{algorithmic}
\label{alg:moo-sam}
\end{algorithm}
}
\vspace*{-\baselineskip}

%% file: sec/5_experiment.tex
\section{Experiments}
\label{sec:exp}


\begin{table*}[!ht]
\renewcommand{\arraystretch}{1.3}

\centering
\resizebox{\textwidth}{!}{%
\begin{tabular}{c|ccc|ccc|ccc}
\hline
 &
  \multicolumn{3}{c}{MultiFashion} &
  \multicolumn{3}{|c|}{MultiMNIST} &
  \multicolumn{3}{c}{MultiFashion+MNIST} \\  \cmidrule(lr){2-4} \cmidrule(lr){5-7} \cmidrule(lr){8-10}
  
\multirow{-2}{*}{Method} &
  Task 1 $\uparrow$&
  Task 2 $\uparrow$&
  Average $\uparrow$&
  Task 1 $\uparrow$&
  Task 2 $\uparrow$&
  Average $\uparrow$&
  Task 1 $\uparrow$&
  Task 2 $\uparrow$&
  Average $\uparrow$
  \\ \cline{1-1}
  STL &
  $ 87.10\pm 0.09$ &
  $86.20 \pm 0.06$ &
  $86.65 \pm 0.02 $ &
  $95.33 \pm 0.08$ &
  $94.16 \pm 0.04$ &
  $94.74 \pm 0.06$ &
  $98.40 \pm 0.02$ &
  $89.42 \pm 0.03$ &
  $93.91 \pm 0.02$ \\
\midrule
MGDA &
  $86.76 \pm 0.09$ &
  $85.78 \pm 0.36$ &
  $86.27 \pm 0.22$ &
  $95.62 \pm 0.02$ &
  $94.49 \pm 0.10$ &
  $ 95.05 \pm 0.06$ &
  $97.24 \pm 0.04$ &
  $88.19 \pm 0.13$ &
  $92.72 \pm 0.07$ \\
\rowcolor{Gray} F-MGDA &
  $\mathbf{88.12 \pm 0.11}$ &
  $\mathbf{87.35 \pm 0.11}$ &
  $\mathbf{ 87.73 \pm 0.09}$ &
  $\mathbf{96.37 \pm 0.06}$ &
  $\mathbf{94.99 \pm 0.06}$ &
  $\mathbf{95.68 \pm 0.00}$ &
  $\mathbf{97.30 \pm 0.09}$ &
  $\mathbf{89.26 \pm 0.14}$ &
  $\mathbf{93.28 \pm 0.03}$ \\ 
\midrule
PCGrad &
  $86.93 \pm 0.17$ &
  $86.20 \pm 0.14$ &
  $86.57 \pm 0.12$ &
  $95.71 \pm 0.03$ &
  $94.41 \pm 0.02$ &
  $95.06 \pm 0.02$ &
  $97.12 \pm 0.16$ &
  $88.45 \pm 0.08$ &
  $92.78 \pm 0.11$ \\
\rowcolor{Gray} F-PCGrad &
  $\mathbf{88.17 \pm 0.14}$ &
  $\mathbf{87.35 \pm 0.27}$ &
  $\mathbf{87.76 \pm 0.07}$ &
  $\mathbf{96.49 \pm 0.05}$ &
  $\mathbf{95.34 \pm 0.10}$ &
  $\mathbf{95.92 \pm 0.07}$ &
  $\mathbf{97.65 \pm 0.06}$ &
  $\mathbf{89.35 \pm 0.07^*}$ 
  & $\mathbf{93.50 \pm 0.01}$ \\ 
\midrule
CAGrad &
  $86.99 \pm 0.17$ &
  $86.04 \pm 0.15$ &
  $86.51 \pm 0.16$ &
  $95.62 \pm 0.05$ &
  $94.39 \pm 0.04$ &
  $95.01 \pm 0.04$ &
  $97.19 \pm 0.06$ &
  $88.18 \pm 0.14$ 
  &92.68 $ \pm 0.04 $ \\
  
\rowcolor{Gray} F-CAGrad &
  $\mathbf{88.19 \pm 0.19}$ &
  $\mathbf{87.45 \pm 0.13}$ &
  $\mathbf{87.82 \pm 0.10^*} $ &
  $\mathbf{96.54 \pm 0.02}$ &
  $\mathbf{95.36 \pm 0.04}$ &
  $\mathbf{95.95 \pm 0.01^*}$ &
  $\mathbf{97.82 \pm 0.05}$ &
  $\mathbf{89.26 \pm 0.22}$ 
  & $\mathbf{93.54 \pm 0.13^*}$ \\
  
\midrule  
 IMTL &
  $87.35 \pm 0.22$ &
  $86.45 \pm 0.09$ &
  $86.90 \pm 0.15$ &
  $95.93 \pm 0.09$ &
  $94.63 \pm 0.13$ &
  $95.28 \pm 0.02$ &
  $97.47 \pm 0.06$ &
  $88.46 \pm 0.11$ 
  &$ 92.97\pm 0.03$\\

\rowcolor{Gray} F-IMTL &
  $\mathbf{88.1 \pm 0.10}$ &
  $\mathbf{87.50 \pm 0.04^*}$ &
  $\mathbf{87.80 \pm 0.06}$ &
  $\mathbf{96.55 \pm 0.07^*}$ &
  $\mathbf{95.16 \pm 0.05}$ &
  $\mathbf{95.85 \pm 0.05}$ &
  $\mathbf{97.59 \pm 0.12}$ &
  $\mathbf{88.99 \pm 0.08}$
  &  $\mathbf{93.29 \pm 0.02}$  \\

\midrule  
 NashMTL &
  ${ 86.91\pm0.09 }$ &
  ${ 86.17\pm0.03 }$ &
    ${ 86.54\pm 0.04}$ &
  ${ 95.54\pm 0.00}$ &
  ${ 94.49\pm 0.09}$ &
    ${ 95.01\pm 0.05}$ &
  ${ 97.00\pm 0.18}$ &
  ${ 88.39\pm 0.16}$ 
  &  ${92.70 \pm0.02 }$  \\

\rowcolor{Gray} F-NashMTL &
  $\mathbf{ 88.21\pm 0.20^*}$ &
  $\mathbf{87.40 \pm 0.03}$ &
  $\mathbf{ 87.81\pm0.11 }$ &
  $\mathbf{96.47\pm0.03 }$ &
  $\mathbf{ 95.40\pm0.10^*}$ &
  $\mathbf{ 95.94\pm0.05 }$ &
  $\mathbf{97.63 \pm 0.12}$ &
  $\mathbf{ 89.33\pm 0.12}$
  &  $\mathbf{ 93.48\pm 0.07}$  \\
  
\midrule  
 FairGrad &
  ${ 86.85\pm 0.15}$ &
  ${ 86.17\pm 0.07}$ &
  ${86.51 \pm 0.04}$ &
  ${ 95.55\pm 0.15}$ &
  ${ 94.29\pm 0.09}$ &
  ${ 94.92\pm 0.11}$ &
  ${ 97.02\pm 0.06}$ &
  ${88.37 \pm 0.16}$
  &  ${ 92.70\pm0.10 }$  \\

\rowcolor{Gray} F-FairGrad &
  $\mathbf{ 88.05\pm 0.08}$ &
  $\mathbf{ 87.41\pm 0.14}$ &
  $\mathbf{ 87.73\pm 0.04 }$ &
  $\mathbf{ 96.48\pm 0.07}$ &
  $\mathbf{ 95.34\pm 0.04}$ &
  $\mathbf{ 95.91\pm 0.05}$ &
  $\mathbf{ 97.94\pm 0.02^*}$ &
  $\mathbf{ 88.99\pm 0.18}$
  &  $\mathbf{ 93.46\pm 0.10}$  \\ 
\bottomrule
\end{tabular}
}
\caption{Evaluation of different methods on three Multi-MNIST datasets. Rows with flat-based minimizers are shaded. Bold numbers denote higher accuracy between flat-based
methods and their baselines. $^*$ denotes the highest accuracy (except for STL as it unfairly exploits multiple neural networks). We also use arrows to indicate that the higher is the better $(\uparrow)$  or vice-versa $(\downarrow)$. \label{tab:mnist} }
\vspace*{-5mm}
\end{table*}

\textbf{Datasets and Baselines.} Our proposed method is evaluated on four MTL benchmarks, including Multi-MNIST \citep{lin2019pareto}, CelebA \citep{liu2015deep} for visual classification, and NYUv2 \citep{silberman2012indoor}, CityScapes \citep{cordts2016cityscapes} for scene understanding. Their descriptions can be found in  Appendix \ref{sec:implement}. 
We show how our framework boosts performance of gradient-based MTL methods by comparing \textit{vanilla} MGDA \citep{sener2018multi}, PCGrad \citep{yu2020gradient}, CAGrad \citep{liu2021conflict}, IMTL \citep{liu2021towards}, NashMTL \cite{navon2022multi}, FairGrad \cite{banfair} to their {flat-based} versions F-MGDA, F-PCGrad, F-CAGrad, F-IMTL, F-NashMTL and F-FairGrad. We also add a single-task learning (STL) baseline for each dataset.
\subsection{Image classification}
\vspace{-1mm}
\textbf{Multi-MNIST.} Following \cite{sener2018multi}, we set up three Multi-MNIST experiments with ResNet18 \citep{he2016deep}, namely: MultiFashion, MultiMNIST and MultiFashion+MNIST. 

As summarized in Table \ref{tab:mnist},  we can see that seeking flatter regions for all tasks can improve the performance of all the baselines across all three datasets. Especially, {flat-based} methods achieve the highest score for each task and for the average, outperforming STL by  $1.2\%$ on MultiFashion and MultiMNIST. We conjecture that the discrepancy between gradient update trajectories to classify digits from MNIST and fashion items from FashionMNIST has resulted in the fruitless performance of baselines, compared to STL on MultiFashion+MNIST. Even if there exists dissimilarity between tasks, our best obtained average accuracy when applying our method to CAGrad is just slightly lower than STL ($<0.4\%$) while employing a single model only.

\textbf{CelebA.} 
CelebA \citep{liu2018large} is a face dataset with 200K images and 40 attributes, forming a 40-class multi-label classification problem. Table \ref{tab:celeba} presents the average errors over 40 tasks, with Linear Scalarization (LS) and Uncertainty Weighting (UW) (Kendall, Gal, and Cipolla 2018) as additional baselines. The best results in each pair and overall are highlighted in bold and $^*$, respectively. Even with a large number of tasks, flat region seeking consistently shows its advantages, with F-CAGrad achieving the lowest average error. Notably, when the optimizer considers flat minima, the performance gaps between PCGrad, IMTL, and CAGrad (8.23, 8.24 vs. 8.22) are smaller than those under conventional ERM training (8.69, 8.88, and 8.52). This suggests that better aggregation of task gradients, and thus reduced conflict, occurs when shared parameters approach a common flat region.

\begin{table}[!ht]


\centering
\resizebox{0.95\columnwidth}{!}{
\begin{tabular}
{l|c|c|c|c|c|c|c}
\toprule
Method & STL & LS & UW & MGDA & PCGrad & CAGrad & IMTL    \\
\midrule
Vanilla & 8.77 & 9.99 & 9.66 & 9.96 & 8.69 & 8.52 & 8.88  \\
\midrule
\rowcolor{Gray} Flat-based   & - & - &- & \textbf{9.22} & \textbf{8.23} & \textbf{8.22*} & \textbf{8.24}  \\
 \bottomrule
\end{tabular}
}
\caption{Mean of error per category of MTL algorithms in multi-label classification on
CelebA dataset.\label{tab:celeba}} 
\vspace{-3mm}
\end{table}

\begin{table*}[!ht]
\setlength{\tabcolsep}{2pt}


\centering
\resizebox{0.8\textwidth}{!}{%
\begin{tabular}{lcc|cc|ccc|ccccccccc|cc}
\toprule
 &  &  & \multicolumn{2}{|c|}{Segmentation} &  & \multicolumn{2}{c|}{Depth} &  & \multicolumn{8}{c|}{Surface Normal}  &   & \\
 \cmidrule(lr){4-5} \cmidrule(lr){7-8} \cmidrule(lr){10-17}
 &  &  & \multirow{2}{*}{mIoU $\uparrow$} & \multirow{2}{*}{Pix Acc $\uparrow$} &  & \multirow{2}{*}{Abs Err $\downarrow$} & \multirow{2}{*}{Rel Err $\downarrow$} &  & \multicolumn{2}{c}{Angle Distance $\downarrow$} &  & \multicolumn{5}{c}{Within $t^\circ$  $\uparrow$} & $ \mathbf{\Delta m \%} \downarrow$ &  \\

 &  &  &  &  &  &  &  &  & Mean & Median &  & 11.25 &  & 22.5 &  & 30 &  &   \\
 \midrule
 & \multicolumn{2}{c|}{STL} & $38.30$ & $63.76$ &  & $0.6754$ & $0.2780$ &  & $25.01$ & $19.21$ &  & $30.14$ &  & $57.20$ &  & $69.15$ & 0.00 &   \\
  \midrule
 & \multicolumn{2}{c|}{LS} & $39.29$ & $65.33$ &  & $0.5493$ & $0.2263$ &  & $28.15$ & $23.96$ &  & $22.09$ &  & $47.50$ &  & $61.08$ &  $5.59$ &  \\
 & \multicolumn{2}{c|}{SI} & $38.45$ & $64.27$ && $0.5354$ & $0.2201$ && $27.60$ & $23.37$ && $22.53$ && $48.57$ && $62.32$ & $4.39$ & \\
 & \multicolumn{2}{c|}{RLW} & $ 37.17 $ & $ 63.77 $ && $ 0.5759 $ & $ 0.2410 $ && $ 28.27 $ & $ 24.18 $ && $ 22.26 $ && $ 47.05 $ && $ 60.62 $ &  $ 7.78 $ & \\
 & \multicolumn{2}{c|}{DWA} & $39.11$ & $65.31$ &  & $0.5510$ & $0.2285$ &  & $27.61$ & $23.18$ &  & $24.17$ &  & $50.18$ &  & $62.39$ &$3.57$ &  \\
 & \multicolumn{2}{c|}{UW} & $36.87$ & $63.17$ &  & $0.5446$ & $0.2260$ &  & $27.04$ & $22.61$ &  & $23.54$ &  & $49.05$ &  & $63.65$ & $4.05$ &  \\
  & \multicolumn{2}{c|}{GradDrop} & $39.39$ & $65.12$ &  & $0.5455$ & $0.2279$ &  & $27.48$ & $22.96$ &  & $23.38$ &  & $49.44$ &  & $62.87$ &$3.58$ &  \\
  & \multicolumn{2}{c|}{Nash-MTL} & $40.13$ & $65.93$ &  & $0.5261^*$ & $0.2171$ &  & $25.26$ & $20.08$ &  & $28.4$ &  & $55.47$ &  & $68.15$ & $-4.04$ &  \\
  \midrule
 & \multicolumn{2}{c|}{MGDA} & $\mathbf{30.47}$ & $\mathbf{59.90}$ &  & $\mathbf{0.6070}$ & $0.2555$ &  & ${24.88}$ & ${19.45}$ &  & ${29.18}$ &  & ${56.88}$ &  & ${69.36}$ &  $1.38$ &  \\
 \rowcolor{Gray}
  & \multicolumn{2}{c|} {F-MGDA} & $26.42$ & $ 58.78$ &  & $0.6078$ & $\mathbf{0.2353}$ &  & $\mathbf{24.34^*}$ & $\mathbf{18.45^*}$ &  & ~~$\mathbf{31.64^*}$ &  & $\mathbf{58.86^*}$ &  & $\mathbf{70.50^*}$ &  $ \mathbf{-0.33} $ &  \\
\midrule 
 & \multicolumn{2}{c|}{PCGrad} & $38.06$ & $64.64$ &  & $0.5550$ & $0.2325$ &  & $27.41$ & $\mathbf{22.80}$ &  & $\mathbf{23.86}$ &  & $\mathbf{49.83}$ &  & $\mathbf{63.14}$ &   $3.97$ &  \\
 \rowcolor{Gray}   &  \multicolumn{2}{c|}{F-PCGrad} & $\mathbf{40.05}$ & $ \mathbf{65.42}$ &  & $\mathbf{0.5429}$ & $\mathbf{0.2243}$ &  & $\mathbf{27.38}$ & $23.00$ &  & ~~$23.47$ &  & $49.35$ &  & $62.74$ & $ \mathbf{3.14} $ &  \\
\midrule   
 & \multicolumn{2}{c|}{CAGrad} & $39.79$ & $65.49$ &  & $0.5486$ & $0.2250$ &  & $26.31$ & $21.58$ &  & $25.61$ &  & $52.36$ &  & $65.58$ &  $0.20$ &  \\
 \rowcolor{Gray}    &  \multicolumn{2}{c|}{F-CAGrad} & $\mathbf{40.93^*}$ & $\mathbf{66.68^*}$ &  & $\mathbf{0.5285}$ & $\mathbf{0.2162}$ &  & $\mathbf{25.43}$ & $\mathbf{20.39}$ &  & ~~$\mathbf{27.99}$ &  & $\mathbf{54.82}$ &  & $\mathbf{67.56}$ &$\mathbf{ -3.78 }$ &  \\
\midrule
 & \multicolumn{2}{c|}{IMTL} & $39.35$ & $ 65.60$ &  & $0.5426$ & $0.2256$ &  & $26.02$ & $21.19$ &  & ~~$26.2$ &  & $53.13$ &  & $66.24$ &  $ -0.76 $ &  \\
  \rowcolor{Gray}   &  \multicolumn{2}{c|}{F-IMTL} & $\mathbf{40.42}$ & $ \mathbf{65.61}$ &  & $\mathbf{0.5389}$ & $\mathbf{0.2121^*}$ &  & $\mathbf{25.03}$ & $\mathbf{19.75}$ &  & ~~$\mathbf{28.90}$ &  & $\mathbf{56.19}$ &  & $\mathbf{68.72}$ &  $ \mathbf{-4.77^*} $ &  \\
 \bottomrule
\end{tabular}
}
\caption{ Test performance for three-task NYUv2 of Segnet \citep{badrinarayanan2017segnet}: semantic segmentation, depth estimation, and surface normal. Using the proposed procedure with gradient-based multi-task learning methods consistently improves their overall performance. \label{tab:nyu}}
\vspace{-4.5mm}
\end{table*}

\subsection{Scene Understanding}
Two datasets used here are NYUv2 \citep{silberman2012indoor} and CityScapes \citep{cordts2016cityscapes}. 
For these two experiments, we additionally include several recent MTL methods, namely, scale-invariant (SI), random
loss weighting (RLW), Dynamic Weight Average (DWA) \citep{liu2019end}, GradDrop \citep{chen2020just}, and Nash-MTL \citep{navon2022multi} whose results are taken from \cite{navon2022multi} and details are in Appendix \ref{sec:implement}. 
Also following the standard protocol used in \citep{liu2019end, liu2021conflict, navon2022multi}, Multi-Task Attention Network \citep{liu2019end} is employed on top of the SegNet architecture \citep{badrinarayanan2017segnet},  our results are averaged over the last $10$ epochs to align with previous work.

\textbf{Evaluation metric.} 
In this experiment, we handle different task types, each with its own metrics. We report the relative task improvement \citep{maninis2019attentive} to compare overall performance. 
Let $M_i$ and $S_i$ be the metrics obtained by the main and the single-task learning (STL) model, respectively. The relative task improvement on $i$-th task is mathematically given by:
$\Delta_i:=100 \cdot(-1)^{l_i}$ $({M_i-S_i})/{S_i}$, 
where $l_i = 1$ if a lower value for the $i$-th criterion is better and $0$ otherwise. We depict our results by the average relative task improvement $\boldsymbol{\Delta m\% }=\frac{1}{m} \sum_{i=1}^m \Delta_i$.

\textbf{NYUv2.} 
Table \ref{tab:nyu} presents the results and relative improvements of each task over STL for different methods. Generally, the flat-based versions achieve comparable or higher results on most metrics, except for F-MGDA in the segmentation task, where it notably decreases the $mIoU$ score. However, F-MGDA significantly boosts performance in other tasks, raising MGDA's overall relative improvement from -1.38\% to +0.33\% above STL. Notably, F-CAGrad and F-IMTL outperform competitors by large margins across all tasks, with top relative improvements of 3.78\% and 4.77\%, respectively.

\begin{table}[!ht]



\setlength{\tabcolsep}{3pt}
\centering
\resizebox{1.\columnwidth}{!}{%

\begin{tabular}{l|cc|cc|c}
        \toprule
         & \multicolumn{2}{c}{Segmentation} & \multicolumn{2}{|c|}{Depth} &        \\
         \midrule Method
         & mIoU $\uparrow$        & Pix Acc $\uparrow$        & Abs Err $\downarrow$    & Rel Err$\downarrow$   & $\boldsymbol{\Delta m\% \downarrow}$  \\
\midrule
STL    & $74.01$ & $93.16$ & $0.0125$ & $27.77$ &       \\
\midrule
LS     & $75.18$ & $93.49$ & $ 0.0155 $ & $46.77$ & $22.60$ \\
SI     & $70.95$ & $91.73$ & $0.0161$ & $33.83$ & $14.11$ \\
RLW    & $74.57$ & $93.41$ & $0.0158$ & $47.79$ & $24.38$ \\
DWA    & $75.24$ & $93.52$ & $0.0160$ & $44.37$ & 21.45 \\
UW     & $72.02$ & $92.85$ & $0.0140$ & $30.13^*$ & 5.89  \\
GradDrop & $75.27$         & $93.53$            & $0.0157$       & $47.54$      & $23.73$  \\
Nash-MTL & $75.41$         & $93.66$            & $0.0129$       & $35.02$      & $6.82$  \\
\midrule
MGDA   & $68.84$ & $91.54$ & $0.0309$ & $33.50$ & $44.14$ \\
\rowcolor{Gray} F-MGDA   & $\mathbf{73.77}$ & $\mathbf{93.12}$ & $\mathbf{0.0129}$ & $\mathbf{27.44^*}$ & $\mathbf{0.67^*}$ \\
\midrule
PCGrad & $75.13$ & $93.48$ & $0.0154$ & $42.07$ & $18.29$ \\
\rowcolor{Gray} F-PCGrad & $\mathbf{75.77}$ & $\mathbf{93.67}$ & $\mathbf{0.0144}$ & $\mathbf{39.60}$ & $\mathbf{13.65}$ \\
\midrule
CAGrad &  ${75.16}$ & $93.48$ & $0.0141$ & $37.60$ & $11.64$ \\
\rowcolor{Gray} F-CAGrad &$\mathbf{76.02}$ & $\mathbf{93.72}$ & $\mathbf{0.0134}$ & $\mathbf{34.64}$ & $\mathbf{7.25}$ \\
\midrule
IMTL                         & $75.33$     & $93.49$ & $0.0135$ & $38.41$ & $11.10$ \\
\rowcolor{Gray} F-IMTL                         & $\mathbf{76.63^*}$     & $\mathbf{93.76^*}$ & $\mathbf{0.0124^*}$ & $\mathbf{31.17}$ & $\mathbf{1.87}$ \\
\bottomrule
\end{tabular}%
}
\caption{Test performance for two-task CityScapes: semantic segmentation and depth estimation.$^*$ denotes the best score for each task's metrics. \label{tab:cityscape} 
}
\vspace*{-4mm}

\end{table}

\textbf{CityScapes.} 
In Table~\ref{tab:cityscape}, the positive impact of seeking flat regions is evident across all metrics and baselines. Notably, MGDA and IMTL show significant relative improvements, achieving the highest and second-best $\boldsymbol{\Delta m \%}$ scores, respectively. PCGrad, CAGrad, and IMTL even surpass STL in segmentation scores. Interestingly, MGDA biases to the depth estimation objective, leading to the predominant performance of F-MGDA on that task, consistent with patterns observed in \citep{liu2021towards} and the NYUv2 experiment.

\begin{figure*}[ht]
  \centering
  \includegraphics[width=2\columnwidth, trim=.cm 0cm 0.cm 0cm,clip]{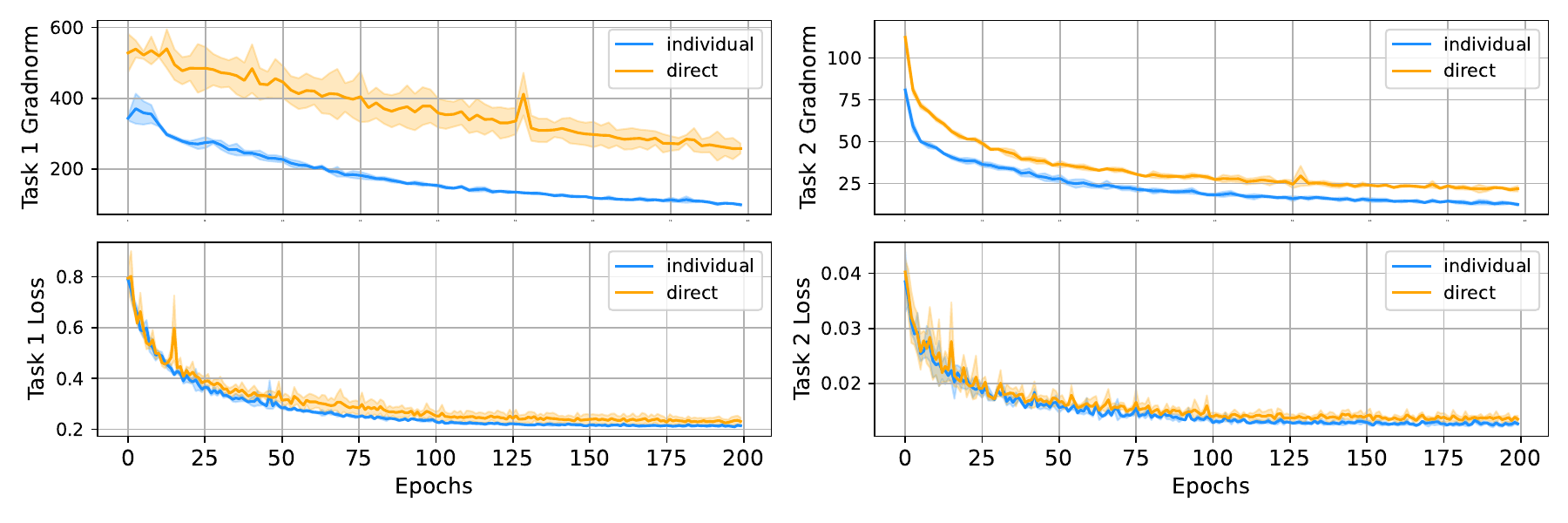}
  \centering
  \vspace*{-4mm}
   \caption{Evolution of gradient norms and task losses across aggregation strategies.}
   \label{fig:decomposition_gradnorm}
\vspace*{-5mm}
\end{figure*}

\subsection{Ablation study}
\label{sec:ablation}
Here, we provide experimental justification for our gradient decomposition, and our method's impact on conventional MTL training by examining task conflict. Appendix \ref{sec:supp_res} provides additional results for model calibration \ref{subsec:calibration}, model robustness \ref{subsec:model_robustness}, \ref{subsec:model_sharpness_supp}, gradient norms \ref{subsec:grad_norm_supp}, loss landscape visualization \ref{subsec:loss_viz}, training curves \ref{subsec:training_curves}, and hyper-param sensitivity \ref{subsec:rho_ablation}.

\textbf{Directly aggregating SAM gradients neglects intra-conflict.} Table \ref{tab:aggregation} compares between the direct aggregation on $\{\boldsymbol{g}_{s h}^{i, \mathrm{SAM}}\}_{i=1}^m$ and our individual aggregation on $\{\boldsymbol{g}_{sh}^{i, \mathrm{flat}}\}_{i=1}^m$, and $\{\boldsymbol{g}_{sh}^{i, \mathrm{loss}}\}_{i=1}^m$ .

\begin{table}[!ht]

\setlength{\tabcolsep}{5pt}
\centering
\resizebox{0.95\columnwidth}{!}{%
\begin{tabular}{l|cc|cc|c}
        \toprule
         & \multicolumn{2}{c}{Segmentation} & \multicolumn{2}{|c|}{Depth} &        \\
         \midrule Method
         & mIoU $\uparrow$        & Pix Acc $\uparrow$        & Abs Err $\downarrow$    & Rel Err$\downarrow$   & $\boldsymbol{\Delta m\% \downarrow}$  \\
\midrule
ERM   & $68.84$ & $91.54$ & $0.0309$ & $33.50$ & $44.14$ \\
 Ours (\textbf{direct})  & ${68.93}$ & ${ 91.41}$ & ${0.0130}$ & ${31.37}$ & ${6.43}$ \\
 \midrule
\rowcolor{Gray} Ours (\textbf{individual})   & $\mathbf{73.77}$ & $\mathbf{93.12}$ & $\mathbf{0.0129}$ & $\mathbf{27.44^*}$ & $\mathbf{0.67}$  \\
\bottomrule
\end{tabular}%
}
\caption{
Direct SAM gradients aggregation vs our proposed gradient aggregation strategy on CityScapes.  \label{tab:aggregation}}
\end{table}
\newpage
Compared to the naive approach, in which per-task SAM gradients are directly aggregated, our decomposition approach consistently improves performance by a large margin across all tasks. Moreover, using our decomposed SAM yields flatter minima and lower loss values throughout the whole training process, as shown in Figure \ref{fig:decomposition_gradnorm}. These results reinforce the rationale behind separately aggregating low-loss directions and flat directions.

\textbf{Other approaches to apply gradient aggregation.} Table \ref{tab:mnist-abl} summarizes the performance of Second-aggre, Each-aggre and Our aggregation strategies: Second-aggre performs an additional step to aggregate $\boldsymbol{g}_{sh}^{loss}$ and $\boldsymbol{g}_{sh}^{flat}$; Each-aggre iteratively aggregates $\boldsymbol{g}_{sh}^{i,loss}$ and $\boldsymbol{g}_{sh}^{i,flat}$ to obtain $\boldsymbol{g}_{sh}^i$ for each task, then aggregate $\boldsymbol{g}_{sh}^i$ one more time. Our strategy requires less aggregation steps yet still achieves comparable performance.
\begin{table}[!ht]
\setlength{\tabcolsep}{2pt}
\centering
\resizebox{\columnwidth}{!}{%
\begin{tabular}{c|ccc|ccc|ccc}
\hline
 &
  \multicolumn{3}{c}{MultiFashion} &
  \multicolumn{3}{|c|}{MultiMNIST} &
  \multicolumn{3}{c}{MultiFashion+MNIST} \\  \cmidrule(lr){2-4} \cmidrule(lr){5-7} \cmidrule(lr){8-10}
  
\multirow{-2}{*}{Method} &
  Task 1 $\uparrow$&
  Task 2 $\uparrow$&
  Avg $\uparrow$&
  Task 1 $\uparrow$&
  Task 2 $\uparrow$&
  Avg $\uparrow$&
  Task 1 $\uparrow$&
  Task 2 $\uparrow$&
  Avg $\uparrow$\\
  
\midrule
 Second-aggre &
  $88.10$ &
  $87.70 $ &
  $87.90$ &
  $96.48 $ &
  $95.24 $ &
  $95.86$ &
  $97.87$ &
  $89.02$ &
  $93.44$ \\ 
\midrule
Each-aggre &
  $87.51 $ &
  $87.68 $ &
  $87.59 $ &
  $96.45 $ &
  $95.40 $ &
  $95.92 $ &
  $97.61 $ &
  $89.17 $ &
  $93.39 $ \\
  \midrule
Ours &
  $88.19$ &
  $87.45$ &
  $87.82 $ &
  $96.54 $ &
  $95.36 $ &
  $95.95 $ &
  $97.82 $ &
  $89.26 $ & 
  $93.54 $ \\ 
\bottomrule
\end{tabular}
}
\caption{Different aggregation strategies applied on CAGrad, on Multi-MNIST datasets.  \label{tab:mnist-abl}}
\vspace*{-1mm}
\end{table}

\textbf{Our improvement is not a mere result of single-task SAM.} To show this point, we provide the results of STL and linear scalarization (LS), casting MTL as a single objective, trained with and without SAM on NYUv2 in Table \ref{tab:reb_nyu}. When equipped with SAM, F-STL and F-LS improve almost all scores of their counterparts.  However, they can not consistently exceed flat gradient-based MTL baselines, which take gradient conflict into account. Particularly, on Segmentation and Depth tasks, F-IMTL achieves the highest scores, while on the Surface Normal task, F-MGDA is the best method. Overall, F-IMTL obtains the best $\boldsymbol{\Delta m\%}$. 
\begin{table}[!ht]
\setlength{\tabcolsep}{2pt}
\centering
\resizebox{1\columnwidth}{!}{%
\begin{tabular}{lcc|cc|ccc|ccccccccc|cc}
\toprule
 &  &  & \multicolumn{2}{|c|}{Segmentation} &  & \multicolumn{2}{c|}{Depth} &  & \multicolumn{8}{c|}{Surface Normal}  &   & \\
 \cmidrule(lr){4-5} \cmidrule(lr){7-8} \cmidrule(lr){10-17}
 &  &  & \multirow{2}{*}{mIoU $\uparrow$} & \multirow{2}{*}{Pix Acc $\uparrow$} &  & \multirow{2}{*}{Abs Err $\downarrow$} & \multirow{2}{*}{Rel Err $\downarrow$} &  & \multicolumn{2}{c}{Angle Distance $\downarrow$} &  & \multicolumn{5}{c}{Within $t^\circ$  $\uparrow$} & $ {\Delta m \%} \downarrow$ &  \\

 &  &  &  &  &  &  &  &  & Mean & Median &  & 11.25 &  & 22.5 &  & 30 &  &   \\
 \midrule
 & \multicolumn{2}{c|}{STL} & $38.30$ & $63.76$ &  & $0.6754$ & $0.2780$ &  & $25.01$ & $19.21$ &  & $30.14$ &  & $57.20$ &  & $69.15$ & 0.00 &   \\
  
\rowcolor{Gray}  & \multicolumn{2}{c|}{F-STL} & $39.07$ & $64.21$ &  & $0.6183$ & $0.2514$ &  & $25.08$ & $18.72$ &  & $30.99$ &  & $58.29$ &  & $69.76$ &  $-3.17$ &  \\
  \midrule
 & \multicolumn{2}{c|}{LS} & $39.29$ & $65.33$ &  & $0.5493$ & $0.2263$ &  & $28.15$ & $23.96$ &  & $22.09$ &  & $47.50$ &  & $61.08$ &  $5.59$ &  \\
 \rowcolor{Gray}  & \multicolumn{2}{c|}{F-LS} & $40.28$ & $65.30$ &  & $0.5360$ & $0.2173$ &  & $27.14$ & $22.56$ &  & $24.45$ &  & $50.29$ &  & $63.57$ &  $1.65$ &  \\
 \midrule
  & \multicolumn{2}{c|} {F-MGDA} & $26.42$ & $ 58.78$ &  & $0.6078$ & ${0.2353}$ &  & ${24.34}$ & ${18.45}$ &  & ~~${31.64}$ &  & ${58.86}$ &  & ${70.50}$ &  $ {-0.33} $ &  \\
  
   &  \multicolumn{2}{c|}{F-IMTL} & ${40.42}$ & $ {65.61}$ &  & ${0.5389}$ & ${0.2121}$ &  & ${25.03}$ & ${19.75}$ &  & ~~${28.90}$ &  & ${56.19}$ &  & ${68.72}$ &  $ {-4.77} $ &  \\
 \bottomrule
\end{tabular}
}
    \centering
    \caption{ Test performance for three-task \textbf{NYUv2} of Segnet.  \label{tab:reb_nyu}}
    \vspace*{-2mm}
\end{table}

\begin{figure}[!ht]
    \centering
     \includegraphics[width=.49\textwidth, trim=.0cm 0cm .0cm 0cm,clip]{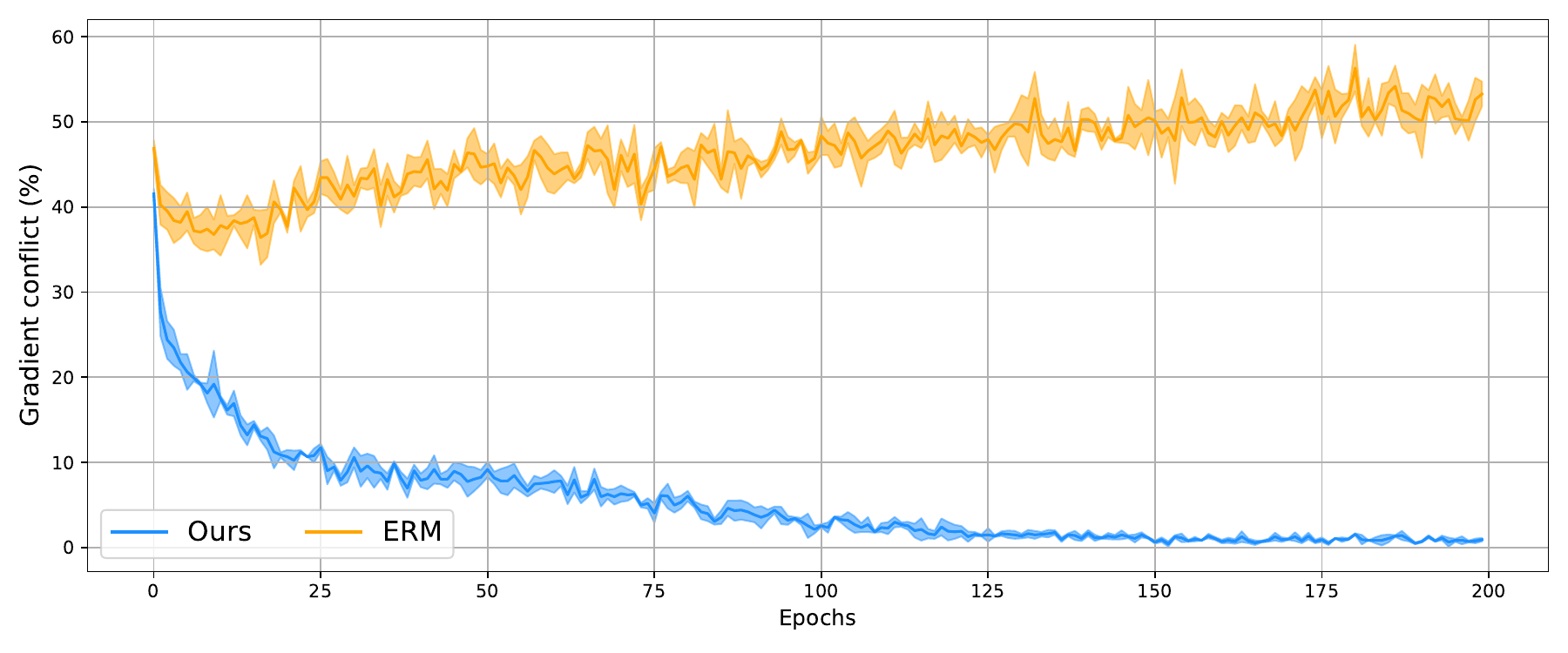}
    \caption{Proportion of conflict between per-task gradients ($\boldsymbol{g}^{1,\text{loss}} \cdot \boldsymbol{g}^{2,\text{loss}} <0$) on Multi-MNIST.}
    \label{fig:conflict}
\end{figure}

\textbf{Task conflict.} To empirically confirm reduced gradient conflict in flat regions, we measured the proportion of minibatches with gradient conflict at each epoch, and present the results in Figure \ref{fig:conflict}. While ERM's gradient conflict rises above 50\%, ours decreases and approaches 0\%. This reduction is also a key objective of recent gradient-based MTL methods aimed at mitigating negative transfer between tasks 
\citep{yu2020gradient, zhugradient, wang2020gradient}. \label{sec:task_conflict}

%% file: sec/6_conclusion.tex
\section{Conclusion}
\label{sec:conclusion}


In this work, we have presented a general framework that can be incorporated into current multi-task learning methods following the gradient balancing mechanism. The core ideas of our proposed method are the employment of flat minimizers in the context of MTL and proving that they can help enhance previous works both theoretically and empirically. Concretely, our method goes beyond optimizing per-task objectives solely to yield models that have both low errors and high generalization capabilities. 


%% file: sec/X_suppl.tex
\clearpage
\setcounter{page}{1}
\maketitlesupplementary
\appendix

%


Due to space constraints, some details were omitted from the main paper. We therefore include additional theoretical developments (section \ref{sec:theory}) and experimental results (section \ref{sec:implement}) in this appendix.

\section{Our Theory Development \label{sec:theory}}

This section contains the proofs and derivations of our theory development to support the main submission.


We first start with the following theorem, which is inspired by the general PAC-Bayes in ~\citep{JMLR:v17:15-290}.
\begin{thm}\label{general_pac} With the assumption that adding Gaussian perturbation will raise the test error: $\mathcal{L}_{\mathcal{D}}(\boldsymbol{\theta}) \leq \mathbb{E}_{\epsilon \sim \mathcal{N}(0, \sigma^2 \mathbb{I})}\left[\mathcal{L}_{\mathcal{D}}(\boldsymbol{\theta}+\boldsymbol{\epsilon})\right]$. Let $T$ be the number of parameter $\boldsymbol{\theta}$, and $N$ be the cardinality of $\mathcal{S}$, then the following inequality is true with the probability $1-\delta$:
\begin{align*} 
&\mathcal{L}_{\mathcal{D}}\left(\boldsymbol{\theta}\right)  \leq \mathbb{E}_{\epsilon \sim \mathcal{N}(0, \sigma^2\mathbb{I})}\left[\mathcal{L}_{\mathcal{S}}  (\boldsymbol{\theta}+\boldsymbol{\epsilon})\right] 
+ \frac{1}{\sqrt{N}}\Bigg[\frac{1}{2} + \\ 
&\frac{T}{2} \log{\Big(1+\frac{||\boldsymbol{\theta}||^2}{T\sigma^2} \Big)} + \log \frac{1}{\delta} + 6\log (N+T) + \frac{L^2}{8}  \Bigg], 
\end{align*} 
where $L$ is the upper-bound of the loss function.
\label{thm:base_thm}
\end{thm}

\begin{proof}
   
We use the PAC-Bayes theory for $P= \mathcal{N}(\mathbf{0},\sigma_P^2\mathbb{I}_T)$ and $Q = \mathcal{N}(\boldsymbol{\theta},\sigma^2 \mathbb{I}_T)$ are the prior and posterior distributions, respectively.

By using the bound in \citep{JMLR:v17:15-290}, with probability at least $1-\delta$ and for all $\beta > 0$, we have: 
    \begin{align*}
        \mathbb{E}_{\boldsymbol{\theta}\sim Q} \left[\mathcal{L}_{\mathcal{D}}(\boldsymbol{\theta})\right]
        &\leq \mathbb{E}_{\boldsymbol{\theta}\sim Q}\left[\mathcal{L}_{\mathcal{S}}(\boldsymbol{\theta})\right] \\
        &+ \frac{1}{\beta}\Big[ \mathsf{KL}(Q\|P) + \log \frac{1}{\delta} + \Psi(\beta,N)\Big],
    \end{align*}
    
    where we have defined:
    \begin{align*}
        \Psi(\beta,N) = \log \mathbb{E}_{P}\mathbb{E}_{\mathcal{S}}\Big[ \exp\Big\{\beta \big(\mathcal{L}_{\boldsymbol{D}}(\boldsymbol{\theta}) - \mathcal{L}_{\mathcal{S}}(\boldsymbol{\theta}) \big) \Big\}\Big].
    \end{align*}

Note that the loss function is bounded by $L$, according to Hoeffding's lemma, we have:
    \begin{align*}
        \Psi(\beta,N) \leq \frac{\beta^2 L^2}{8N}.
    \end{align*}

By Cauchy inequality:
\begin{align*}
    &\quad\frac{1}{\sqrt{N}}\Bigg[ \frac{T}{2}\log\Big(1+\frac{||{\boldsymbol{\theta}}||^2}{T\sigma^2} \Big) + \frac{L^2}{8} \Bigg] \\
    &\geq \frac{L}{2\sqrt{N}}\sqrt{T\log\Big(1+\frac{||{\boldsymbol{\theta}}||^2}{T\sigma^2} \Big)} 
    \geq L, 
\end{align*}
which means that the theorem is proved since the loss function is upper bounded by $L$, following assumptions, if $||\boldsymbol{\theta}||^2 \geq T\sigma^2 \Big[\exp\frac{4N}{T}-1\Big]$.

Now, we only need to prove the theorem under the case: $||\boldsymbol{\theta}||^2 \leq T\sigma^2 \Big[\exp\frac{4N}{T}-1\Big]$.

We need to specify $P$ in advance since it is a prior distribution. However, we do not know in advance the value of $\boldsymbol{\theta}$ that affects the KL divergence term. Hence, we build a family of distribution $P$ as follows:
    \begin{align*}
        \mathfrak{P} = \Big\{P_j &= \mathcal{N}(\mathbf{0},\sigma_{P_j}^2\mathbb{I}_T): 
        \sigma_{P_j}^2 = c \exp\big(\frac{1-j}{T}\big), \\
        c &= \sigma^2 \big(1 + \exp\frac{4N}{T} \big), j = 1,2,\ldots\Big\}.
    \end{align*}
Set $\delta_j = \frac{6\delta}{\pi^2j^2}$, the below inequality holds with probability at least $1-\delta_j$:
 \begin{align*}
       \mathbb{E}_{\boldsymbol{\theta}\sim Q} \left[ \mathcal{L}_{\mathcal{D}}(\boldsymbol{\theta})\right] 
       &\leq \mathbb{E}_{\boldsymbol{\theta}\sim Q} \left[\mathcal{L}_{\mathcal{S}}(\boldsymbol{\theta})\right] \\
       &+ \frac{1}{\beta}\Big[ \mathsf{KL}(Q\|P_j) + \log \frac{1}{\delta_j} + \frac{\beta^2L^2}{8N}\Big].
    \end{align*}
Or it can be written as:
 \begin{align*}
       \mathbb{E}_{\epsilon \sim \mathcal{N}(0, \sigma^2\mathbb{I})} \left[ \mathcal{L}_{\mathcal{D}}(\boldsymbol{\theta}+\boldsymbol{\epsilon})\right]   &\leq \mathbb{E}_{\epsilon \sim \mathcal{N}(0, \sigma^2\mathbb{I})} \left[\mathcal{L}_{\mathcal{S}}(\boldsymbol{\theta}+\boldsymbol{\epsilon})\right] \\
       &+ \frac{1}{\beta}\Big[ \mathsf{KL}(Q\|P_j) + \log \frac{1}{\delta_j} + \frac{\beta^2L^2}{8N}\Big].
    \end{align*}
Thus, with probability $1-\delta$ the above inequalities hold for all $P_j$.  We choose:
\begin{align*}
    j^* =  \left \lfloor 1 + T\log\left(\frac{\sigma^2\big(1+\exp\{ 4N/T\} \big)}{\sigma^2 +\|\boldsymbol{\theta}\|^2/T}\right) \right\rfloor. 
\end{align*}
Since $\frac{\|\boldsymbol{\theta}\|^2}{T} \leq \sigma^2 \big[\exp\frac{4N}{T} -1 \big]$, we get $\sigma^2 + \frac{\|\boldsymbol{\theta}\|^2}{T} \leq \sigma^2 \exp\frac{4N}{T} $, thus $j^*$ is well-defined.  We also have:

\begin{align*}
   &\hspace{8mm}T \log\frac{c}{\sigma^2+ \|\boldsymbol{\theta}\|^2/T } \leq j^* \leq 1 + T\log \frac{c}{\sigma^2 + \|\boldsymbol{\theta}\|^2/T} \\
   &\Rightarrow\quad  \log\frac{c}{\sigma^2+ \|\boldsymbol{\theta}\|^2/T }\leq \frac{j^*}{T} \leq \frac{1}{T} + \log \frac{c}{\sigma^2 + \|\boldsymbol{\theta}\|^2/T}\\
   &\Rightarrow \quad  -\frac{1}{T} + \log \frac{\sigma^2 + \|\boldsymbol{\theta}\|^2/T}{c} \leq \frac{-j^*}{T} \leq \log\frac{\sigma^2 + \|\boldsymbol{\theta}\|^2/T}{c} \\
   &\Rightarrow \quad  e^{-1/T} \frac{\sigma^2+ \|\boldsymbol{\theta}\|^2/T}{c} \leq e^{-j^*/T} \leq \frac{\sigma^2 + \|\boldsymbol{\theta}\|^2/T}{c} \\
   &\Rightarrow \quad \sigma^2 + \frac{\|\boldsymbol{\theta}\|^2}{T} \leq c e^{\frac{1-j^*}{T}} \leq e^{\frac{1}{T}}\Big(\sigma^2 + \frac{\|\boldsymbol{\theta}\|^2}{T} \Big)\\
   &\Rightarrow \quad   \sigma^2 + \frac{\|\boldsymbol{\theta}\|^2}{T} \leq \sigma_{P_{j^*}}^2 \leq e^{\frac{1}{T}}\Big(\sigma^2 + \frac{\|\boldsymbol{\theta}\|^2}{T}\Big).
\end{align*}

 Hence, we have:
\begin{align*}
    &\mathsf{KL}(Q\|P_{j^*}) =\frac{1}{2}\Big[\frac{T\sigma^2 +\|\boldsymbol{\theta}\|^2}{\sigma_{P_{j^*}}^2} - T + T\log\frac{\sigma_{P_{}j^*}^2}{\sigma^2} \Big]  \\ &\leq \frac{1}{2}\Big[\frac{T\sigma^2 +\|\boldsymbol{\theta}\|^2}{\sigma^2 + \|\boldsymbol{\theta}\|^2/T} - T + T\log\frac{e^{1/T}\big(\sigma^2 + \|\boldsymbol{\theta}\|^2/T \big)}{\sigma^2}  \Big] \\
    &\leq  \frac{1}{2}\Big[1+ T\log\big(1 + \frac{\|\boldsymbol{\theta}\|^2}{T\sigma^2}\big) \Big].
\end{align*}

For the term $\log\frac{1}{\delta_{j^*}}$, use the inequality $\log(1+e^t) \leq 1 + t$ for $t>0$:
\begin{align*}
    &\log\frac{1}{\delta_{j^*}} = \log \frac{(j^*)^2\pi^2}{6\delta} = \log\frac{1}{\delta}  + \log\Big(\frac{\pi^2}{6}\Big) + 2\log(j^*) \\
    &\leq \log\frac{1}{\delta} + \log\frac{\pi^2}{6} + 2\log \Big( 1+T\log\frac{\sigma^2\big(1+ \exp(4N/T)\big)}{\sigma^2 + \|\boldsymbol{\theta}\|^2/T}\Big)  \\
    &\leq \log\frac{1}{\delta} + \log\frac{\pi^2}{6} + 2\log\Big(1+ T\log\big(1+\exp(4N/T)\big)\Big) \\
    &\leq \log\frac{1}{\delta} + \log\frac{\pi^2}{6} + 2\log\Big(1+ T\big(1+\frac{4N}{T} \big) \Big) \\
    &\leq \log\frac{1}{\delta} + \log\frac{\pi^2}{6} + \log(1+T + 4N).
\end{align*}

Choosing $\beta = \sqrt{N}$, with probability at least $1-\delta$ we get:
\begin{align*}
    &\frac{1}{\beta} \Big[\mathsf{KL}(Q\|P_{j^*}) + \log \frac{1}{\delta_{j^*}} + \frac{\beta^2 L^2}{8N} \Big]  \\
    &\leq \frac{1}{\sqrt{N}}\Big[\frac{1}{2}+\frac{T}{2}\log\Big(1+\frac{\|\boldsymbol{\theta}\|^2}{T\sigma^2}\Big) + \log \frac{1}{\delta} + 6\log(N+T)\Big] \\
    &+ \frac{L^2}{8\sqrt{N}}.
\end{align*}

Thus the theorem is proved.
\end{proof}

 Back to our context of multi-task learning in which we have $m$ tasks with each task model: $\boldsymbol{\theta}^i = [\boldsymbol{\theta}_{sh}, \boldsymbol{\theta}^i_{ns}]$, we can prove the following theorem. 

\begin{thm} \label{thm:sup_main_theo}
 With the assumption that adding Gaussian perturbation will rise the test error: $\mathcal{L}_{\mathcal{D}}(\boldsymbol{\theta}^i) \leq \mathbb{E}_{\epsilon \sim \mathcal{N}(0, \sigma^2 \mathbb{I})}\left[\mathcal{L}_{\mathcal{D}}(\boldsymbol{\theta}^i+\boldsymbol{\epsilon})\right]$. Let $T_i$ be the number of parameter $\boldsymbol{\theta^i}$ and $N$ be the cardinality of $\mathcal{S}$.
 We have the following inequality holds with probability $1-\delta$ (over the choice of training set $\mathcal{S} \sim \mathcal{D}$):
\begin{equation}
\left[\mathcal{L}_{\mathcal{D}}^{i}\left(\boldsymbol{\theta}^{i}\right)\right]_{i=1}^{m}\leq \big[\mathbb{E}_{\epsilon \sim \mathcal{N}(0, \sigma^2\mathbb{I})}\left[\mathcal{L}_{\mathcal{S}}(\boldsymbol{\theta}^i+\boldsymbol{\epsilon})\right]+f^{i}\left(\Vert\boldsymbol{\theta}^{i}\Vert_{2}^{2}\right)\big]_{i=1}^{m},
\end{equation} 
where
\begin{align*}
    f^{i}\left(\Vert\boldsymbol{\theta}^{i}\Vert_{2}^{2}\right) &= \frac{1}{\sqrt{N}}\Bigg[\frac{1}{2} + \frac{T_i}{2} \log\Big(1+\frac{||{\boldsymbol{\theta}}||^2}{T_i\sigma^2} \Big) \\
    &+ \log\frac{1}{\delta}
    + 6\log(N+T_i) + \frac{L^2}{8}  \Bigg].
\end{align*}

\end{thm}

\textbf{Proof.} The result for the base case $m=1$ can be achieved by using Theorem \ref{thm:base_thm} where $\xi = \delta$ and $f^1$ is defined accordingly. 
 We proceed by induction, suppose that Theorem \ref{thm:sup_main_theo} is true for all  $i \in [n]$ with probability $1 - \delta/2$, which also means:
  \begin{equation*}
\left[\mathcal{L}_{\mathcal{D}}^{i}\left(\boldsymbol{\theta}^{i}\right)\right]_{i=1}^{n}\leq \big[\mathbb{E}_{\epsilon \sim \mathcal{N}(0, \sigma\mathbb{I})}\left[\mathcal{L}_{\mathcal{S}}(\boldsymbol{\theta}^i+\boldsymbol{\epsilon})\right]+f^{i}\left(\Vert\boldsymbol{\theta}^{i}\Vert_{2}^{2}\right)\big]_{i=1}^{n}.
\end{equation*} 

Using Theorem \ref{thm:base_thm} for $\boldsymbol{\theta}^{n+1}$ and $\xi = \delta/2$, with probability $1 - \delta/2$, we have:
\begin{align*}
    \mathcal{L}_{\mathcal{D}}^{n+1}\left(\boldsymbol{\theta}^{n+1}\right) \leq &\mathbb{E}_{\epsilon \sim \mathcal{N}(0, \sigma\mathbb{I})}\left[\mathcal{L}_{\mathcal{S}}(\boldsymbol{\theta}^{n+1}+\boldsymbol{\epsilon})\right]\\
    &+f^{n+1}\left(\Vert\boldsymbol{\theta}^{n+1}\Vert_{2}^{2}\right).
\end{align*} 

  Using the inclusion–exclusion principle, with probability at least $1 - \delta$, we reach the conclusion for $m = n+1$.

We next prove the result in the main paper. Let us begin by formally restating the main theorem as follows:

\begin{thm}
For any perturbation radius $\rho_{sh}, \rho_{ns} > 0$, with probability $1-\delta$ (over the choice of training set $\mathcal{S} \sim \mathcal{D}$) we obtain:
\begin{align}
&\left[\mathcal{L}_{\mathcal{D}}^{i}\left(\boldsymbol{\theta}^{i}\right)\right]_{i=1}^{m}\leq \\
&\max_{\Vert\boldsymbol{\epsilon}_{sh}\Vert_{2}\leq \rho_{sh}}\Biggl[
\max_{\Vert\boldsymbol{\epsilon}_{ns}^{i}\Vert_2\leq\rho_{ns}}\mathcal{L}_{\mathcal{S}}^{i}\left(\boldsymbol{\theta}_{sh}+\boldsymbol{\epsilon}_{sh},\boldsymbol{\theta}_{ns}^{i}+\boldsymbol{\epsilon}_{ns}^{i}\right) \\
&\quad\quad\quad+f^{i}\left(\Vert\boldsymbol{\theta}^{i}\Vert_{2}^{2}\right)\Biggr]_{i=1}^{m},
\end{align} 

where $f^{i}\left(\Vert\boldsymbol{\theta}^{i}\Vert_{2}^{2}\right)$ is defined the same as in Theorem \ref{thm:sup_main_theo}.
\end{thm}

\textbf{Proof.} Theorem \ref{thm:sup_main_theo} gives us
\begin{align*}
&\left[\mathcal{L}_{\mathcal{D}}^{i}\left(\boldsymbol{\theta}^{i}\right)\right]_{i=1}^{m} \\
&\leq\left[\mathbb{E}_{\boldsymbol{\epsilon}\sim N(0,\sigma^2\mathbb{I})}\left[\mathcal{L}_{\mathcal{S}}^{i}\left(\boldsymbol{\theta}^{i}+\boldsymbol{\epsilon}\right)\right]+f^{i}\left(\Vert\boldsymbol{\theta}^{i}\Vert_{2}\right)\right]_{i=1}^{m}\\
&=\biggl[\int\mathbb{E}_{\boldsymbol{\epsilon}_{ns}^{i}}\left[\mathcal{L}_{\mathcal{S}}^{i}\left(\boldsymbol{\theta}_{sh}+\boldsymbol{\epsilon}_{sh},\boldsymbol{\theta}_{ns}^{i}+\boldsymbol{\epsilon}_{ns}^{i}\right)\right]p\left(\boldsymbol{\epsilon}_{sh}\right)d\boldsymbol{\epsilon}_{sh} \\
&\qquad+f^{i}\left(\Vert\boldsymbol{\theta}^{i}\Vert_{2}\right)\biggr]_{i=1}^{m} \\
&=\mathbb{E}_{\boldsymbol{\epsilon}_{sh}}\left[\mathbb{E}_{\boldsymbol{\epsilon}_{ns}^{i}}\left[\mathcal{L}_{\mathcal{S}}^{i}\left(\boldsymbol{\theta}_{sh}+\boldsymbol{\epsilon}_{sh},\boldsymbol{\theta}_{ns}^{i}+\boldsymbol{\epsilon}_{ns}^{i}\right)\right]+f^{i}\left(\Vert\boldsymbol{\theta}^{i}\Vert_{2}\right)\right]_{i=1}^{m},
\end{align*}

where $p(\boldsymbol{\epsilon}_{sh})$ is the density function of Gaussian distribution; $\boldsymbol{\epsilon}_{sh}$ and $\boldsymbol{\epsilon}^i_{ns}$ are drawn from their corresponding Gaussian distributions. 

We have $\boldsymbol{\epsilon}_{ns}^{i}\sim N(0,\sigma^2\mathbb{I}_{ns})$ with the dimension $T_{i,ns}$, therefore $\Vert\boldsymbol{\epsilon}_{ns}^{i}\Vert$ follows the Chi-square distribution. As proven in \citep{laurent2000adaptive}, we have for all $i$:
 \begin{equation*}
     P\left(\|\boldsymbol{\epsilon}^i_{ns}\|_2^2 \geq T_{i,ns} \sigma^2 + 2 \sigma^2 \sqrt{T_{i,ns} t}+2 t \sigma^2\right) \leq e^{-t}, \forall t>0
 \end{equation*}
 
 \begin{equation*}
     P\left(\|\boldsymbol{\epsilon}^i_{ns}\|_2^2 < T_{i,ns} \sigma^2 + 2 \sigma^2 \sqrt{T_{i,ns} t}+2 t \sigma^2\right) > 1 - e^{-t}
 \end{equation*}
 for all $t>0$.    
 
 Select $t=\ln(\sqrt{N})$, we derive the following bound for the noise magnitude in terms of the perturbation radius $\rho_{ns}$ for all $i$:
   \begin{align}
    &P\left(\|\boldsymbol{\epsilon}^i_{ns}\|_2^2 \leq \sigma^2(2 \ln (\sqrt{N})+T_{i,ns} + 2 \sqrt{T_{i,ns} \ln (\sqrt{N})})\right) \nonumber \\
    &\qquad> 1 - \frac{1}{\sqrt{N}}. \label{eq:ns}
 \end{align}
 Moreover, we have $\boldsymbol{\epsilon}_{sh}\sim N(0,\sigma^2\mathbb{I}_{sh})$ with the dimension $T_{sh}$, therefore $\Vert\boldsymbol{\epsilon}_{sh}\Vert$ follows the Chi-square distribution. As proven in \citep{laurent2000adaptive}, we have:
 \begin{align*}
     P\left(\|\boldsymbol{\epsilon}_{sh}\|_2^2 \geq T_{sh} \sigma^2 + 2 \sigma^2 \sqrt{T_{sh} t}+2 t \sigma^2\right) \leq e^{-t}, \forall t>0\\
     P\left(\|\boldsymbol{\epsilon}_{sh}\|_2^2 < T_{sh} \sigma^2 + 2 \sigma^2 \sqrt{T_{sh} t}+2 t \sigma^2\right) > 1 - e^{-t}
 \end{align*}
 for all $t>0$.
 
 Select $t=\ln(\sqrt{N})$, we derive the following bound for the noise magnitude in terms of the perturbation radius $\rho_{sh}$:
   \begin{align}
    &P\left(\|\boldsymbol{\epsilon}_{sh}\|_2^2 \leq \sigma^2(2 \ln (\sqrt{N})+T_{sh}+2 \sqrt{T_{sh} \ln (\sqrt{N})})\right) \nonumber \\ 
    &\qquad> 1 - \frac{1}{\sqrt{N}}. \label{eq:sh}
 \end{align}

 By choosing $\sigma$ less than $\frac{\rho_{sh}}{\sqrt{2\ln N^{1/2}+T_{sh}+2\sqrt{T_{sh}\ln N^{1/2}}}}$ and $\min_{i} \frac{\rho_{ns}}{\sqrt{2\ln N^{1/2}+T_{i,ns}+2\sqrt{T_{i,ns}\ln N^{1/2}}}}$,
 and referring to (\ref{eq:ns},\ref{eq:sh}), we achieve both:
 \begin{align*}
P\left(\Vert\boldsymbol{\epsilon}_{ns}^{i}\Vert<\rho_{ns}\right) & >1-\frac{1}{N^{1/2}}, \forall i,\\
P\left(\Vert\boldsymbol{\epsilon}_{sh}\Vert<\rho_{sh}\right) & >1-\frac{1}{N^{1/2}}.
\end{align*}

Finally, we finish the proof as:
\begin{align*}
&\left[\mathcal{L}_{\mathcal{D}}^{i}\left(\boldsymbol{\theta}^{i}\right)\right]_{i=1}^{m} \\
 &\leq  \mathbb{E}_{\boldsymbol{\epsilon}_{sh}}\left[\mathbb{E}_{{\boldsymbol{\epsilon}_{ns}^{i}}}\left[\mathcal{L}_{\mathcal{S}}^{i}\left(\boldsymbol{\theta}_{sh}+\boldsymbol{\epsilon}_{sh},\boldsymbol{\theta}_{ns}^{i}+\boldsymbol{\epsilon}_{ns}^{i}\right)\right]+f^{i}\left(\Vert\boldsymbol{\theta}^{i}\Vert_{2}\right)\right]_{i=1}^{m} \\
 &\leq 
 \mathbb{\max}_{||\boldsymbol{\epsilon}_{sh}||<\rho_{sh}}\biggl[\mathbb{\max}_{||\boldsymbol{\epsilon}_{ns}^{i}||<\rho_{ns}}\mathcal{L}_{\mathcal{S}}^{i}\left(\boldsymbol{\theta}_{sh}+\boldsymbol{\epsilon}_{sh},\boldsymbol{\theta}_{ns}^{i}+\boldsymbol{\epsilon}_{ns}^{i}\right)\\
 &+\left(1-\frac{1}{\sqrt{N}}\right)\frac{L}{\sqrt{N}}+\frac{1}{\sqrt{N}}+f^{i}\left(\Vert\boldsymbol{\theta}^{i}\Vert_{2}\right)\biggr]_{i=1}^{m}.
\end{align*}


To reach the final conclusion, we redefine: $$f^{i}\left(\Vert\boldsymbol{\theta}^{i}\Vert_{2}\right)=\left(1-\frac{1}{\sqrt{N}}\right)\frac{L}{\sqrt{N}}+\frac{1}{\sqrt{N}}+f^{i}\left(\Vert\boldsymbol{\theta}^{i}\Vert_{2}\right).$$

Here we note that we reach the final inequality due to the following derivations: 
\begin{flalign*}
&\mathbb{E}_{\boldsymbol{\epsilon}_{sh}}\left[\mathbb{E}_{\boldsymbol{\epsilon}_{ns}^{i}} \left[ \mathcal{L}_{\mathcal{S}}^{i} \left(\boldsymbol{\theta}_{sh}+ \boldsymbol{\epsilon}_{sh}, \boldsymbol{\theta}_{ns}^{i}+ \boldsymbol{\epsilon}_{ns}^{i} \right)\right]\right]_{i=1}^{m}  \\ 
& \hspace{0mm}\leq \int_{B_{sh}}\biggl[\int_{B^i_{ns}}\mathcal{L}_{\mathcal{S}}^{i}\left(\boldsymbol{\theta}_{sh}+\boldsymbol{\epsilon}_{sh},\boldsymbol{\theta}_{ns}^{i}+\boldsymbol{\epsilon}_{ns}^{i}\right)d\boldsymbol{\epsilon}_{ns}^{i} \\
&\qquad\qquad\qquad\qquad\qquad\qquad\qquad\qquad+\frac{1}{\sqrt{N}}\biggl]_{i=1}^{m}d\boldsymbol{\epsilon}_{sh}  \\
&+ \int_{B_{sh}^{c}}\biggl[\int_{B^i_{ns}}\mathcal{L}_{\mathcal{S}}^{i}\left(\boldsymbol{\theta}_{sh}+\boldsymbol{\epsilon}_{sh},\boldsymbol{\theta}_{ns}^{i}+\boldsymbol{\epsilon}_{ns}^{i}\right)d\boldsymbol{\epsilon}_{ns}^{i}\\
&\qquad\qquad\qquad\qquad\qquad\qquad\qquad\qquad+\frac{1}{\sqrt{N}}\biggr]_{i=1}^{m}d\boldsymbol{\epsilon}_{sh}  \\
& \hspace{0mm}\leq \int_{B_{sh}}\biggl[\int_{B^i_{ns}}\mathcal{L}_{\mathcal{S}}^{i}\left(\boldsymbol{\theta}_{sh}+\boldsymbol{\epsilon}_{sh},\boldsymbol{\theta}_{ns}^{i}+\boldsymbol{\epsilon}_{ns}^{i}\right)d\boldsymbol{\epsilon}_{ns}^{i} \\
& \qquad\qquad\qquad\qquad+ \left(1-\frac{1}{\sqrt{N}}\right)\frac{L}{\sqrt{N}}+\frac{1}{\sqrt{N}}\biggr]_{i=1}^{m}d\boldsymbol{\epsilon}_{sh}  \\ 
& \hspace{0mm}\leq \mathbb{\max}_{||\boldsymbol{\epsilon}_{sh}||<\rho_{sh}}\biggl[\\
&\qquad\qquad\mathbb{\max}_{||\boldsymbol{\epsilon}_{ns}^{i}||<\rho_{ns}}\biggl[\mathcal{L}_{\mathcal{S}}^{i}\left(\boldsymbol{\theta}_{sh}+\boldsymbol{\epsilon}_{sh},\boldsymbol{\theta}_{ns}^{i}+\boldsymbol{\epsilon}_{ns}^{i}\right) \\
&\qquad\qquad\qquad\qquad +\left(1-\frac{1}{\sqrt{N}}\right)\frac{L}{\sqrt{N}}+\frac{1}{\sqrt{N}} \biggr]\biggr]_{i=1}^m,
\end{flalign*}
where $B_{sh}=\left\{ \boldsymbol{\epsilon}_{sh}:||\boldsymbol{\epsilon}_{sh}||\leq\rho_{sh}\right\} $,  $B_{sh}^{c}$ is the complement set, and $B_{ns}^{i}=\left\{ \boldsymbol{\epsilon}_{ns}^{i}:||\boldsymbol{\epsilon}_{ns}^{i}||\leq\rho_{ns}\right\}$.

We also provide a theoretical justification for our gradient decomposition. 
\begin{thm} \label{thm:decomposition_app}
 If $\left\langle \boldsymbol{g}_{sh}^{\text{flat}},\boldsymbol{g}_{sh}^{i,\text{loss}}\right\rangle \leq0,\forall i$
and $\left\langle \boldsymbol{g}_{sh}^{\text{loss}},\boldsymbol{g}_{sh}^{i,\text{flat}}\right\rangle \leq0,\forall i$, the gradient decomposition strategy yields a smaller sum of the losses and the gradient norms. More formally, we have
\begin{align*}
&\left[l^{i}\left(\boldsymbol{\theta}_{sh}-\eta \boldsymbol{g}_{sh}^{\text{SAM,dec}}\right)+\rho_{sh}s^{i}\left(\boldsymbol{\theta}_{sh}-\eta \boldsymbol{g}_{sh}^{\text{SAM,dec}}\right)\right]_{i} \\
&\leq\left[l^{i}\left(\boldsymbol{\theta}_{sh}-\eta \boldsymbol{g}_{sh}^{\text{SAM,dir}}\right)+\rho_{sh}s^{i}\left(\boldsymbol{\theta}_{sh}-\eta \boldsymbol{g}_{sh}^{\text{SAM,dir}}\right)\right]_{i}.
\end{align*}
This theorem shows that if the aggregation vector $\boldsymbol{g}_{sh}^{\text{flat}}$ of the flatness component is \textit{non-congruent} to $\boldsymbol{g}_{sh}^{i, \text{loss}}$ of the loss component (i.e., they form an obtuse angle) and the aggregation $\boldsymbol{g}_{sh}^{\text{loss}}$ of the loss component is also \textit{non-congruent} to $\boldsymbol{g}_{sh}^{i, \text{flat}}$ of the flatness component, which possibly happens in the early stage of training, the gradient decomposition strategy optimizes the loss and the gradient norm better. The empirical evidence is given in Figure \ref{fig:decomposition_gradnorm} when the gradient decomposition strategy gains lower loss values and gradient norms than the direct strategy.
\end{thm}

\begin{proof}
    Given $\boldsymbol{\theta}^i_{ns}$, we denote $l^i(\boldsymbol{\theta}_{sh}) = \mathcal{L}_S(\boldsymbol{\theta}_{sh}, \boldsymbol{\theta}^i_{ns})$ and $s^i(\boldsymbol{\theta}_{sh}) = \Vert \nabla_{\boldsymbol{\theta}_{sh}} \mathcal{L}_S(\boldsymbol{\theta}_{sh}, \boldsymbol{\theta}^i_{ns}) \Vert_2$. We have the $\boldsymbol{g}^{i, \text{SAM}}_{sh}$ minimizes $h^i(\boldsymbol{\theta}_{sh}) = l^i(\boldsymbol{\theta}_{sh}) + \rho_{sh}s^i(\boldsymbol{\theta}_{sh})$. Therefore, their aggregation $\boldsymbol{g}^{ \text{SAM,dir}}_{sh}$ minimizes $[h^i(\boldsymbol{\theta}_{sh})]_i$. We have
    \begin{align*}
&h^{i}\left(\boldsymbol{\theta}_{sh}\right)-h^{i}\left(\boldsymbol{\theta}_{sh}-\eta \boldsymbol{g}_{sh}^{\text{SAM,dir}}\right) \\
\approx&\eta\left\langle \boldsymbol{g}_{sh}^{\text{SAM,dir}},\nabla_{\boldsymbol{\theta}_{sh}}h^{i}\left(\boldsymbol{\theta}_{sh}\right)\right\rangle \\
=& \eta\left\langle \boldsymbol{g}_{sh}^{\text{SAM,dir}},\nabla_{\boldsymbol{\theta}_{sh}}l^{i}\left(\boldsymbol{\theta}_{sh}\right)\right\rangle +\eta\rho_{sh}\left\langle \boldsymbol{g}_{sh}^{\text{SAM,dir}},\nabla_{\boldsymbol{\theta}_{sh}}s^{i}\left(\boldsymbol{\theta}_{sh}\right)\right\rangle \\
= & \eta\left\langle \boldsymbol{g}_{sh}^{\text{SAM,dir}},\boldsymbol{g}_{sh}^{i,\text{loss}}\right\rangle +\eta\rho_{sh}\left\langle \boldsymbol{g}_{sh}^{\text{SAM,dir}},\boldsymbol{g}_{sh}^{i,\text{flat}}\right\rangle .
\end{align*}
\begin{align*}
&h^{i}\left(\boldsymbol{\theta}_{sh}\right)-h^{i}\left(\boldsymbol{\theta}_{sh}-\eta \boldsymbol{g}_{sh}^{\text{SAM,dec}}\right)\\
\approx&\eta\left\langle \boldsymbol{g}_{sh}^{\text{SAM,dec}},\nabla_{\boldsymbol{\theta}_{sh}}h^{i}\left(\boldsymbol{\theta}_{sh}\right)\right\rangle \\
= & \eta\left\langle \boldsymbol{g}_{sh}^{\text{SAM,dec}},\nabla_{\boldsymbol{\theta}_{sh}}l^{i}\left(\boldsymbol{\theta}_{sh}\right)\right\rangle +\eta\rho_{sh}\left\langle \boldsymbol{g}_{sh}^{\text{SAM,dec}},\nabla_{\boldsymbol{\theta}_{sh}}s^{i}\left(\boldsymbol{\theta}_{sh}\right)\right\rangle \\
= & \eta\left\langle \boldsymbol{g}_{sh}^{\text{SAM,dec}},\boldsymbol{g}_{sh}^{i,\text{loss}}\right\rangle +\eta\rho_{sh}\left\langle \boldsymbol{g}_{sh}^{\text{SAM,dec}},\boldsymbol{g}_{sh}^{i,\text{flat}}\right\rangle \\
= & \eta\left\langle \boldsymbol{g}_{sh}^{\text{loss}}+\boldsymbol{g}_{sh}^{\text{flat}},\boldsymbol{g}_{sh}^{i,\text{loss}}\right\rangle +\eta\rho_{sh}\left\langle \boldsymbol{g}_{sh}^{\text{loss}}+\boldsymbol{g}_{sh}^{\text{flat}},\boldsymbol{g}_{sh}^{i,\text{flat}}\right\rangle \\
= & \eta\left\langle \boldsymbol{g}_{sh}^{\text{loss}},\boldsymbol{g}_{sh}^{i,\text{loss}}\right\rangle +\eta\rho_{sh}\left\langle \boldsymbol{g}_{sh}^{\text{flat}},\boldsymbol{g}_{sh}^{i,\text{flat}}\right\rangle \\
&+\eta\left\langle \boldsymbol{g}_{sh}^{\text{flat}},\boldsymbol{g}_{sh}^{i,\text{loss}}\right\rangle +\eta\rho_{sh}\left\langle \boldsymbol{g}_{sh}^{\text{loss}},\boldsymbol{g}_{sh}^{i,\text{flat}}\right\rangle \\
\leq & \eta\left\langle \boldsymbol{g}_{sh}^{\text{loss}},\boldsymbol{g}_{sh}^{i,\text{loss}}\right\rangle +\eta\rho_{sh}\left\langle \boldsymbol{g}_{sh}^{\text{flat}},\boldsymbol{g}_{sh}^{i,\text{flat}}\right\rangle .
\end{align*}

Due to the definition of $\boldsymbol{g}_{sh}^{\text{loss}}$ and $\boldsymbol{g}_{sh}^{\text{flat}}$,
we have $\left\langle \boldsymbol{g}_{sh}^{\text{loss}},\boldsymbol{g}_{sh}^{i,\text{loss}}\right\rangle \geq\left\langle \boldsymbol{g}_{sh}^{\text{SAM,dir}},\boldsymbol{g}_{sh}^{i,\text{loss}}\right\rangle $
and $\left\langle \boldsymbol{g}_{sh}^{\text{flat}},\boldsymbol{g}_{sh}^{i,\text{flat}}\right\rangle \geq\left\langle \boldsymbol{g}_{sh}^{\text{SAM,dir}},\boldsymbol{g}_{sh}^{i,\text{flat}}\right\rangle $.
This follows that 
\begin{align*}
&h^{i}\left(\boldsymbol{\theta}_{sh}\right)-h^{i}\left(\boldsymbol{\theta}_{sh}-\eta \boldsymbol{g}_{sh}^{\text{SAM,dir}}\right)\\
&\leq h^{i}\left(\boldsymbol{\theta}_{sh}\right)-h^{i}\left(\boldsymbol{\theta}_{sh}-\eta \boldsymbol{g}_{sh}^{\text{SAM,dec}}\right) \\
\Rightarrow &h^{i}\left(\boldsymbol{\theta}_{sh}-\eta \boldsymbol{g}_{sh}^{\text{SAM,dec}}\right)\leq h^{i}\left(\boldsymbol{\theta}_{sh}-\eta \boldsymbol{g}_{sh}^{\text{SAM,dir}}\right) \\
\Rightarrow &l^{i}\left(\boldsymbol{\theta}_{sh}-\eta \boldsymbol{g}_{sh}^{\text{SAM,dec}}\right)+\rho_{sh}s^{i}\left(\boldsymbol{\theta}_{sh}-\eta \boldsymbol{g}_{sh}^{\text{SAM,dec}}\right)\\
&\leq l^{i}\left(\boldsymbol{\theta}_{sh}-\eta \boldsymbol{g}_{sh}^{\text{SAM,dir}}\right)+\rho_{sh}s^{i}\left(\boldsymbol{\theta}_{sh}-\eta \boldsymbol{g}_{sh}^{\text{SAM,dir}}\right),\forall i \\
\Rightarrow &\left[l^{i}\left(\boldsymbol{\theta}_{sh}-\eta \boldsymbol{g}_{sh}^{\text{SAM,dec}}\right)+\rho_{sh}s^{i}\left(\boldsymbol{\theta}_{sh}-\eta \boldsymbol{g}_{sh}^{\text{SAM,dec}}\right)\right]_{i} \\ 
&\leq\left[l^{i}\left(\boldsymbol{\theta}_{sh}-\eta \boldsymbol{g}_{sh}^{\text{SAM,dir}}\right)+\rho_{sh}s^{i}\left(\boldsymbol{\theta}_{sh}-\eta \boldsymbol{g}_{sh}^{\text{SAM,dir}}\right)\right]_{i}.
\end{align*}

\end{proof}

\section{Gradient aggregation strategies overview  \label{sec:grad_agg}}
This section details how the gradient\_aggregate operation is defined according to recent gradient-based multi-task learning methods that we employed as baselines in the main paper, including  MGDA \citep{sener2018multi}, PCGrad \citep{yu2020gradient}, CAGrad \citep{liu2021conflict} and IMTL \citep{liu2021towards}. Assume that we are given $m$ vectors $\boldsymbol{g}^{1}, \boldsymbol{g}^{2}, \dots, \boldsymbol{g}^{m}$ represent task gradients. Typically, we aim to find a combined gradient vector as:

$$
\boldsymbol{g}= \text{gradient\_aggregate}(\boldsymbol{g}^{1}, \boldsymbol{g}^{2}, \dots, \boldsymbol{g}^{m}).
$$

\subsection{Multiple-gradient descent algorithm - MGDA}
\cite{sener2018multi} applies MGDA \citep{desideri2012multiple} to find the minimum-norm gradient vector that lies in the convex hull composed by task gradients $\boldsymbol{g}^{1}, \boldsymbol{g}^{2}, \dots, \boldsymbol{g}^{m}$:
 \begin{equation*}
     \boldsymbol{g} = \text{argmin} ||\sum_{i=1}^m w_i\boldsymbol{g}^i||^2, s.t.\sum_{i=1}^m w_i=1  \quad \textrm{and}  \quad w_i \geq 0  \forall i.
 \end{equation*}
This approach can guarantee that the obtained solutions lie on the Pareto front of task objective functions.

\subsection{Projecting conflicting gradients - PCGrad}
PCgrad resolves the disagreement between tasks by projecting gradients that conflict with each other, i.e. $\langle \boldsymbol{g}^i, \boldsymbol{g}^j\rangle < 0$, to the orthogonal direction of each other. Specifically, $\boldsymbol{g}^i$ is replaced by its projection on the normal plane of $\boldsymbol{g}^j$: 
\begin{equation*}
    \boldsymbol{g}^i_{\mathrm{PC}} = \boldsymbol{g}^i - \frac{\boldsymbol{g}^i \cdot \boldsymbol{g}^j}{||\boldsymbol{g}^j||^2}\boldsymbol{g}^j.
\end{equation*}
Then compute the aggregated gradient based on these deconflict vectors $\mathbf{g}=\sum_i^m \mathbf{g}^i_{\mathrm{PC}}$.
\subsection{Conflict Averse Gradient Descent - CAGrad}
CAGrad \citep{liu2021conflict} seeks a worst-case direction in a local ball around the average gradient of all tasks, $\boldsymbol{g}_0$, that minimizes conflict with all of the gradients. The updated vector is obtained by optimizing the following problem: 
\begin{equation*}
    \max_{ \mathbf{g} \in R} \min_{i\in[m]}\langle \mathbf{g}^i,\mathbf{g} \rangle \quad s.t. \quad || \mathbf{g}-\mathbf{g}^0||\leq c ||\mathbf{g}^0||,
\end{equation*}
where $ \mathbf{g}^0=\frac{1}{m} \sum_i^m  \mathbf{g}^i$ is the averaged gradient and $c$ is a hyper-parameter.

\subsection{Impartial multi-task learning - IMTL}

IMTL \citep{liu2021towards} proposes to balance per-task gradients by 
finding the combined vector $\mathbf{g}$, whose projections onto $\{\mathbf{g}^i\}_{i=1}^m$ are equal. Following this, they obtain the closed-form solution for the simplex vector $\boldsymbol{w}$ for reweighting task gradients:
$$\boldsymbol{w}=\boldsymbol{g}^1 \boldsymbol{U}^{\top}\left(\boldsymbol{D} \boldsymbol{U}^{\top}\right)^{-1}$$
where $\boldsymbol{u}^i=\boldsymbol{g}^i /\left\|\boldsymbol{g}^i\right\|$,  $\boldsymbol{U}=\left[\boldsymbol{u}^1-\boldsymbol{u}^2, \cdots, \boldsymbol{u}^1-\boldsymbol{u}^m\right]$, and $\boldsymbol{D}=\left[\boldsymbol{g}^1-\boldsymbol{g}^2, \cdots, \boldsymbol{g}^1-\boldsymbol{g}^m\right]$
The aggregated vector is then calculated as $\mathbf{g}=\sum_i^m w_i \mathbf{g}^i$.

\section{Implementation Details \label{sec:implement}}

In this part, we provide implementation details regarding the empirical evaluation in the main paper, along with additional comparison experiments. All experiments are run on a single A100 GPU (40 GB VRAM).

\subsection{Baselines}
In this subsection, we briefly introduce some of the comparative methods that appeared in the main text:
\begin{itemize}
    \item {Linear scalarization} (LS) minimizes the unweighted sum of task objectives $\sum_i^m\mathcal{L}^i(\boldsymbol{\theta})$.
    \item {Scale-invariant} (SI) aims toward obtaining similar convergent solutions even if losses are scaled with different coefficients via minimizing $\sum_i^m\log\mathcal{L}^i(\boldsymbol{\theta})$.
    \item {Random loss weighting} (RLW) \citep{linreasonable} is a simple yet effective method for balancing task losses or gradients by random weights.
    \item Dynamic Weight Average (DWA) \citep{liu2019end} simply adjusts the weighting coefficients by taking the rate of
change of loss for each task into account.
    \item {GradDrop }  \citep{chen2020just} presents a probabilistic masking process that algorithmically eliminates all gradient values having the opposite sign w.r.t a predefined direction.
\end{itemize}

\subsection{Image classification}
\textbf{Datasets.} \\
- Multi-MNIST\footnote{https://github.com/Xi-L/ParetoMTL}. Following the protocol of \cite{sener2018multi}, we set up three Multi-MNIST experiments with ResNet18 \citep{he2016deep}, namely: MultiFashion, MultiMNIST and MultiFashion+MNIST. In each dataset, two images are sampled uniformly from the MNIST \citep{lecun1998gradient} or Fashion-MNIST \citep{xiao2017fashion}, then one is placed on the top left, and the other is on the bottom right. We thus obtain a two-task learning that requires predicting the categories of the digits or fashion items on the top left (task 1) and the bottom right (task 2), respectively.  \\
- CelebA \citep{liu2018large} is a face dataset with 200K images and 40 attributes, forming a 40-class multi-label classification problem.

\textbf{Network Architectures.} For two datasets in this problem, {Multi-MNIST} and {CelebA}, we replicate experiments from \citep{sener2018multi, lin2019pareto} by respectively using the Resnet18 (11M parameters) and Resnet50 (23M parameters)  \citep{he2016deep} with the last output layer removed as the shared encoders and constructing linear classifiers as the task-specific heads, i.e. 2 heads for Multi-MNIST and 40 for CelebA, respectively.  

\textbf{Training Details.} We train the all the models under our proposed framework and baselines using:
\begin{itemize}
    \item Multi-MNIST: Adam optimizer \citep{kingma2014adam} with a learning rate of $0.001$ for $200$ epochs using a batch size of 256. Images from the three datasets are resized to $36\times36$.
    \item CelebA: Batch-size of $256$ and images are resized to $64\times64\times3$. Adam  \citep{kingma2014adam} is used again with a learning rate of $0.0005$, which is  decayed by $0.85$ for every 10 epochs,  our model is trained for $50$ epochs in total.
\end{itemize} 

Regarding the hyper-parameter for SAM \citep{foret2021sharpnessaware}, we use their adaptive version \citep{pmlr-v139-kwon21b} where both $\rho_{sh}$ and $\rho_{ns}$ are set equally and extensively tuned from $0.005$ to  $5$. More details can be found in the public source code.

\subsection{Scene understanding}
Two datasets used in this problem are  NYUv2 and CityScapes:

- NYUv2\footnote{\label{note1}https://github.com/Cranial-XIX/CAGrad} is an indoor scene dataset that contains 3 tasks: 13-class semantic segmentation, depth estimation, and surface normal prediction. 

- CityScapes\textsuperscript{\ref{note1}} has 19 classes of street-view images, which are coarsened into 7 categories to create two tasks: semantic segmentation and depth estimation.

Similar to \citep{navon2022multi}, all images in the NYUv2 dataset are resized to $288 \times 384$ while all images in the  CityScapes dataset are resized to $128 \times 256$ to speed up the training process. 
We follow the exact protocol in \citep{navon2022multi} for implementation. Specifically, SegNet \citep{badrinarayanan2017segnet} is adopted as the architecture for the backbone and Multi-Task Attention Network MTAN \citep{liu2019end} is applied on top of it. 
We train each method for 200 epochs using Adam optimizer \citep{kingma2014adam} with an initial learning rate of $1e^{-4}$ and reduce it to $5e^{-5}$ after 100 epochs. We use a batch size of 2 for NYUv2 and 8 for CityScapes. The last 10 epochs are averaged to get the final results, and all experiments are run with three random seeds. More details can be found in the public source code.

\section{Additional Results}
\label{sec:supp_res}

To further show the improvement of our proposed training framework over the conventional one,
this section provides additional comparison results
in terms of qualitative results, predictive performance, convergent behavior, loss landscape, model sharpness, and gradient norm.
We also complete the ablation study in the main paper by providing results on all three datasets in the Multi-MNIST dataset.

\subsection{Image segmentation qualitative result}
\label{subsec:segmen_qual}

\begin{figure*}[!ht]
\subfloat[A training sample (after augmentation)]{%
\centering
  \includegraphics[width=\textwidth, trim=3cm 7cm 2.5cm 6.5cm,clip]{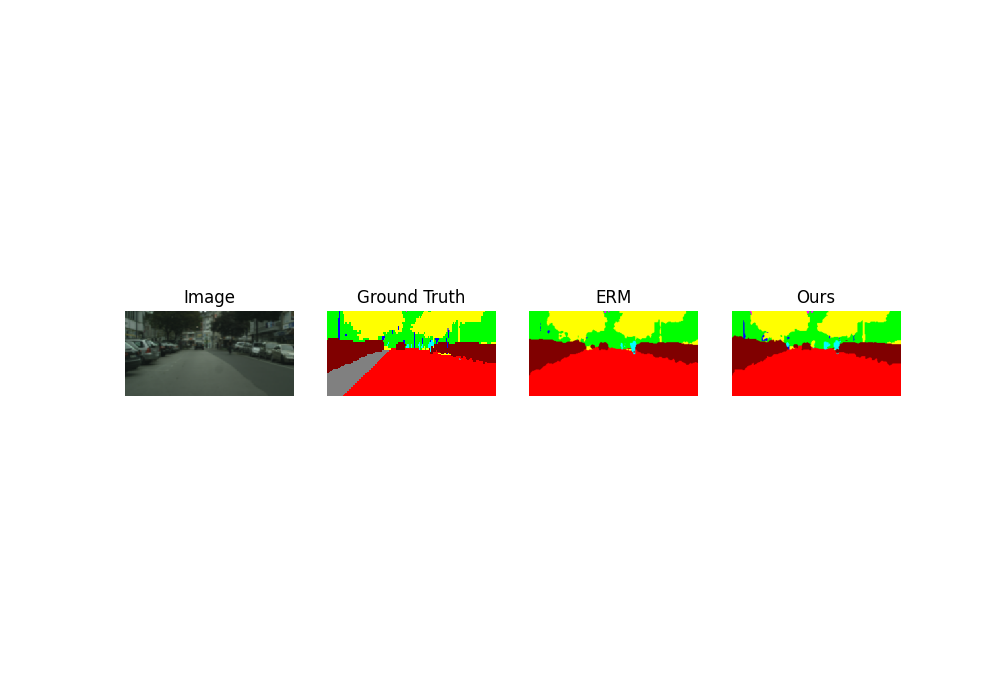}\label{fig:train_aug}
}

\subfloat[Corresponding original image (before augmentation)]{%
\centering
  \includegraphics[width=\textwidth, trim=3cm 7cm 2.5cm 6.5cm,clip]{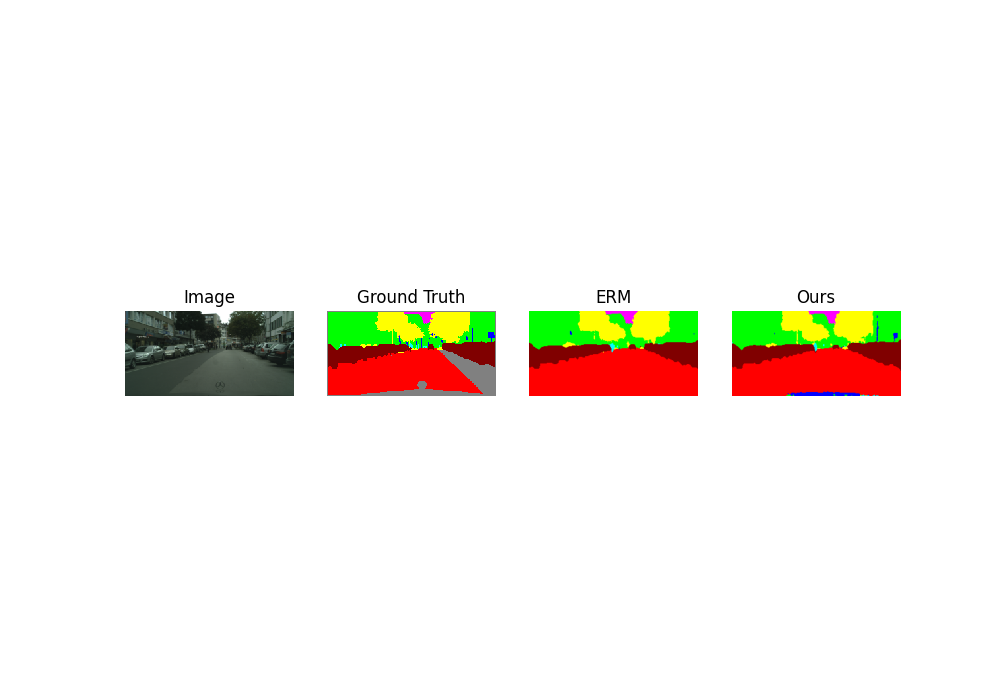}\label{fig:train_data}
}

\subfloat[Predictions on an unseen image]{%
\centering
  \includegraphics[width=\textwidth, trim=3cm 7cm 2.5cm 6.5cm,clip]{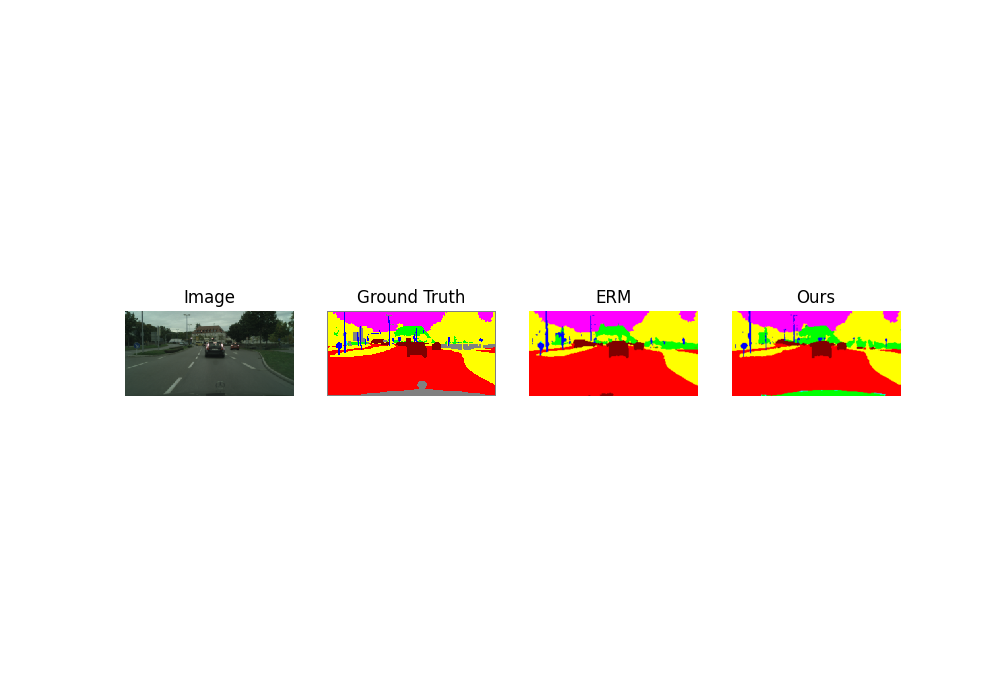}\label{fig:unseen1}
}

\subfloat[Predictions on an unseen image]{%
\centering
  \includegraphics[width=\textwidth, trim=3cm 7cm 2.5cm 6.5cm,clip]{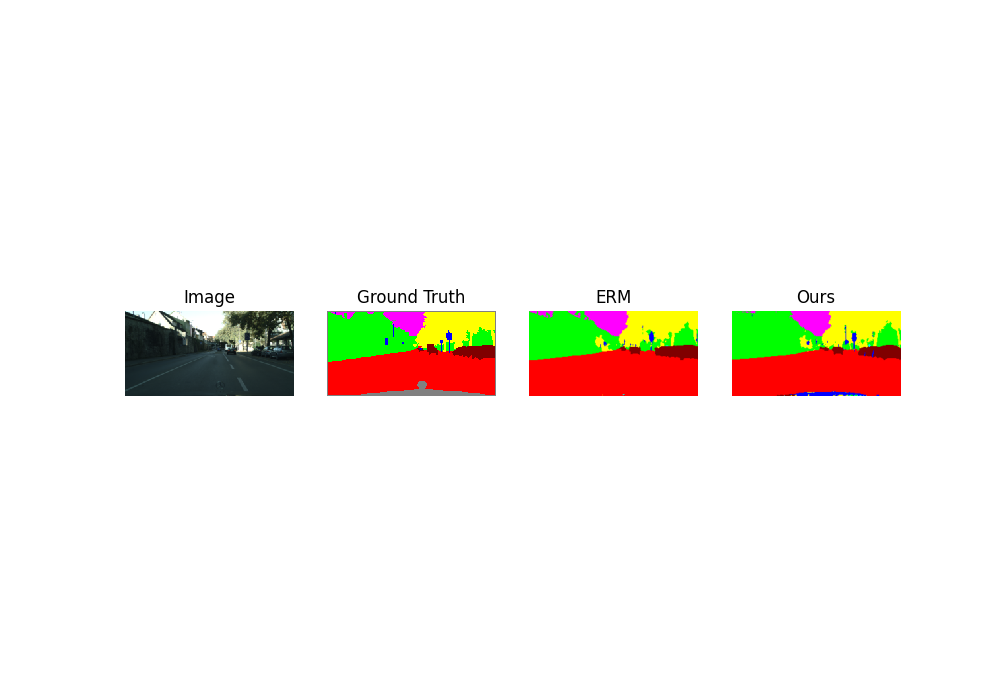}\label{fig:unseen2}
}

\caption{ Semantic segmentation prediction comparison on CityScapes \label{fig:quanlitative}. From left to right are input images, ground truth, and segmentation outputs from SegNet \citep{badrinarayanan2017segnet} using ERM training and sharpness-aware training. Regions that are represented in gray color are ignored during training. (Best viewed in color).}
\end{figure*}

In this section, we provide qualitative results of our method of the  CityScapes experiment. We compare our proposed method against its main baseline by highlighting typical cases where our method excels in generalization performance. Figure \ref{fig:quanlitative} shows some visual examples of segmentation outputs on the test set. Note that in the CityScapes dataset, the ``void" class is identified as unclear and pixels labeled as void do not contribute to either objective or score \citep{cordts2016cityscapes}.

While there is only a small gap between the segmentation performance of ERM and ours, we found that a small area, which is the car hood and located at the bottom of images, is often incorrectly classified. For example, in Figure \ref{fig:quanlitative}, the third and fourth rows compare the prediction of SegNet \citep{badrinarayanan2017segnet} with ERM training and with our proposed method. It can be seen that both of them could not detect this area correctly, this is because this unclear ``void" class did not appear during training. Even worse, the currently employed data augmentation technique in the codebase of Nash-MTL and other recent multi-task learning methods \cite{navon2022multi, liu2021conflict} consists of RandomCrop, which often unintentionally excludes edge regions. For example, Figure \ref{fig:train_aug} shows an example fed to the neural network for training, which excludes the car hood and its logo, compared to the original image (Figure \ref{fig:train_data}). Therefore, we can consider this "void" class as a novel class in this experiment, since its appearance is ignored in both training and evaluation. Even though, in Figures \ref{fig:unseen1} and \ref{fig:unseen2} our training method is still able to distinguish between this unknown area and other nearby known classes, which empirically shows the robustness and generalization ability of our method over ERM.

\subsection{Predictive performance}
\label{subsec:calibration}

In this part, we provide experimental justification for an intriguing insight into the connection between model sharpness and model calibration. Empirically, we found that when a model converges to flatter minima, it tends to be more calibrated. We start by giving the formal definition of a well-calibrated classification model and three metrics to measure the calibration of a model, then we analyze our empirical results.

Consider a $C$-class classification problem with a test set of $N$ samples given in the form $(x_i, y_i)_{i=1}^N$ where $y_i$ is the true label for the sample $x_i$. Model outputs the predicted probability for a given sample $x_i$ to fall into $C$ classes, is given by $$\hat{\boldsymbol{p}}(x_i) = [\hat{p}(y=1|x_i), \dots, {\hat{p}}(y=C|x_i)].$$
$\hat{p}(y=c|x_i)$ is also the confidence of the model when assigning the sample $x_i$ to class $c$.
The predicted label $\hat{y}_i$ is the class with the highest predicted value, $\hat{p}(x_i) := \max_{c}\hat{p}(y=c|x_i)$. We refer to $\hat{p}(x_i)$ as the confidence score of a sample $x_i$.

\textbf{Model calibration} is a desideratum of modern deep neural networks, which indicates that the predicted probability of a model should match its true probability. This means that the classification network should be not only accurate but also confident about its prediction, i.e. being aware of when it is likely to be incorrect. Formally stated, the \textit{perfect calibration} \citep{guo2017calibration} is: 
\begin{equation}
  P(\hat{y} = y | \hat{p} = q) = q,  \forall q \in [0,1].
  \label{perfect_cali}
\end{equation}

\textbf{Metric.} The exact computation of Equation \ref{perfect_cali} is infeasible, thus we need to define some metrics to evaluate how well-calibrated a model is. 
\begin{itemize}
    \item {Brier score $\downarrow$} (BS) \citep{brier1950verification} assesses the accuracy of a model's predicted probability by taking into account the absolute difference between its confidence for a sample to fall into a class and the true label of that sample. Formally, 
    \begin{equation*}
        BS = \frac{1}{N} \sum_{i=1}^N \sum_{c=1}^C \left(\hat{p}(y=c|x_i) - \boldsymbol{1}[y_i=c]\right)^2.
    \end{equation*}

    \item {Expected calibration error $\downarrow$ } (ECE) compares the predicted probability (or confidence) of a model to its accuracy \citep{naeini2015obtaining, guo2017calibration}. To compute this error, we first bin the confidence interval $[0,1]$ into $M$ equal bins, then categorize data samples into these bins according to their confidence scores. 
    We finally compute the absolute value of the difference between the average confidence and the average accuracy within each bin, and report the average value over all bins as the ECE. Specifically, let $B_m$ denote the set of indices of samples having their confidence scores belonging to the $m^{th}$ bin. The average accuracy and the average confidence within this bin are:  
    \begin{equation*}
    \begin{aligned}
        acc(B_m) &= \frac{1}{|B_m|} \sum_{i \in B_m} \boldsymbol{1}[\hat{y_i} = y_i], \\
        conf(B_m) &= \frac{1}{|B_m|} \sum_{i \in B_m}\hat{p}(x_i).
    \end{aligned}
    \end{equation*}
    Then the ECE of the model is defined as:
    \begin{equation*}
        ECE = \sum_{m=1}^M \frac{|B_m|}{N} |acc(B_m) - conf(B_m)|.
    \end{equation*}
In short, the lower ECE neural networks obtain, the more calibrated they are.
    
    \item {Predictive entropy} (PE) is a widely-used measure of uncertainty \citep{ovadia2019can, lakshminarayanan2017simple, malinin2018predictive} via the predictive  probability of the model output.
 When encountering an unseen sample, a well-calibrated model is expected to yield a high PE, representing its uncertainty in predicting out-of-domain (OOD) data.
    \begin{equation*}
        PE = \frac{1}{C} \sum_{c=1}^C -\hat{p}(y=c|x_i) \log \hat{p}(y=c|x_i).
    \end{equation*}
\end{itemize}

\begin{figure}[!ht]
\centering

  \centering
  \includegraphics[width=1.\columnwidth, trim=.cm 0cm 0.cm 0cm,clip]{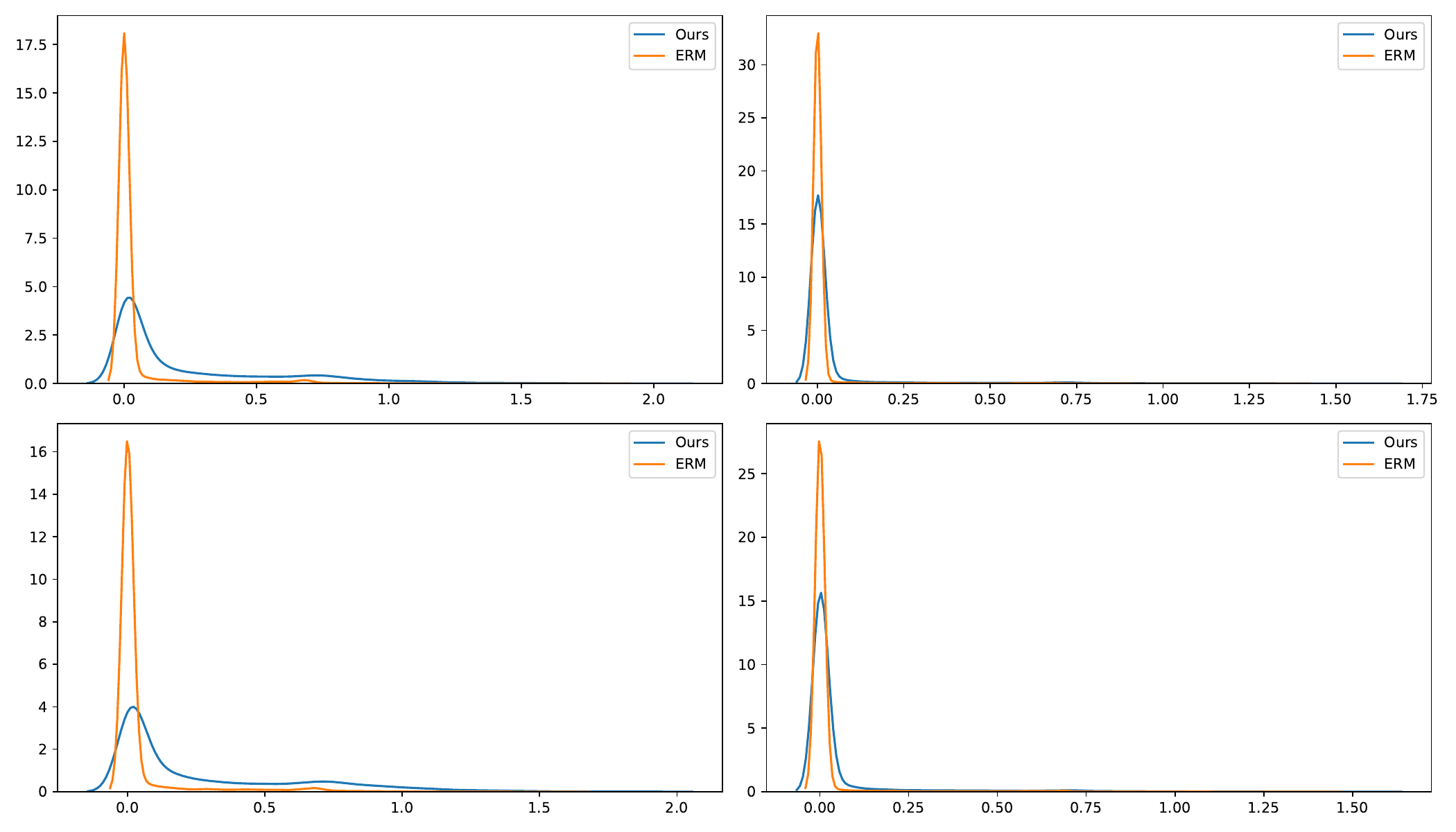}
  \caption{Histograms of predictive entropy of ResNet18 \citep{he2016deep}  on in-domain dataset, train and test on MultiMNIST (left) and MultiFashion (right). We use the \textcolor{orange}{orange} lines to denote ERM training while  \textcolor{blue}{blue} lines indicate our proposed method.}\label{fig:in-entropy}
\end{figure}
\begin{figure}[!ht]
  \centering
  \includegraphics[width=1.\columnwidth, trim=.cm 0cm 0.cm 0cm,clip]{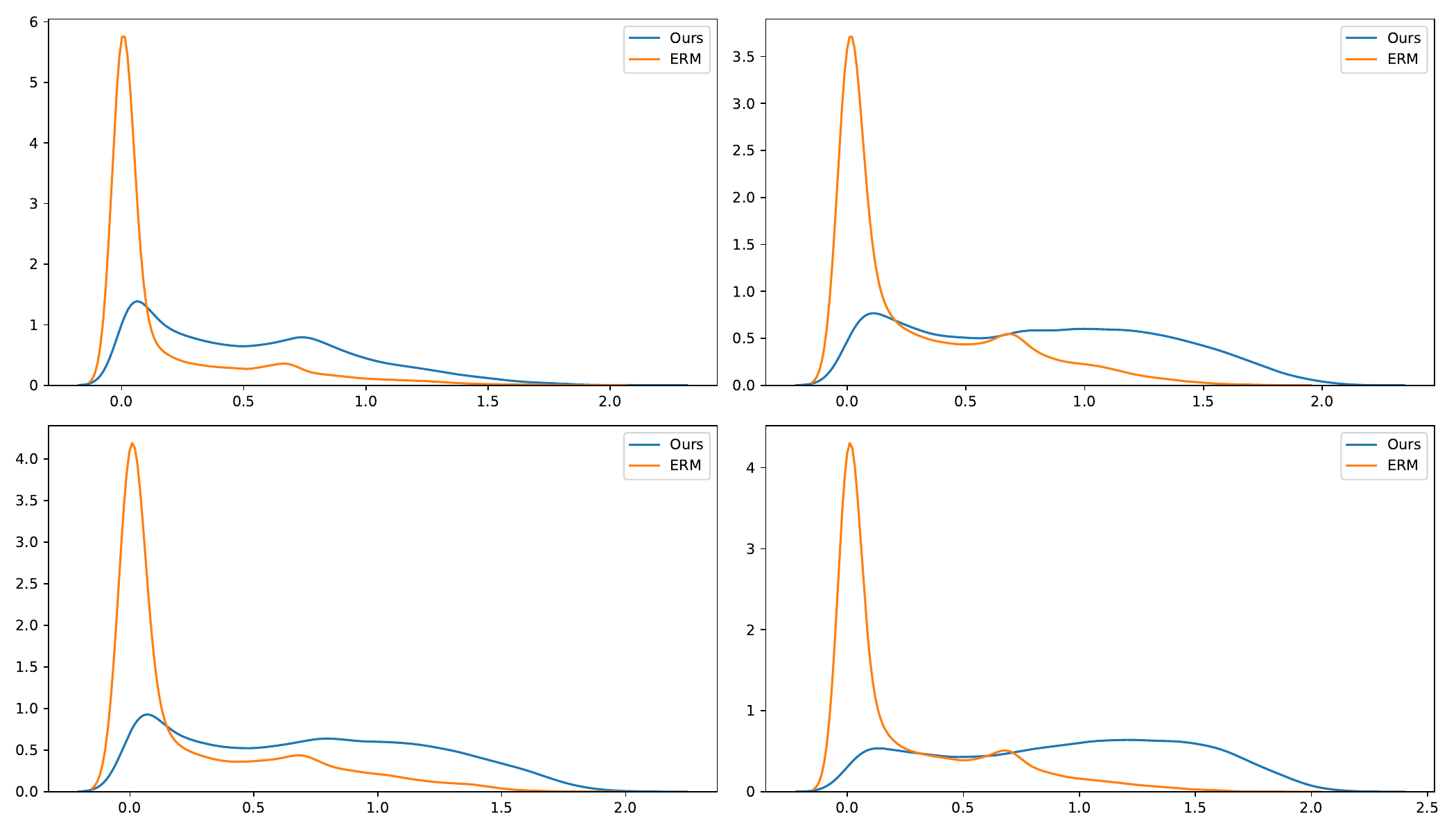}
  \caption{Out of domain: model is trained on MultiMNIST, then tested on MultiFashion (left) and vice versa (right). Models trained with ERM give over-confident predictions as their predictive entropy concentrates around $0$.}\label{fig:out-entropy}
\end{figure}

 Figures \ref{fig:in-entropy} and \ref{fig:out-entropy} plot the distribution of the model's predicted entropy in the case of in-domain and out-domain testing, respectively. We can see when considering the flatness of minima, the model shows higher predictive entropy on both in-domain and out-of-domain, compared to ERM. This also means that our model outputs high uncertainty prediction when it is exposed to a sample from a different domain. 

\begin{figure}[!ht]
  \centering
  \includegraphics[width=.95\columnwidth, trim=.cm 0cm 0.cm 0cm,clip]{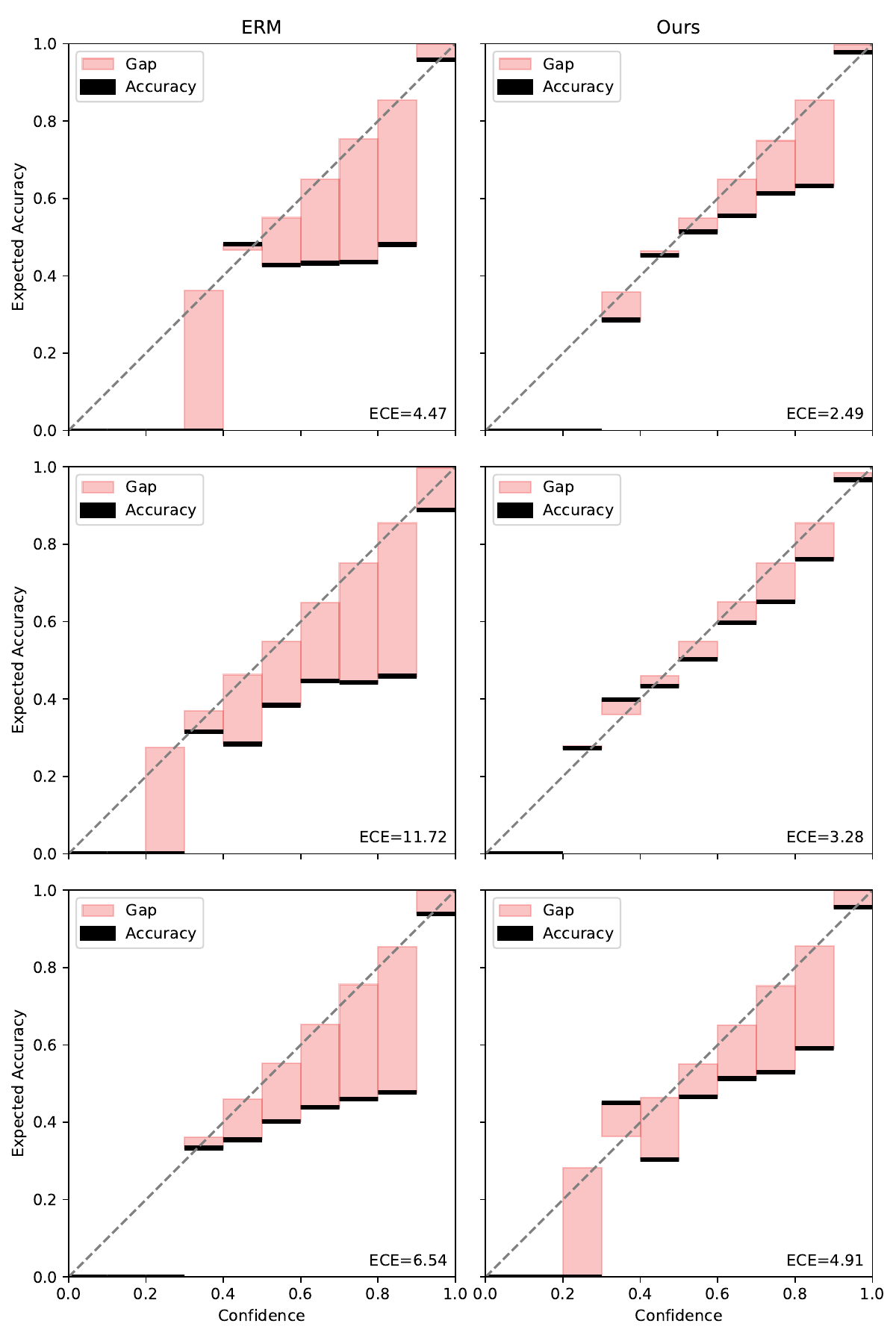}
\caption{ The predictive performance (measured by the expected calibration error) of neural networks has been enhanced by using our proposed training method (right column). }
\label{fig:ECE}
\end{figure}

Here, we calculate the results for both tasks 1 and 2 as a whole and plot their ECE in Figure \ref{fig:ECE}. When we look at the in-domain prediction in more detail, our model still outperforms ERM in terms of expected calibration error. We hypothesize that considering a flat minima optimizer not only lowers errors across tasks but also improves the predictive performance of the model.

\begin{table}[!ht]

\centering{}\resizebox{\columnwidth}{!}{
\begin{tabular}{c|c|c|c|c}
\toprule
Dataset & Task & Multi-Fashion & Multi-Fashion+MNIST & MultiMNIST \tabularnewline
\midrule
\multirow{3}{*}{ERM} & Top left & 0.237 & 0.055 & 0.082 \tabularnewline
 & Bottom right & 0.254 & 0.217 & 0.106 \tabularnewline
 \cmidrule{2-5}
 & Average & 0.246 & 0.136 & 0.094 \tabularnewline
\midrule
\multirow{3}{*}{Ours} & Top left & \textbf{0.172}  & \textbf{0.037} & \textbf{0.059} \tabularnewline
 & Bottom right & \textbf{0.186} & \textbf{0.189} & \textbf{0.075} \tabularnewline
 \cmidrule{2-5}
 & Average & \textbf{0.179} & \textbf{0.113} & \textbf{0.067} \tabularnewline
\bottomrule
\end{tabular}
}
\caption{Brier score on Multi-Fashion, Multi-Fashion+MNIST and MultiMNIST
datasets. We use the \textbf{bold} font to highlight the best results.}
\label{tab:mnist-brier}
\end{table}

We also report the Brier score and ECE for each task in Table \ref{tab:mnist-brier} and Table \ref{tab:mnist-ece}. As can be observed from these tables, our method shows consistent improvement in the model calibration when both scores decrease over all scenarios.

\begin{table}[!ht]
\centering{}\resizebox{\columnwidth}{!}{%
\begin{tabular}{c|c|c|c|c}
\toprule
Dataset & Task & Multi-Fashion & Multi-Fashion+MNIST & MultiMNIST \tabularnewline
\midrule
\multirow{3}{*}{ERM} & Top left & 0.113 & 0.027 & 0.039 \tabularnewline
 & Bottom right & 0.121 & 0.104 & 0.050 
  \tabularnewline
  \cmidrule{2-5}
 & Average & 0.117 & 0.066 & 0.045 \tabularnewline
\midrule
\multirow{3}{*}{Ours} & Top left & \textbf{0.034}  & \textbf{0.015} & \textbf{0.022} \tabularnewline
 & Bottom right & \textbf{0.032} & \textbf{0.083} & \textbf{0.028} \tabularnewline
 \cmidrule{2-5}
  & Average & \textbf{0.033} & \textbf{0.049} & \textbf{0.025} \tabularnewline
\bottomrule
\end{tabular}}
\caption{Expected calibration error on Multi-Fashion, Multi-Fashion+MNIST and MultiMNIST datasets. Here we set the number of bins equal to $10$.}
\label{tab:mnist-ece}
\end{table}

\subsection{Effect of choosing perturbation radius 
$\rho$.} 
\label{subsec:rho_ablation}
\begin{figure}[!ht]
    \centering
     \includegraphics[width=\columnwidth, trim=0cm 0.cm 0.0cm 0cm,clip]{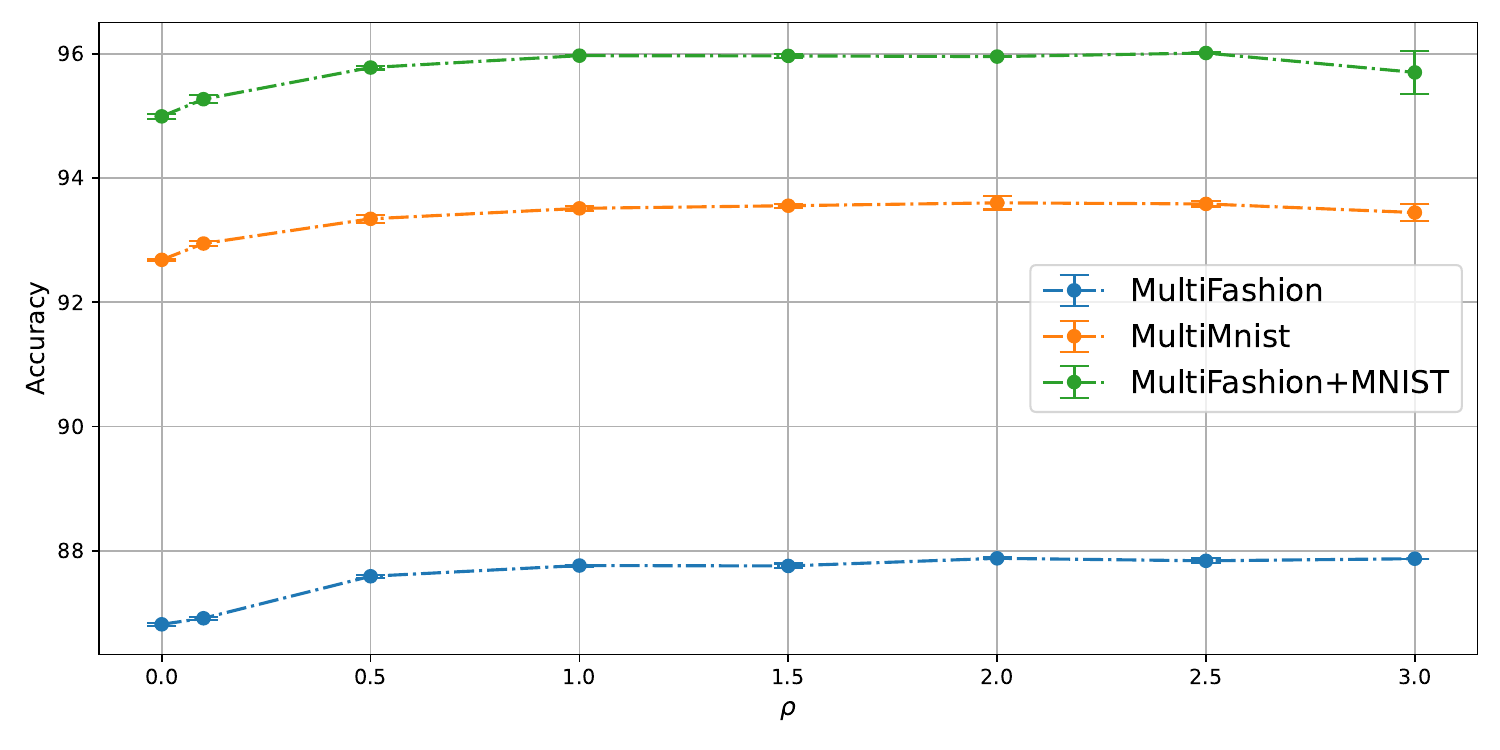}
    \caption{Average accuracy when varying $\rho$ from $0$ to $3.0$ (with error bar from three independent runs).}
    \label{fig:rho}
\end{figure}

The experimental results analyzing the sensitivity of the model w.r.t $\rho$ are given in Figure \ref{fig:rho}. We evenly picked $\rho$ from $0$ to $3.0$ to run experiments on three Multi-MNIST datasets.  We find that the average accuracy of each task is rather stable from $\rho=0.5$, which means the effect of different values of $\rho$ in a reasonably small range is similar.  It can also be easy to notice that the improvement tends to saturate when $\rho\geq1.5$.

 \subsection{Loss landscape}
\label{subsec:loss_viz}

\begin{figure}[!h]
\centering

\begin{subfigure}{\linewidth}
    \centering
    \includegraphics[width=1.\columnwidth,scale=0.35, trim=2.5cm 2.5cm 2cm 4cm,clip]{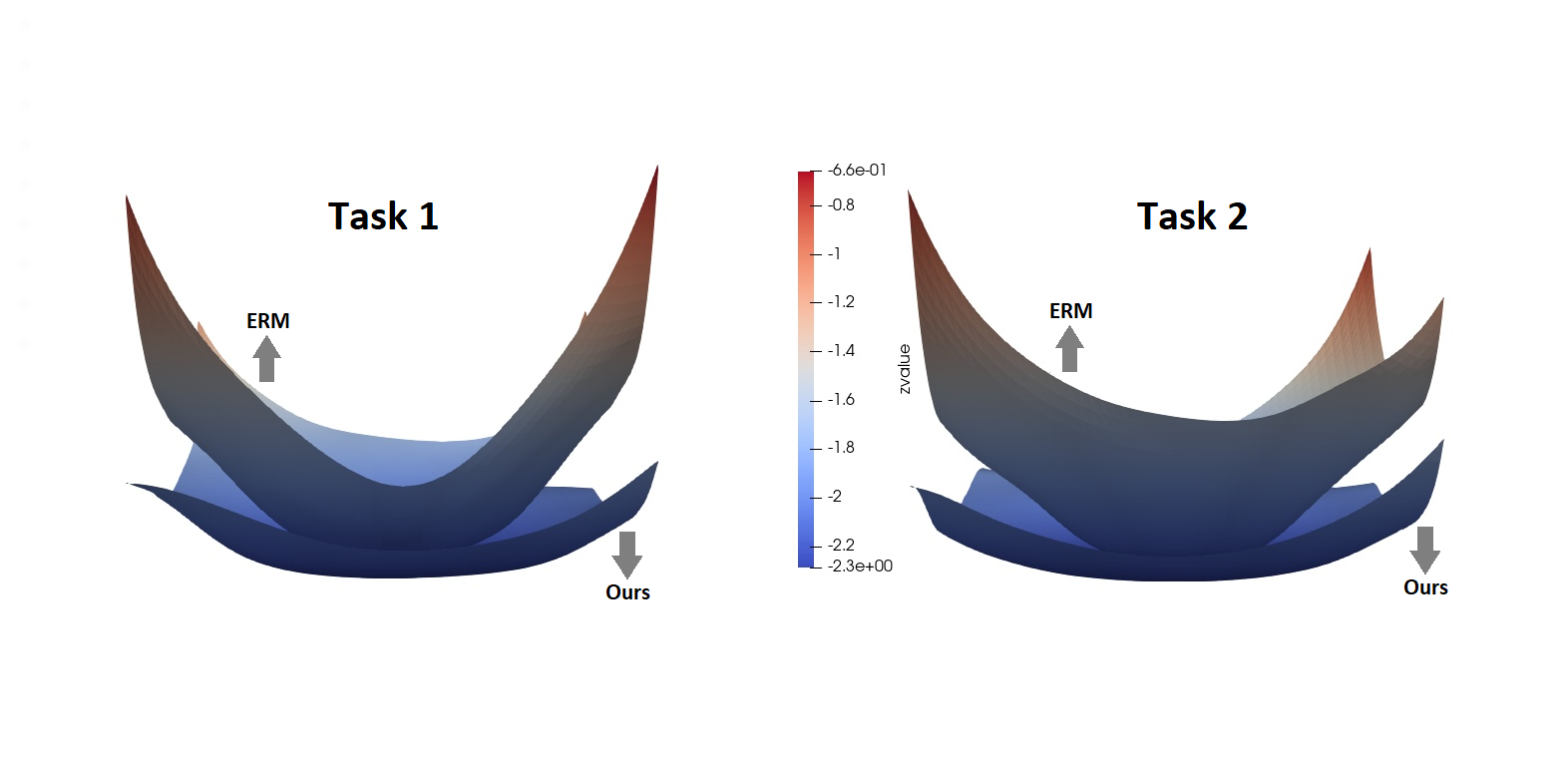}
    \caption{MultiMNIST}\label{fig:mnist-loss}
\end{subfigure}

\begin{subfigure}{\linewidth}
    \centering
    \includegraphics[width=1.\columnwidth, scale=0.35, trim=2.5cm 2.5cm 2cm 4cm,clip]{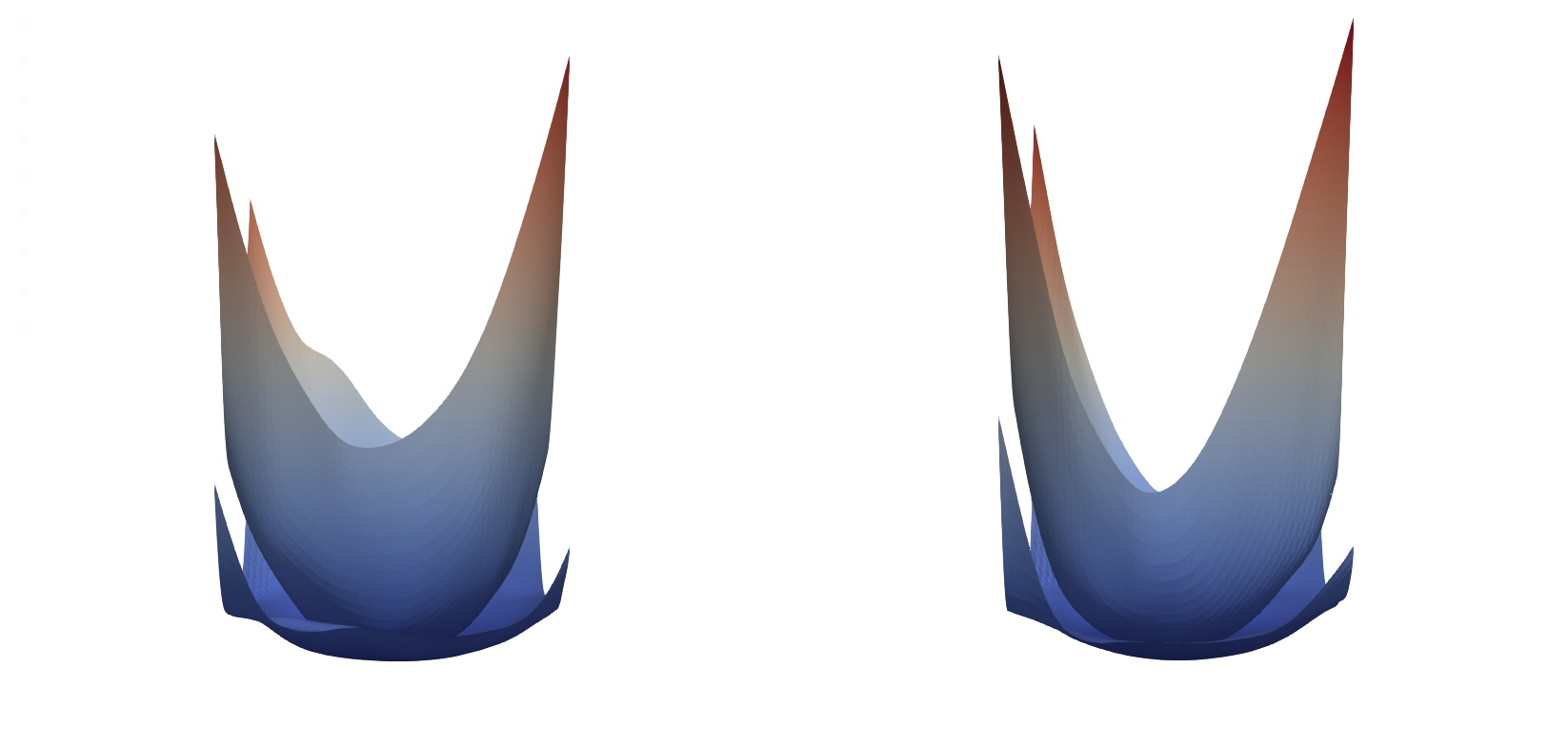}
    \caption{MultiFashion}\label{fig:fashion-loss}
\end{subfigure}

\begin{subfigure}{\linewidth}
  \centering
  \includegraphics[width=1.\columnwidth, scale=0.35, trim=2.5cm 2.5cm 2cm 4cm,clip]{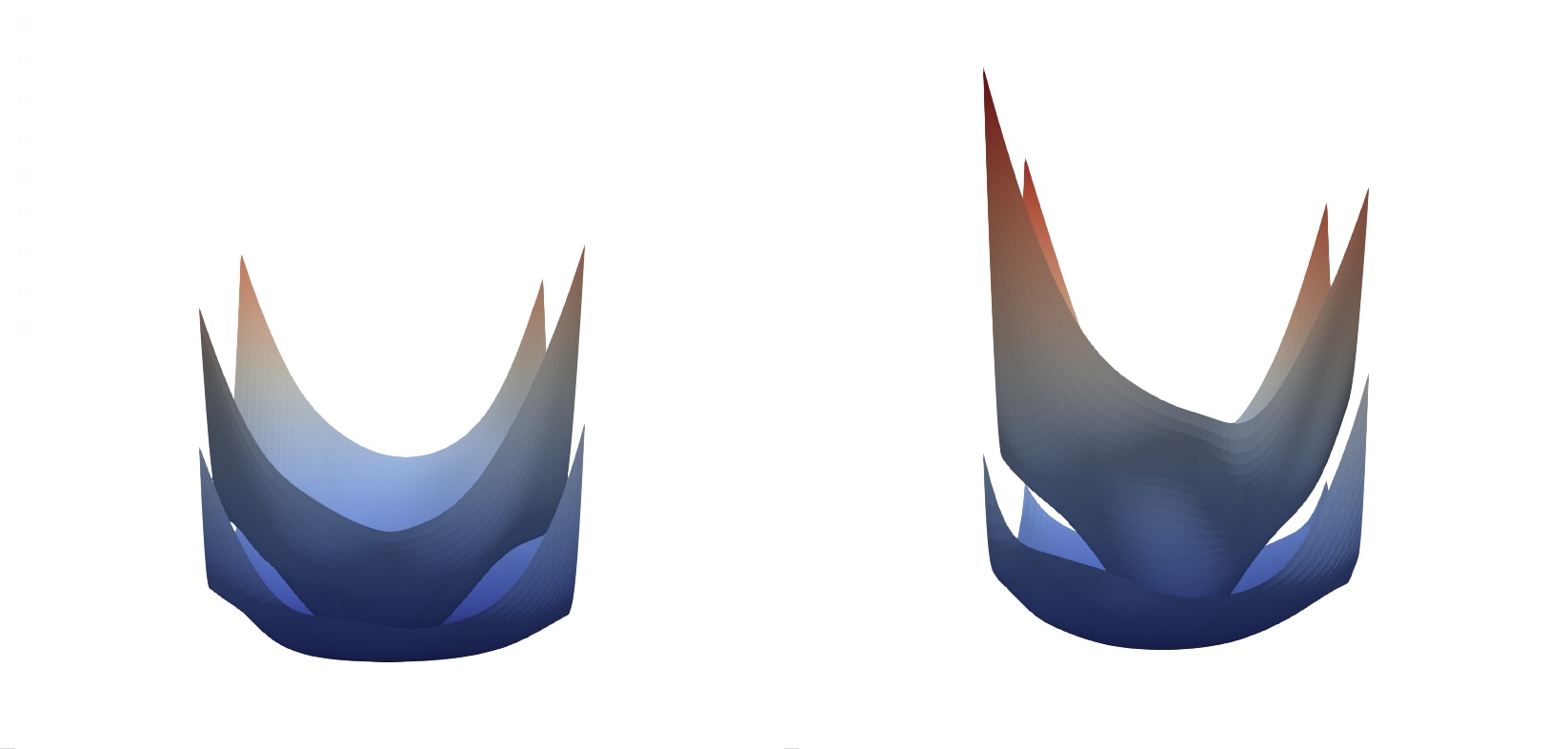}
  \caption{MultiFashion+MNIST}\label{fig:fashion-MNIST-loss}
\end{subfigure} 

\caption{Loss landscapes of task 1 and task 2 on MultiMNIST, MultiFashion and MultiFashion+MNIST, respectively. }
\label{fig:loss}
\end{figure}

Firstly, following \cite{li2018visualizing}, we provide additional visual comparisons of the loss landscapes trained with standard training and with our framework across two tasks of three datasets of Multi-MNIST in Figure \ref{fig:loss}. These are test loss surfaces of checkpoints that have the highest validation accuracy. The solution found by our proposed method not only mitigates the test loss sharpness for both tasks but also can reduce the test loss value itself, in comparison to traditional ERM. This is a common behavior when using flat minimizers as the gap between train and test performance has been narrowed \citep{DBLP:conf/uai/IzmailovPGVW18, kaddour2022fair}.

\subsection{Model robustness against weight perturbation}
\label{subsec:model_robustness}
Thirdly, to verify that SAM can orient the model to the common flat and low-loss region of all tasks, we measure the model performance within a $r$-radius Euclidean ball. To be more specific, we perturb parameters of two converged models by $\epsilon$, which lies in a $r$-radius ball and plot the accuracy of the perturbed models of each task as we increase $r$ from $0$ to $1000$. At each value of $r$, $10$ different models around the $r-$radius ball of the converged model are sampled.

\begin{figure}[!h]
    \centering
     \includegraphics[width=1\columnwidth, trim=.cm 0cm 0.cm 0cm,clip]{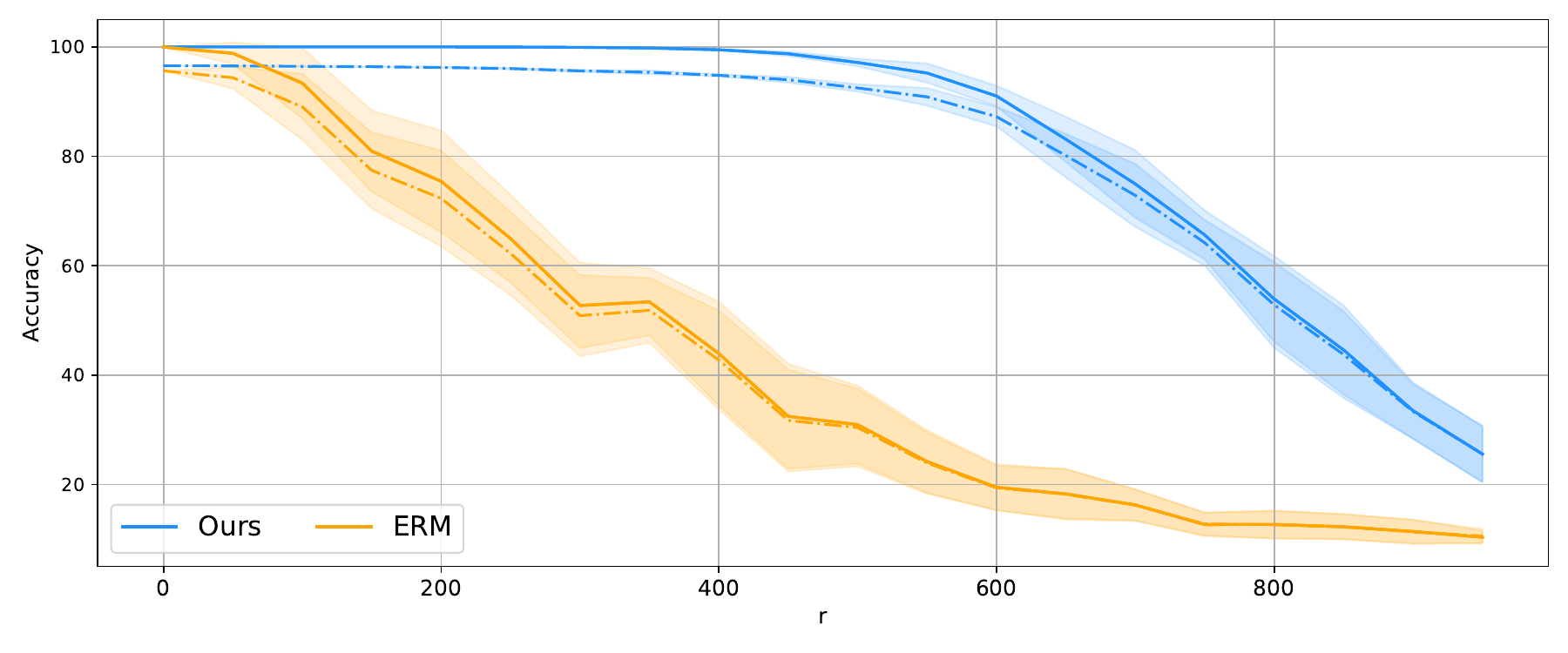}
    \caption{\textbf{Accuracy} within $r$-radius ball. Solid/dashed lines denote performance on train/test sets.
    \label{fig:local-flatness}}
\end{figure}

In Figure \ref{fig:local-flatness}, the accuracy of the model trained using our method remains at a high level when noise keeps increasing until $r=800$. This also gives evidence that our model found a region that changes slowly in loss. By contrast, the naively trained model loses its predictive capabilities as soon as the noise appears and becomes a dummy classifier that attains $10\%$ accuracy in a $10$-way classification. 

\subsection{Gradient conflict}
\begin{figure*}[!ht]
    \centering
     \includegraphics[width=1\linewidth, trim=.cm 0cm 0.cm 0cm,clip]{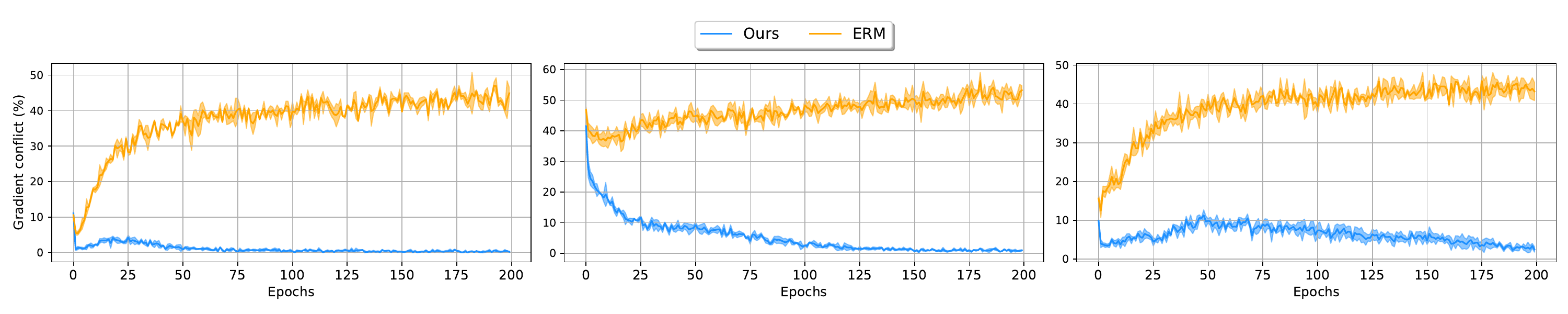}
    \caption{\textbf{Task gradient conflict proportion} of models trained with our proposed method and ERM across MultiFashion,  MultiFashion+MNIST, and MultiMNIST datasets (columns).}
    \label{fig:sup_conflict}
\end{figure*}

Secondly, in the main paper, we measure the percentage of gradient conflict on the MultiFashion+MNIST dataset. Here, we provide the full results on three different datasets. As can be seen from Figure \ref{fig:sup_conflict}, there is about half of the mini-batches lead to the conflict between task 1 and task 2 when using traditional training. Conversely, our proposed method significantly reduces such confliction to less than $5\%$ via updating the parameter toward flat regions.

\vspace*{1cm}

\subsection{Training curves}
\label{subsec:training_curves}

\begin{figure*}[!h]
    \centering
     \includegraphics[width=1\textwidth, trim=.cm 0cm 0.cm 0cm,clip]{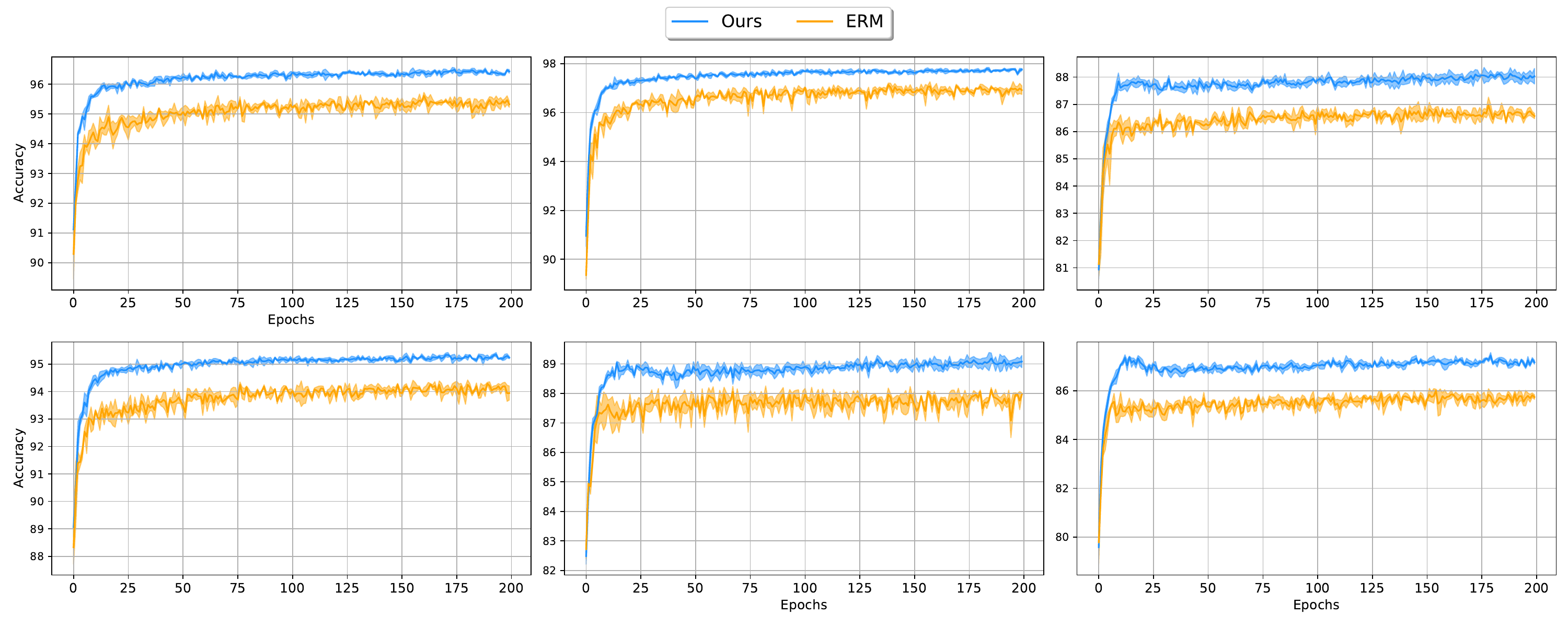}
    \caption{\textbf{Test accuracy} of models trained with our proposed method and ERM across 2 tasks (rows) of MultiFashion,  MultiFashion+MNIST and MultiMNIST datasets (columns).}
    \label{fig:test-accuracy}
\end{figure*}

Thirdly, we compare the test accuracy of trained models under the two settings in Fig. \ref{fig:test-accuracy}. It can be seen that from the early epochs ($20$-th epoch), the \textit{flat-based} method outperforms the \textit{ERM-based} method on all tasks and datasets. Although the ERM training model is overfitted after such a long training, our model retains a high generalizability, as discussed throughout previous sections.

\begin{figure*}[!h]
    \centering
     \includegraphics[width=1\textwidth, trim=.cm 0cm 0.cm 0cm,clip]{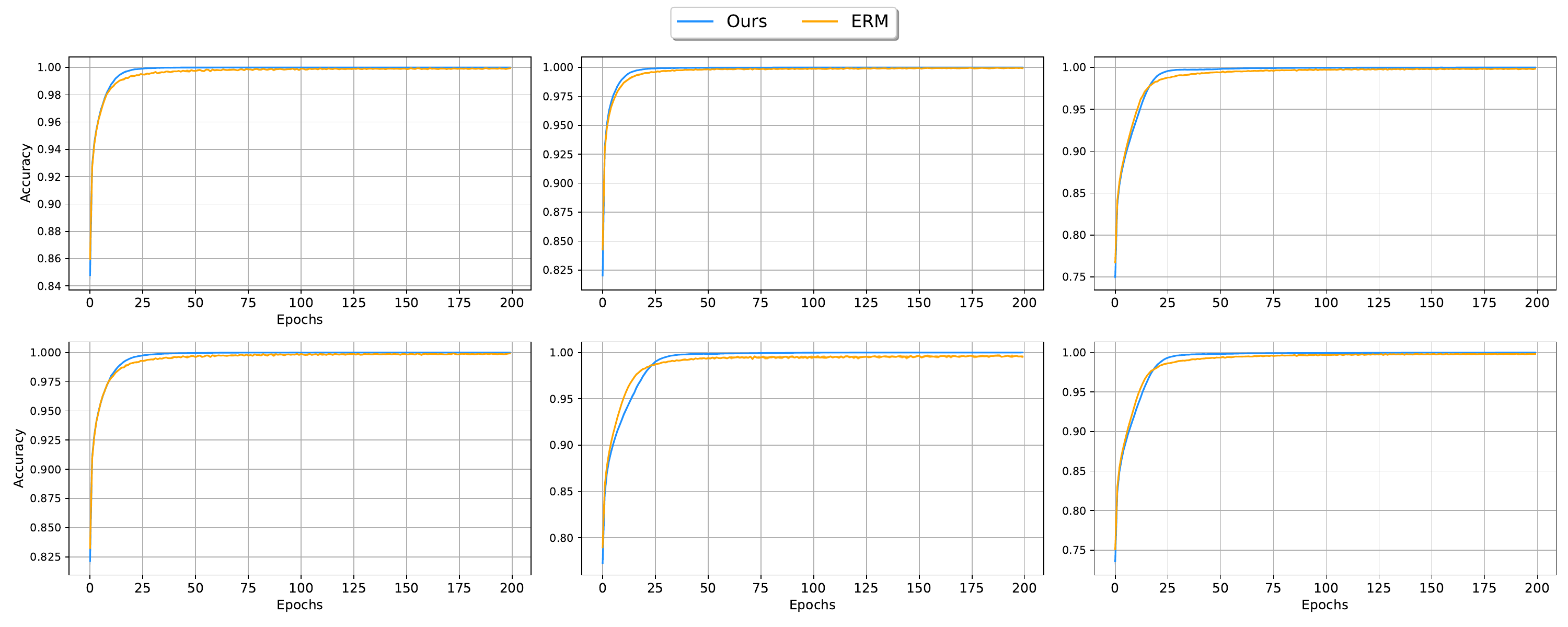}
    \caption{\textbf{Train accuracy} of models trained  with our proposed method and ERM across 2 tasks (rows) of MultiFashion,  MultiFashion+MNIST and MultiMNIST datasets (columns).}
    \label{fig:train-accuracy}
\end{figure*}

Furthermore, we also plot the training accuracy curves across experiments in Figure \ref{fig:train-accuracy} to show that training accuracy scores of both ERM and our proposed method are similar and reach $\approx100\%$ from $50$-th epoch, which illustrates that the improvement is associated with generalization enhancement, not better training.

\subsection{Model sharpness}
\label{subsec:model_sharpness_supp}

\begin{figure*}[!ht]
\centering

\begin{subfigure}{\linewidth}
    \centering
    \includegraphics[width=.95\textwidth, trim=.cm 0cm 0.cm 0cm,clip]{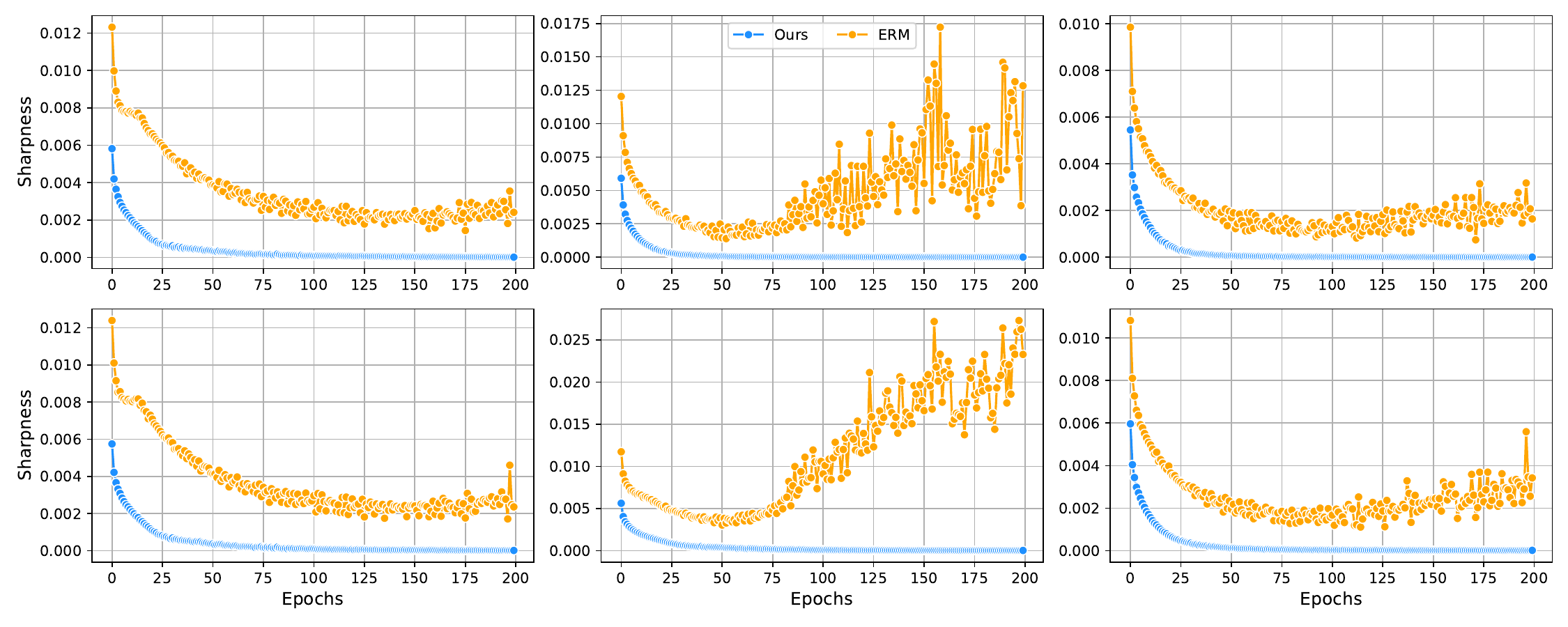}
    \caption{$\rho=0.005$}\label{fig:sharp_0.005}
\end{subfigure}

\begin{subfigure}{\linewidth}
    \centering
    \includegraphics[width=.95\textwidth, trim=.cm 0cm 0.cm 0cm,clip]{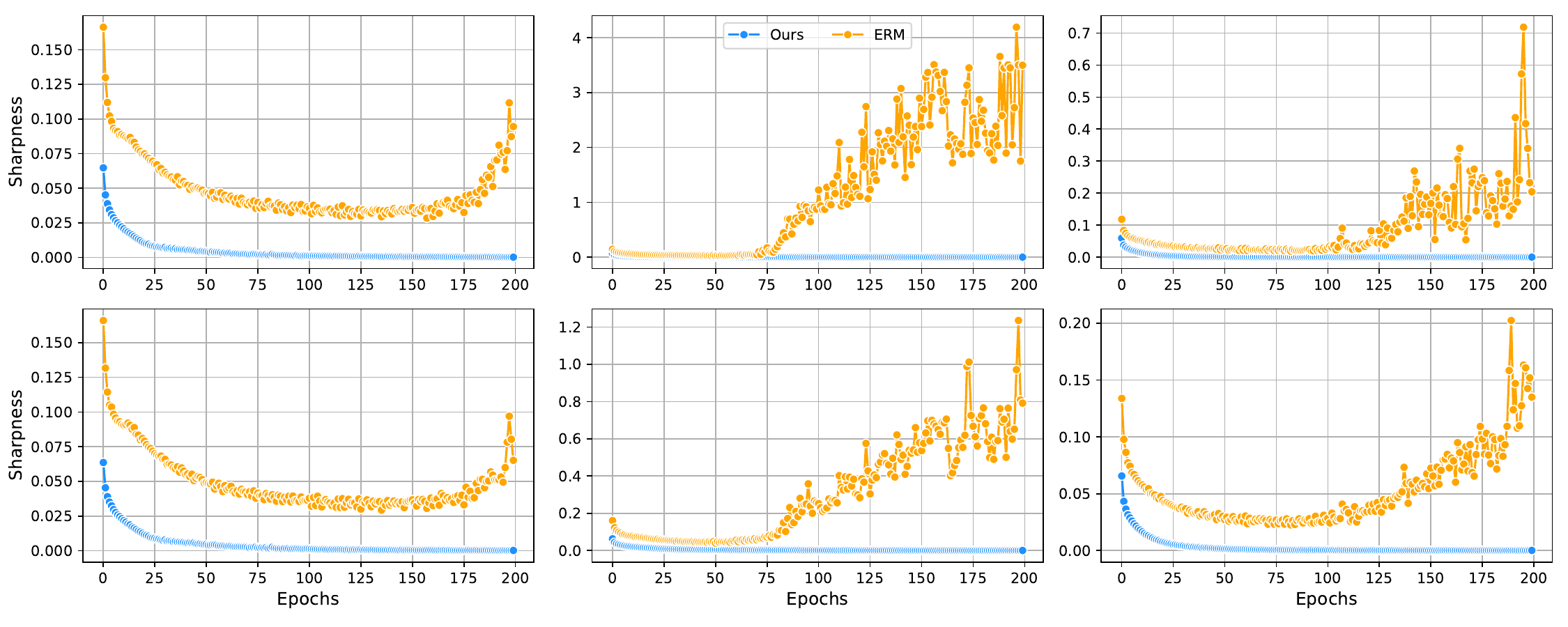}
    \caption{$\rho=0.05$}\label{fig:sharp_0.05}
\end{subfigure}

\begin{subfigure}{\linewidth}
    \centering
    \includegraphics[width=.95\textwidth, trim=.cm 0cm 0.cm 0cm,clip]{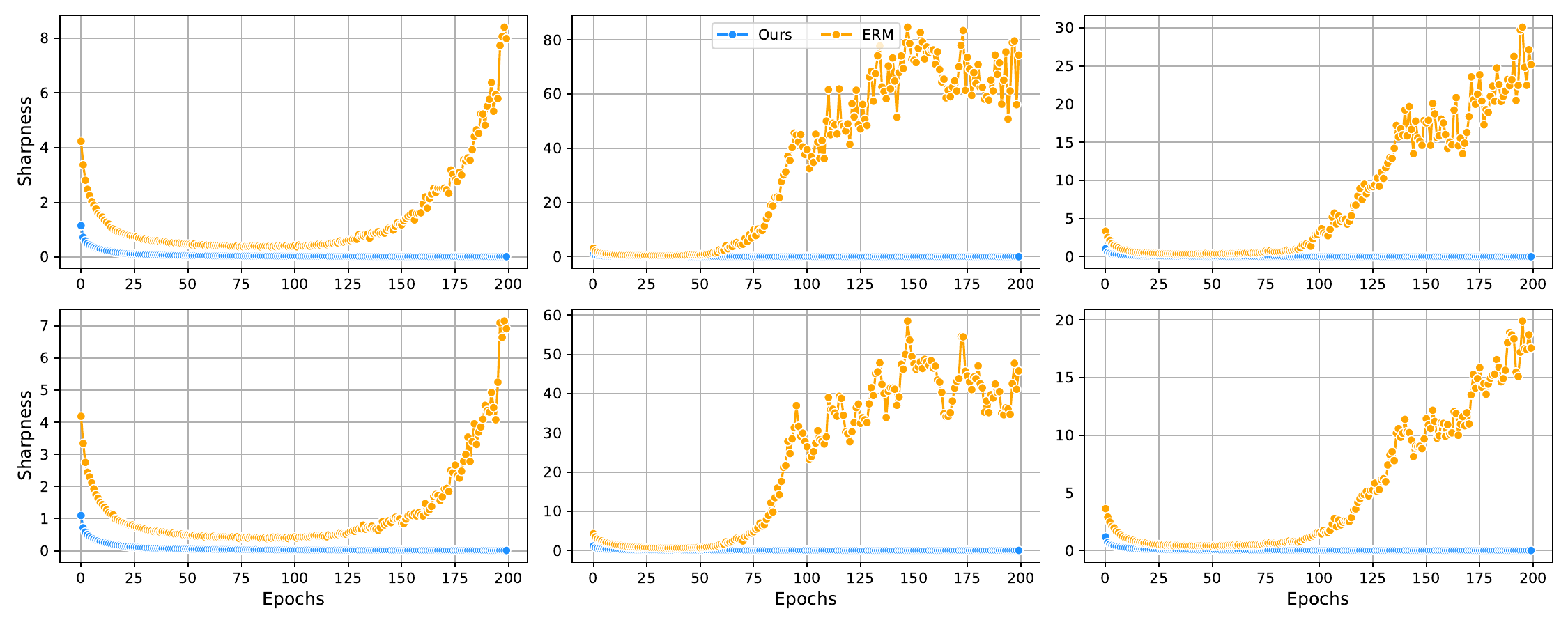}
    \caption{$\rho=0.5$}\label{fig:sharp_0.5}
\end{subfigure}

\caption{\textbf{Sharpness} of models trained  with our proposed method and ERM with different values of $\rho$. For each $\rho$, the top and bottom row respectively represents the first and second task, and each column respectively represents each dataset in Multi-MNIST: from left to right are MultiFashion, MultiFashion+MNIST, MultiMNIST.}
\label{fig:sharpness-supp}
\end{figure*}

 Fourthly,  Figure \ref{fig:sharpness-supp} displays the evolution of $\rho$-sharpness of models along training epochs under conventional loss function (ERM) and worst-case loss function (ours) on training sets of three datasets from Multi-MNIST, with multiple values of $\rho$. We can clearly see that under our framework, for both tasks, the model can guarantee uniformly low loss value in the $\rho$-ball neighborhood of parameter across training process. In contrast, ERM suffers from sharp minima from certain epochs when the model witnesses a large gap between the loss of worst-case perturbed model and current model. This is the evidence for the benefit that our framework brings to gradient-based methods, which is all tasks can concurrently find flat minima thus achieving better generalization. 

\subsection{Gradient norm}
\label{subsec:grad_norm_supp}

\begin{figure*}[!ht]
\centering

\begin{subfigure}{\linewidth}
    \centering
    \includegraphics[width=.95\textwidth, trim=.cm 0cm 0.cm 0cm,clip]{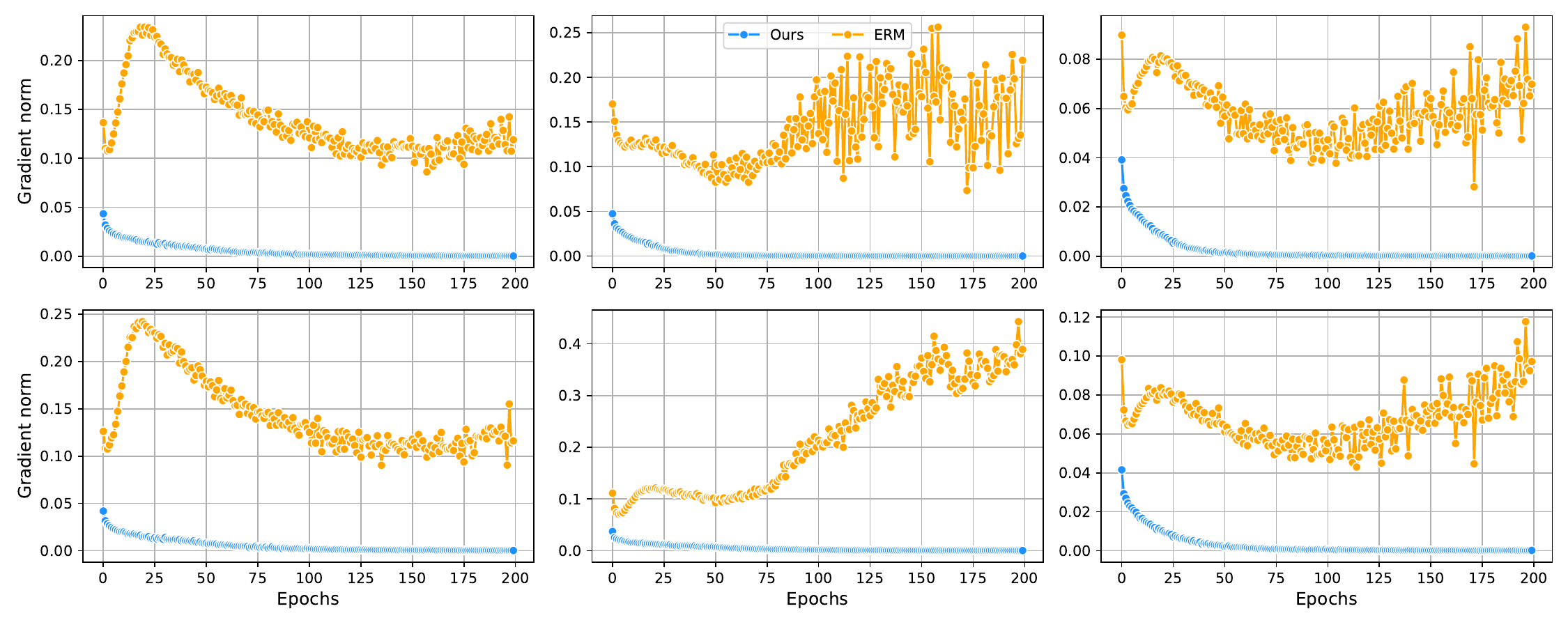}
    \caption{$\rho=0.005$}\label{fig:grad_0.005}
\end{subfigure}

\begin{subfigure}{\linewidth}
    \centering
    \includegraphics[width=.95\textwidth, trim=.cm 0cm 0.cm 0cm,clip]{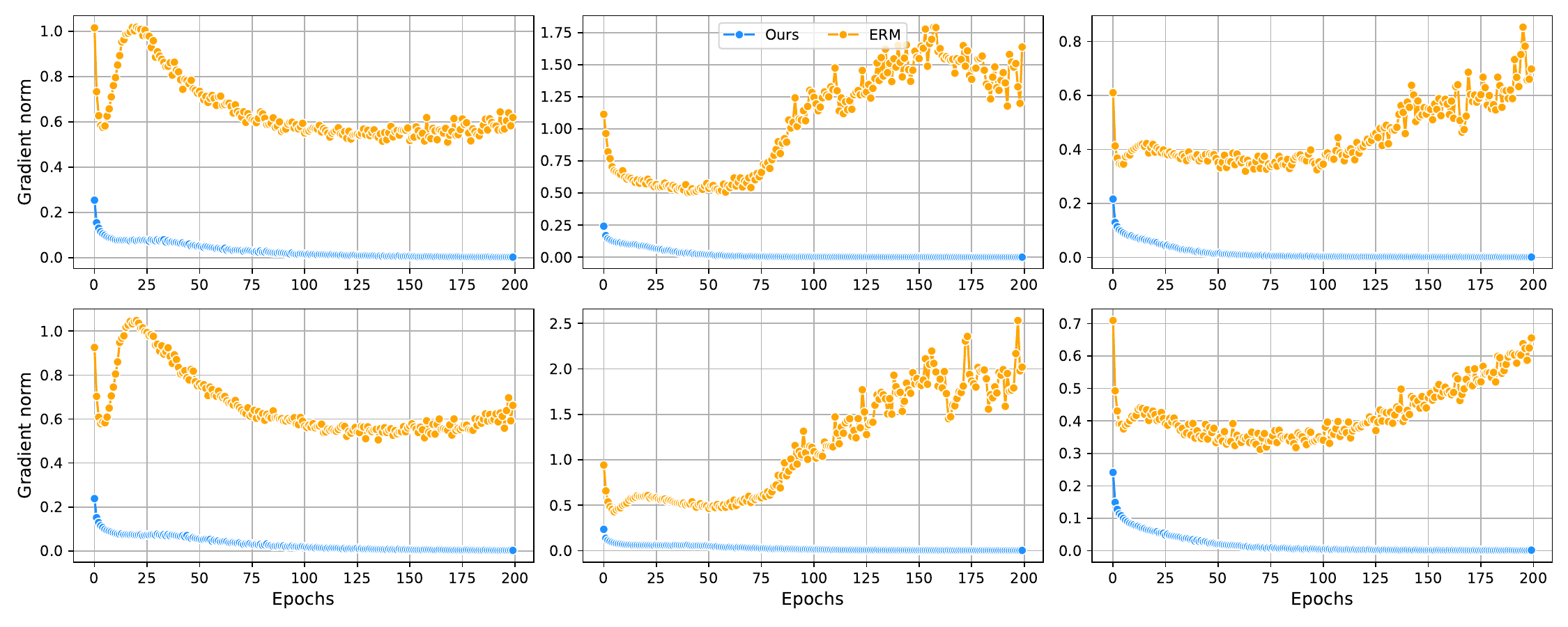}
    \caption{$\rho=0.05$}\label{fig:grad_0.05}
\end{subfigure}

\begin{subfigure}{\linewidth}
    \centering
    \includegraphics[width=.95\textwidth, trim=.cm 0cm 0.cm 0cm,clip]{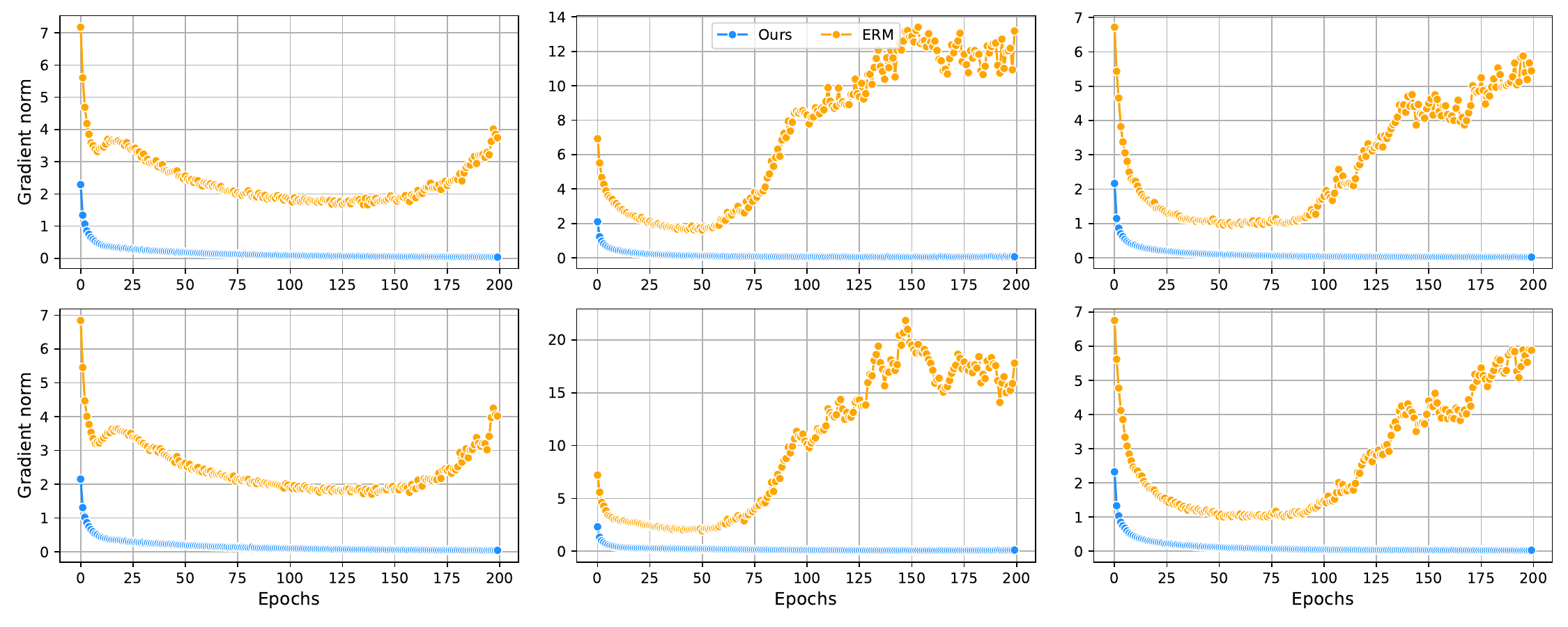}
    \caption{$\rho=0.5$}\label{fig:grad_0.5}
\end{subfigure}

\caption{\textbf{Gradient magnitude} at the worst-case perturbations of models trained with our proposed method and  ERM with different values of $\rho$. For each $\rho$, the top and bottom row respectively represents the first and second task, and each column respectively represents each dataset in Multi-MNIST: from left to right are MultiFashion, MultiFashion+MNIST, MultiMNIST.}
\label{fig:grad}
\end{figure*}

Finally, we demonstrate the gradient  norm of the loss function w.r.t the worst-case perturbed parameter of each task. On the implementation side, we calculate the magnitude of the flat gradient $\boldsymbol{g}^{i,\text{flat}}$ for each task
at different values of $\rho$ in Figure \ref{fig:grad}. As analyzed by Equation~6 
from the main paper, following the negative direction of $\boldsymbol{g}_{sh}^{i,SAM}$ will lower the $\mathcal{L}_2$ norm of the gradient, which orients the model towards flat regions. This is empirically verified in Figure \ref{fig:grad}. In contrast, as the number of epochs increases, gradnorm of the model trained with ERM tends to increase or fluctuate around a value higher than that of model trained with SAM.

\section{Discussion and Limitations}
The primary limitations of our work lie in time and space complexity. Specifically, our method demands an additional forward-backward pass to compute the worst-case gradient for each task, resulting in approximately twice the runtime compared to ERM counterparts. This could potentially be mitigated by employing a periodic update strategy as in ~\citep{Liu_2022_CVPR}, or by applying weight perturbation on a randomly chosen set of weights and data ~\citep{du2021efficient}, or even applying proposed training procedure on last few epochs \cite{zhou2024sharpness}. However, we leave this exploration for future work, as the main focus of our paper is to demonstrate the effectiveness of encouraging flatness in MTL.
In terms of space complexity, our approach requires approximately double the memory compared to traditional gradient-based methods. This is due to the need to store both the flat gradient and the loss gradient for each task as part of our gradient decomposition process.